\PassOptionsToPackage{table}{xcolor}
\documentclass{article}

\usepackage{iclr2027_conference,times}
\usepackage[T1,OT1]{fontenc}
\usepackage{microtype}

\usepackage{amsmath,amsfonts,bm}

\def\eqref#1{equation~\ref{#1}}
\def\1{\bm{1}}

\DeclareMathAlphabet{\mathsfit}{\encodingdefault}{\sfdefault}{m}{sl}
\SetMathAlphabet{\mathsfit}{bold}{\encodingdefault}{\sfdefault}{bx}{n}

\usepackage{booktabs}
\usepackage{array}
\usepackage{graphicx}
\usepackage{amssymb}
\usepackage[hidelinks]{hyperref}
\usepackage{url}
\usepackage[table]{xcolor}
\usepackage{placeins}
\usepackage{float}
\usepackage{wrapfig}
\usepackage{needspace}
\usepackage{tikz}
\usepackage{tabularx}
\usepackage[skins]{tcolorbox}
\usepackage{booktabs}
\usepackage{enumitem}

\usetikzlibrary{arrows.meta,calc,fit,positioning,shapes.geometric}

\definecolor{sbInk}{HTML}{253447}
\definecolor{sbBlue}{HTML}{3B82F6}
\definecolor{sbOrange}{HTML}{F59E0B}
\definecolor{sbGreen}{HTML}{10B981}
\definecolor{sbPurple}{HTML}{8B5CF6}
\definecolor{sbRed}{HTML}{EF4444}
\definecolor{sbPanel}{HTML}{F5F7FA}
\definecolor{sbTableHead}{HTML}{EAF0F5}
\definecolor{sbTableSubhead}{HTML}{F4F7FA}

\newcommand{\tbest}[2][sbBlue]{\cellcolor{#1!18}\textbf{#2}}
\newcommand{\tsecond}[2][sbBlue]{\cellcolor{#1!8}#2}

\usepackage{xcolor}

\newcommand{\compacttable}{%
  \renewcommand{\arraystretch}{1.05}%
  \setlength{\extrarowheight}{0pt}%
  \setlength{\aboverulesep}{2pt}%
  \setlength{\belowrulesep}{2pt}%
  \setlength{\abovetopsep}{0pt}%
  \setlength{\belowbottomsep}{0pt}%
  \setlength{\defaultaddspace}{3pt}%
}
\AddToHook{env/table/begin}{\compacttable}
\AddToHook{env/table*/begin}{\compacttable}

\AddToHook{env/figure/begin}{\setlength{\parskip}{0pt}}
\AddToHook{env/figure*/begin}{\setlength{\parskip}{0pt}}
\AddToHook{env/table/begin}{\setlength{\parskip}{0pt}\setlength{\abovecaptionskip}{0pt}\setlength{\belowcaptionskip}{4pt}}
\AddToHook{env/table*/begin}{\setlength{\parskip}{0pt}\setlength{\abovecaptionskip}{0pt}\setlength{\belowcaptionskip}{4pt}}

\newtcolorbox[auto counter]{takeaway}[1][]{
  enhanced, frame hidden, boxrule=0pt, sharp corners,
  colback=sbPanel, borderline west={1.2pt}{0pt}{sbInk!45},
  boxsep=0pt, left=7pt, right=7pt, top=5pt, bottom=5pt,
  before skip=7pt, after skip=7pt, fontupper=\normalsize,
  before upper={\textbf{\textcolor{sbInk}{Takeaway~\thetcbcounter.}}\ },
  #1
}

\graphicspath{{figures/}}

\title{Code4Scene: Benchmarking Coding Agents for Constructing and Editing 3D Scenes}

\input{authors}

\usepackage{etoolbox}
\makeatletter
\patchcmd{\@maketitle}{\vskip 0.3in minus 0.1in}{\vskip 0.12in}{}{}
\makeatother

\iclrfinalcopy
\AddToHook{cmd/maketitle/after}{%
  \fancyhead{}\lhead{Preprint}\renewcommand{\headrulewidth}{0.4pt}%
}
\hypersetup{
  pdftitle={Code4Scene: Benchmarking Coding Agents for Constructing and Editing 3D Scenes},
  pdfauthor={Xiaokang Ye, Siddhant Hitesh Mantri, Zimeng Chen, Edward Zhang, Zhaoxu Zheng, Yuanheng Li, Yizhao Chen, Tianyang Huang, Lianhui Qin},
  pdfsubject={Preprint working draft}
}

\begin{document}
% Keep natural paragraph spacing instead of stretching gaps to fill float pages.
\raggedbottom

\maketitle
% The contents list at the start of the appendix excludes main-text sections.
\addtocontents{toc}{\protect\setcounter{tocdepth}{-2}}

% Keep the author block, teaser, and complete abstract on the title page.
\begin{figure}[!ht]
    \centering
    \includegraphics[width=0.90\linewidth,trim=0bp 4.5bp 0bp 18bp,clip]{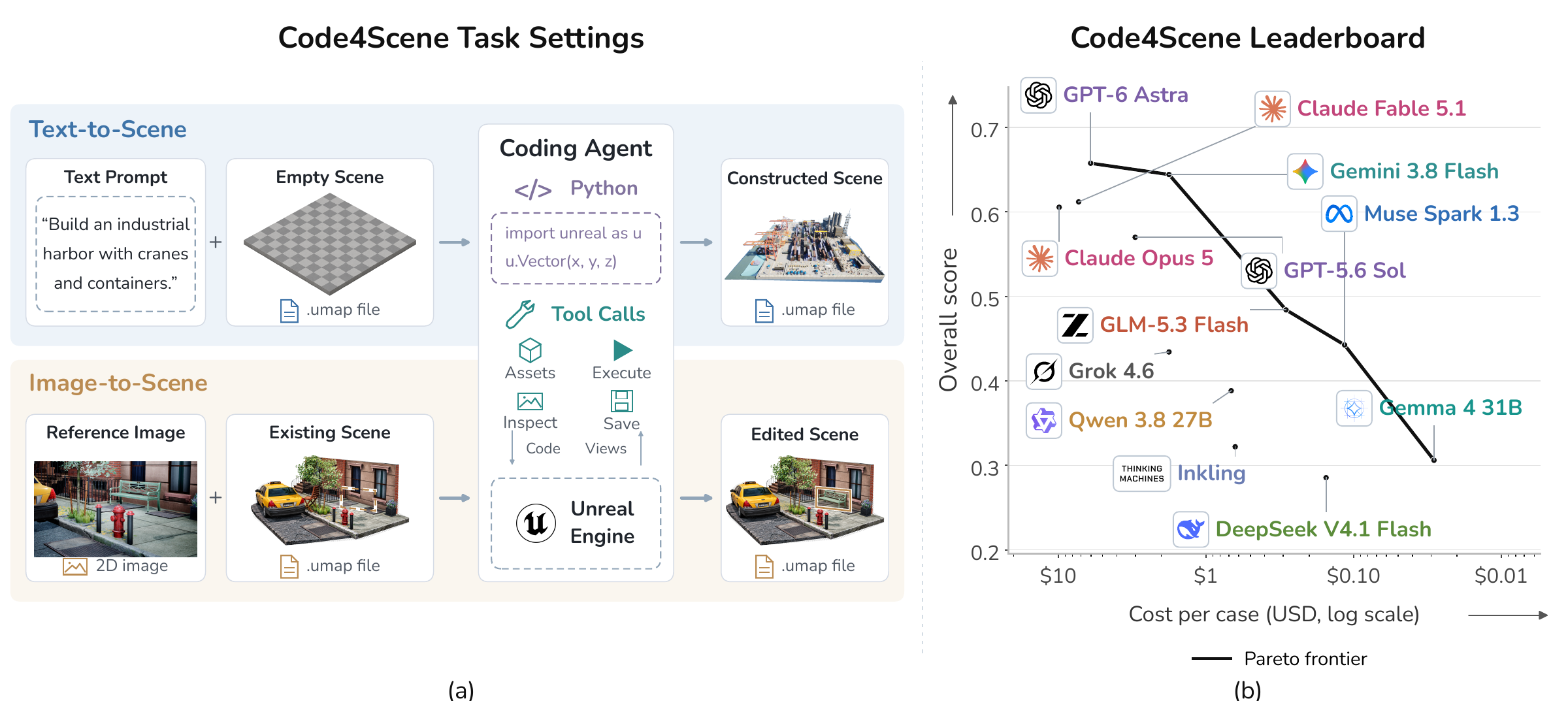}
    \caption{Code4Scene tasks and leaderboard. (a) Coding agents use Python and tool calls to construct or edit persistent Unreal Engine scenes based on text prompt or image reference. (b) Current overall scores versus matched estimated token cost per case on the retained public set, using fixed September 2026 list rates (Appendix~\ref{app:cost-estimate}). The line marks the frontier among displayed systems; Table~\ref{tab:rerun-cost-coverage} reports accounting coverage.}
    % "Construction Scene" too samll
    % "Edited Scene" not obvious
    \label{fig:teaser}
\end{figure}
\FloatBarrier

\begin{abstract}

Frontier coding agents can now write and execute code that authors 3D environments, but whether they reliably understand 3D structure and precisely control scene state remains unclear. The generated 3D scene is a persistent, executable artifact: a convincing render can hide incorrect spatial relations, intersecting objects, or unintended modifications. We introduce \textbf{Code4Scene}, a benchmark of 170 Unreal Engine cases built from human-assembled scenes that evaluates coding agents on two complementary settings under a shared execution interface. \emph{Construction} tests scene-level spatial reasoning from open-ended language specifications, where many realizations are valid; \emph{editing} tests precise control of scene state, where the agent must recover the target scene from reference images while preserving everything else. Rather than scoring code or rendered views, Code4Scene evaluates the generated engine-native scene for task fulfillment, artifact integrity, and static physical validity, with edits additionally compared against withheld ground truth. Across 14 coding-agent systems on the 83-case public set, construction and editing performance are strongly correlated but not interchangeable (Spearman $\rho = 0.67$): Claude Fable 5.1 leads construction, Gemini 3.8 Flash leads editing. Spatial Composition is the weakest construction category for every agent, while editing remains imprecise: the best Repair F1 is only 0.562, and 34.2\% of edits that fully recover the target still introduce unintended changes elsewhere in the scene. These results expose a gap between plausible 3D generation and reliable spatial reasoning and state control. The project webpage is at \url{https://simworld.org/code4scene}.

\end{abstract}

\clearpage
% The preprint has no page limit: give body floats and captions natural space.
% Keep the title-page author/teaser/abstract arrangement unchanged.
\setlength{\textfloatsep}{14pt plus 3pt minus 2pt}
\setlength{\floatsep}{10pt plus 2pt minus 2pt}
\setlength{\intextsep}{10pt plus 2pt minus 2pt}
\setlength{\dbltextfloatsep}{14pt plus 3pt minus 2pt}
\setlength{\dblfloatsep}{10pt plus 2pt minus 2pt}
\setlength{\abovecaptionskip}{6pt}
\setlength{\belowcaptionskip}{2pt}
% Prefer mixed text/float pages; reserve float-only pages for large figures.
\renewcommand{\floatpagefraction}{0.80}
\renewcommand{\dblfloatpagefraction}{0.80}
\renewcommand{\topfraction}{0.88}
\renewcommand{\textfraction}{0.12}
\AddToHook{env/table/begin}{\setlength{\belowcaptionskip}{6pt}}
\AddToHook{env/table*/begin}{\setlength{\belowcaptionskip}{6pt}}
\makeatletter
\setlength{\@fpsep}{12pt plus 2pt minus 2pt}
\makeatother
\section{Introduction}
\label{sec:introduction}

Frontier coding agents are beginning to directly author 3D environments. Recent systems can write and execute scene-editing code, operate graphics engines, and iteratively refine scenes from visual feedback. Demonstrations such as GPT-6 Astra constructing a model in Blender~\citep{blender2026} and transferring it into a walkable Unreal Engine~\citep{epic2026unreal} scene~\citep{openai2026gpt6astra} suggest rapid progress in spatially grounded generation. This capability could have broad value in game development, industrial design, and robotics simulation. But its reliability remains unclear: \textbf{can frontier coding agents reliably understand 3D structure and precisely control executable scene state?}

3D scene authoring makes this question difficult to answer because the output is not just code or an image, but a persistent artifact whose state must remain correct after execution. An agent must translate textual or visual intent into engine operations, inspect the resulting changes, and revise the scene without breaking what already works. A convincing render can still hide incorrect spatial relations, unsupported or intersecting objects, broken dependencies, or unintended modifications.  

Recent work has begun to automate 3D scene creation with language models and coding agents, while new benchmarks evaluate scene generation, reconstruction, editing, and executable worlds~\citep{kang2026simworldstudio,liu2026simworlds,gu2025blendergym,xu2026muse,lu2026worldcoder}. These evaluations, however, differ substantially in engine, scene representation, agent interface, and evaluation target, ranging from rendered outputs to structured scene states and engine-specific artifacts. It remains difficult to compare general-purpose coding agents in a common setting or distinguish visual plausibility from spatial correctness, physical validity, and precise state control.

To evaluate these capabilities systematically, we study two complementary aspects of 3D scene authoring: \textbf{construction} and \textbf{editing}. Construction probes scene-level spatial reasoning: given an open-ended language specification, an agent must compose a coherent scene, and many realizations may be valid. Editing probes precise control of scene state: given an existing scene and reference images of the desired result, the agent must infer the required changes and apply them while preserving everything else. The two settings therefore capture complementary aspects of 3D scene authoring and require different notions of correctness.

We introduce \textbf{Code4Scene}, a benchmark of 170 Unreal Engine cases for evaluating coding agents on 3D scene construction and editing under a shared execution interface. Rather than scoring only code or rendered views, Code4Scene evaluates the generated, persistent engine-native scene, including task fulfillment, artifact integrity, and static physical validity. Construction is evaluated by semantic and spatial adherence to the specification, while editing is compared against withheld ground truth using actor-level Repair F1. This lets us distinguish scenes that merely look plausible from those that are correct, valid, and precisely controlled.

We evaluate 14 coding-agent systems on the 83-case public set. Construction and editing performance are strongly correlated but not interchangeable (Spearman $\rho = 0.67$): Claude Fable 5.1 leads construction, Gemini 3.8 Flash leads editing, and GPT-6 Astra has the highest observed overall mean, with its advantage over Gemini unresolved by the paired source-cluster interval. \textbf{Spatial Composition is the weakest construction category for every agent}, and even the best agent's (GPT-6 Astra) Spatial Composition score reaches only 0.587. In the same selected runs, 19.1\% of analyzable requirements in the Spatial Composition family whose relevant entities are confirmed present are still judged mismatched, showing that generating the right elements does not imply arranging them correctly. Editing remains difficult as well, with the best Repair F1 reaching only 0.562; even among edits that fully recover the target, 34.2\% still introduce unintended changes elsewhere in the scene. Together, these results expose a clear gap between generating plausible 3D content and reliably reasoning about its spatial structure and controlling its state.

% s
% % TODO: Keep this findings paragraph synchronized with the experiment takeaways.

% \Needspace{3\baselineskip}
% Our contributions are threefold.

% \begin{itemize}[leftmargin=*]
% \item We introduce \textbf{Code4Scene}, a 190-case Unreal benchmark for
% open-ended construction and reference-guided editing under a shared interface
% and submission protocol.

% \item We develop an independent evaluation framework for artifact integrity,
% physical safety, and task fulfillment. We assess Text-to-Scene agreement with
% human judgments, fixed-evidence Detailed Alignment repeatability, and downstream
% usability with a deployed player agent.

% \item We compare 14 coding-agent systems on 95 public cases, separating
% task fulfillment from physical safety and identifying spatial-composition
% challenges. We analyze Python-based editing and historical correlations between
% construction scores and navigation utility.
% \end{itemize}

% Queue the comparison after the introduction so it floats to page 3.
\begin{table}[!t]
\centering
\caption{Comparison of related benchmarks and their reported coverage. Input
references specify the desired outcome and exclude initial-scene observations.
Open-ended settings admit multiple valid scene realizations; ground-truth-based
(GT-based) settings use a hidden target state for evaluation. Full related work discussions appear in Appendix~\ref{app:additional-related-work}.}
\label{tab:benchmark-positioning}
\footnotesize
\setlength{\tabcolsep}{1.5pt}
\begin{tabularx}{\linewidth}{>{\raggedright\arraybackslash}Xl*{7}{c}}
\toprule
\rowcolor{sbTableHead}
\textbf{Benchmark} & \textbf{Artifact} &
\multicolumn{2}{c}{\textbf{Input}} &
\multicolumn{2}{c}{\textbf{Task Setting}} &
\multicolumn{3}{c}{\textbf{Evaluation}} \\
\cmidrule(lr){3-4}\cmidrule(lr){5-6}\cmidrule(l){7-9}
& & Text & Images & Open-ended & GT-based & Semantic & Physical & Preservation \\
\midrule
BlenderGym \citeyearpar{gu2025blendergym} & Script &
& $\checkmark$ & & $\checkmark$ & $\checkmark$ & & \\
SceneActBench \citeyearpar{zhao2026sceneactbench} & GLB/pose &
& $\checkmark$ & & $\checkmark$ & $\checkmark$ & & \\
4DBuildBench \citeyearpar{liu2026simworlds} & \textrm{.blend} &
$\checkmark$ & & $\checkmark$ & & $\checkmark$ & $\checkmark$ & \\
CutsceneBench \citeyearpar{he2026cutscene} & Sequence &
$\checkmark$ & & $\checkmark$ & $\checkmark$ & $\checkmark$ & & \\
WorldCoder-Bench \citeyearpar{lu2026worldcoder} & HTML &
$\checkmark$ & & $\checkmark$ & & $\checkmark$ & $\checkmark$ & \\
AuthorBench \citeyearpar{xu2026muse} & JSON &
$\checkmark$ & & $\checkmark$ & & $\checkmark$ & $\checkmark$ & $\checkmark$ \\
VWE-Bench \citeyearpar{ning2026vibeworlding} & Asset map &
$\checkmark$ & & $\checkmark$ & $\checkmark$ & $\checkmark$ & $\checkmark$ & $\checkmark$ \\
\midrule
\rowcolor{sbBlue!9}
\textbf{Code4Scene (ours)} & \textrm{.umap} &
$\checkmark$ & $\checkmark$ & $\checkmark$ & $\checkmark$ & $\checkmark$ & $\checkmark$ & $\checkmark$ \\
\bottomrule
\end{tabularx}
\end{table}

\section{Code4Scene Benchmark}
\label{sec:scenebench}

\subsection{Task Settings}
\label{sec:problem-formulation}

Code4Scene formulates 3D scene authoring as \textit{agentic code generation for 3D scenes}.
Rather than evaluating dedicated 3D reconstruction methods
(e.g., 3D Gaussian Splatting~\citep{kerbl2023gaussian}) or LLM-based layout
generation systems~\citep{feng2023layoutgpt,yang2024holodeck}, we evaluate
general-purpose coding agents that directly construct and edit 3D scenes
through code and engine tools.
The agent decides how to retrieve assets, write and execute code, inspect
the generated scene, and revise it in Unreal Engine. The output is an
engine-native scene (\texttt{.umap}) that can be reopened, edited, and
simulated. We evaluate this generated scene and record the agent's
interaction trajectory for analysis (Appendix~\ref{app:agent-interface}).

Within this formulation, we consider two complementary scene-authoring
activities: creating a new scene from textual requirements and modifying an
existing scene to match a reference. The first tests whether an agent can
turn open-ended intent into a spatially coherent scene, while the second tests
whether it can infer the required changes and apply them without disrupting
the rest of the scene. Code4Scene instantiates these as
\textbf{Text-to-Scene Construction} and \textbf{Image-to-Scene Editing} under
the same engine and agent interface.

Formally, each case derives from a human-assembled scene
\(\mathcal{L}^{\star}\). From it we build an asset catalog \(\mathcal{C}\), an
initial scene \(\mathcal{L}^{0}\), agent-visible evidence \(E\), and, for
editing, evaluator-only references \(H\). Under a shared interface
\(\mathcal{A}\) and a resource budget \(B\), an agent configuration produces
the generated scene through
\(f:(\mathcal{L}^{0},\mathcal{C},E,\mathcal{A},B)\rightarrow\hat{\mathcal{L}}\).
The two settings differ in \(\mathcal{L}^{0}\), \(E\), and \(H\).

\textbf{Text-to-Scene Construction.}
The agent starts from an empty scene and receives a scene description and the
asset catalog (\(\mathcal{L}^{0}\) is blank, \(E\) is the description, and
\(H=\varnothing\)). The description admits many valid scenes, so
\(\mathcal{L}^{\star}\) guides prompt writing and the catalog but is not a
reconstruction target. The agent must identify the requested entities, counts,
attributes, and style, and match each request to assets in the catalog. It must
place objects so that the stated spatial relations hold, keep them supported
and free of interpenetration, and save a scene whose dependencies resolve.
In our test set, a prompt contains 58 checkable requirements on average (up to
92). These demands motivate complementary checks of prompt fulfillment,
physical plausibility, and artifact validity, captured by the
\textit{Semantic Verifier}, \textit{Physical Safety}, and
\textit{Candidate Integrity}, respectively
(Sections~\ref{sec:evaluation-protocol} and~\ref{sec:evaluation}).

\textbf{Image-to-Scene Editing.}
The agent receives a corrupted copy of a human-assembled scene
(\(\mathcal{L}^{0}\)) and reference views of the original scene (\(E\)).
\(H\) contains the original scene, held-out views, and its serialized state,
accessible only to the evaluator (Section~\ref{sec:evaluation-protocol}).
The agent must compare the references with a scene containing approximately
95 to 6{,}700 actors and identify objects that were removed, inserted,
duplicated, replaced, moved, rotated, rescaled, or mirrored. It must restore
the intended scene state while preserving content that does not require repair.
Restored objects must meet tight tolerances: 5\,cm in position,
\(5^{\circ}\) in rotation, and 5\% in scale. Missed targets lower recall,
while incorrect target states and collateral edits lower precision
(Section~\ref{sec:evaluation}). The target is specified through images, so
the agent must infer the required changes from the reference views.

\subsection{Data Curation}
\label{sec:benchmark-suite}

\begin{figure}[!t]
\centering
\includegraphics[width=\linewidth,trim=0bp 11.5bp 0bp 8bp,clip]{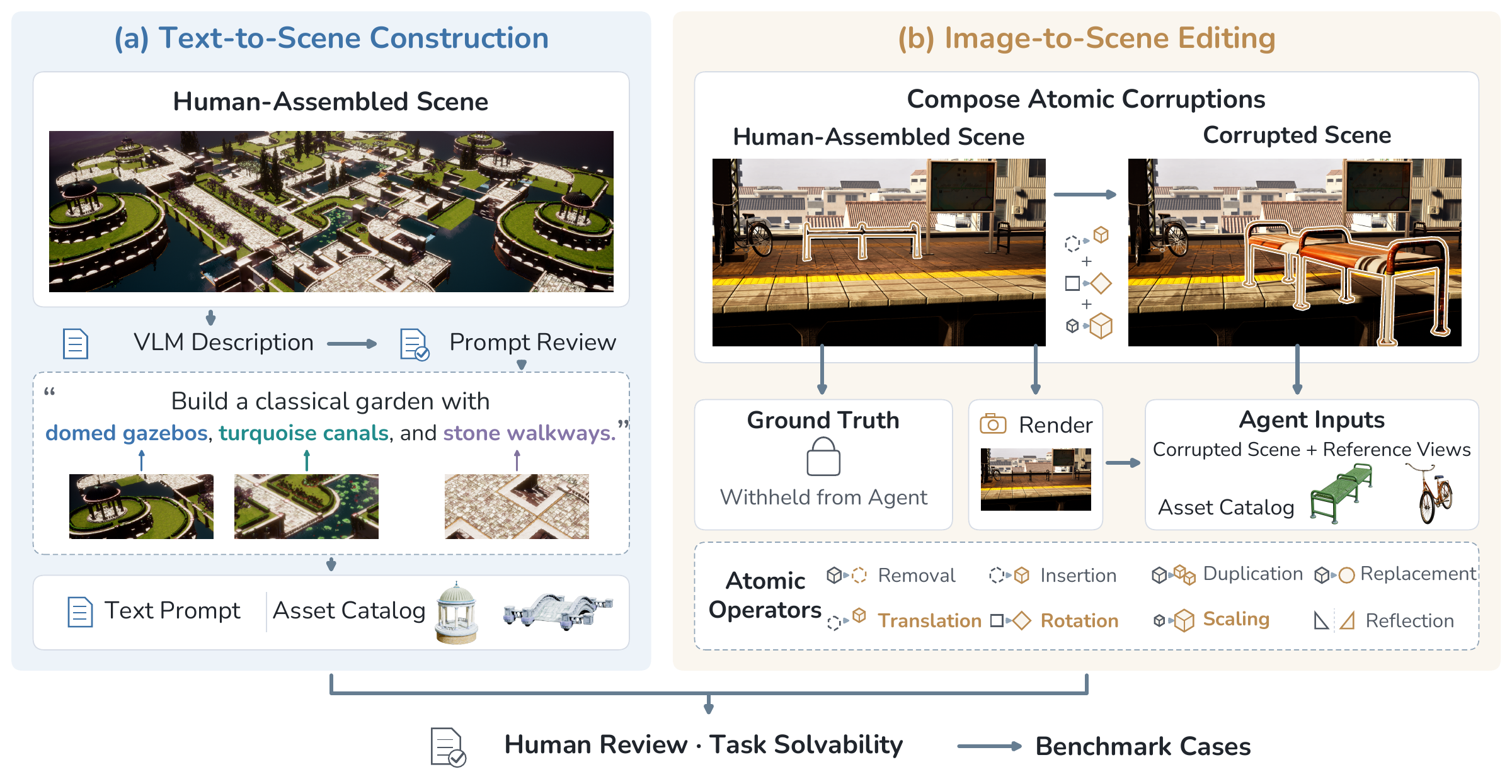}
\caption{Task construction. Text-to-Scene prompts describe human-assembled
scenes. Image-to-Scene inputs are corrupted copies, with the human-assembled
scene retained as ground truth. Human review checks task solvability in both
settings.}
\label{fig:task-construction}
\end{figure}

% Scene terminology standardized as Human-Assembled Scene.
% Delete empty scene
% Assets catalog need concrete assets
% (b) Initial Scnee -> Corrupted Scene
% (b) Arrows need to be bigger

% \textbf{Scale.}

Code4Scene derives construction and editing tasks from human-assembled Unreal Engine scenes to reflect the complexity of scene authoring in practice (Figure~\ref{fig:task-construction}). The benchmark contains 170 retained cases: 26 Text-to-Scene Construction tasks and 144 Image-to-Scene Editing tasks, comprising 72 indoor and 72 outdoor cases. Each case involves a long-horizon authoring task in a production-scale scene, with construction runs allowing up to 800 tool calls and 180 minutes. Every case is manually curated and reviewed for solvability. We therefore prioritize realistic, verifiable tasks over maximizing case count. The benchmark is comparable in size to other execution-based 3D benchmarks (Table~\ref{tab:benchmark-positioning}), including 4DBuildBench (50 scenes), CutsceneBench (65 scenarios), and BlenderGym (245 start--goal pairs) \citep{liu2026simworlds,he2026cutscene,gu2025blendergym}.

% We curate cases around three requirements: each case should reflect realistic scene-authoring workflows, support reliable evaluation, and be solvable from the evidence available to the agent (Figure~\ref{fig:task-construction}).

\paragraph{Human-assembled scenes.}
We select human-assembled Unreal Engine scenes from a source collection of 522 Fab content packs\footnote{Epic Games' digital asset marketplace: \url{https://www.fab.com/}.} \citep{fab}, requiring high visual quality, non-trivial compositional complexity, and assets that can be loaded and identified reproducibly. The collection spans historical, urban, industrial, domestic, and stylized environments. The retained benchmark has 26 Text-to-Scene cases and eight editing tasks for each of 18 Image-to-Scene source scenes (nine indoor and nine outdoor).

An asset-use audit excluded cases requiring Bazaar\_Meshingun, ChemicalPlantEnv, DesertTown, MiddleEast or DetectiveOffice, whose listings were identified as not allowing AI use. We also excluded the retired TrainStation cases because AI-use permission was unconfirmed. This removes 20 cases from the original 190-case suite, including 12 of the original 95 public cases: seven involving the five No-AI packs and five involving TrainStation. All configurations use the same retained-case manifest; failed submissions on retained tasks still count as zero. Replacement scenes require fresh model evaluations and are not included in these results. Human-assembled scenes expose agents to irregular layouts and intentional object arrangements; for editing, they also provide exact ground truth for the intended repair.

\paragraph{Text-to-Scene task construction.}
We ground each text prompt in the assets and composition of a human-assembled scene. We first render screenshots and use a vision-language model (VLM) to generate an initial description. Two reviewers independently verify the description against the scene and consolidate their corrections into the final prompt. At test time, the agent starts from an empty scene and receives only the text prompt and the corresponding asset catalog. The human-assembled scene guides prompt and asset selection, while the prompt defines the construction task and allows multiple valid solutions. We automatically extract a fixed set of requirements from each prompt and reuse it across submissions, so all agents face the same checklist (Appendix~\ref{app:semantic-verifier}).

\paragraph{Image-to-Scene task construction.}
Image-to-Scene evaluates fine-grained editing against an exact ground-truth target. We retain the human-assembled scene as withheld ground truth and create each test input by applying one or more atomic corruptions to a copy, such as \textit{removal}, \textit{insertion}, \textit{duplication}, or \textit{replacement}. The original scene defines the intended repair, allowing us to measure both target recovery and preservation of the surrounding scene. Authors experienced with Unreal Engine design each corruption to ensure that the required edit is meaningful in context. At test time, the agent receives the corrupted scene, the asset catalog, and rendered reference views of the original scene: two views for each indoor case and one for each outdoor case. We select scenes whose reference views provide sufficient visual evidence for the repair and whose editable objects are stable, individually addressable actors (Appendix~\ref{app:case-construction-details}).

\paragraph{Human review of solvability.}
Three independent reviewers who did not author the cases review every case for solvability from the resources available to the agent. For Text-to-Scene, they check that the prompt can be satisfied using the provided asset catalog; for Image-to-Scene, they check that every required repair can be inferred from the reference images.

\paragraph{Public and private sets.}
Following previous work including ARC-AGI-3 \citep{foundation2026arc}, Code4Scene separates public and private evaluation cases. The public set supports reproducible development and evaluation, while the private set retains unreleased cases to reduce direct tuning to benchmark tasks. Given the cost of long authoring runs, we select the public set to approximate full-suite performance at lower evaluation cost. The retained suite comprises an 83-case public set (16 Text-to-Scene, 22 indoor Image-to-Scene, and 45 outdoor Image-to-Scene) and an 87-case private set (10, 50, and 27, respectively). The sets contain distinct cases, although they can share source scenes. Across the ten agent configurations evaluated on both sets, public-set and full-suite overall scores correlate at \(r=0.996\), and no pair whose full-suite overall scores differ by at least 0.02 reverses order on the public set (Appendix~\ref{app:public-private}; Table~\ref{tab:public-private-validation}).

\subsection{Evaluation Pipeline}
\label{sec:evaluation-protocol}

\begin{figure}[!t]
\centering
\includegraphics[width=\linewidth,trim=0bp 5.5bp 0bp 1.5bp,clip]{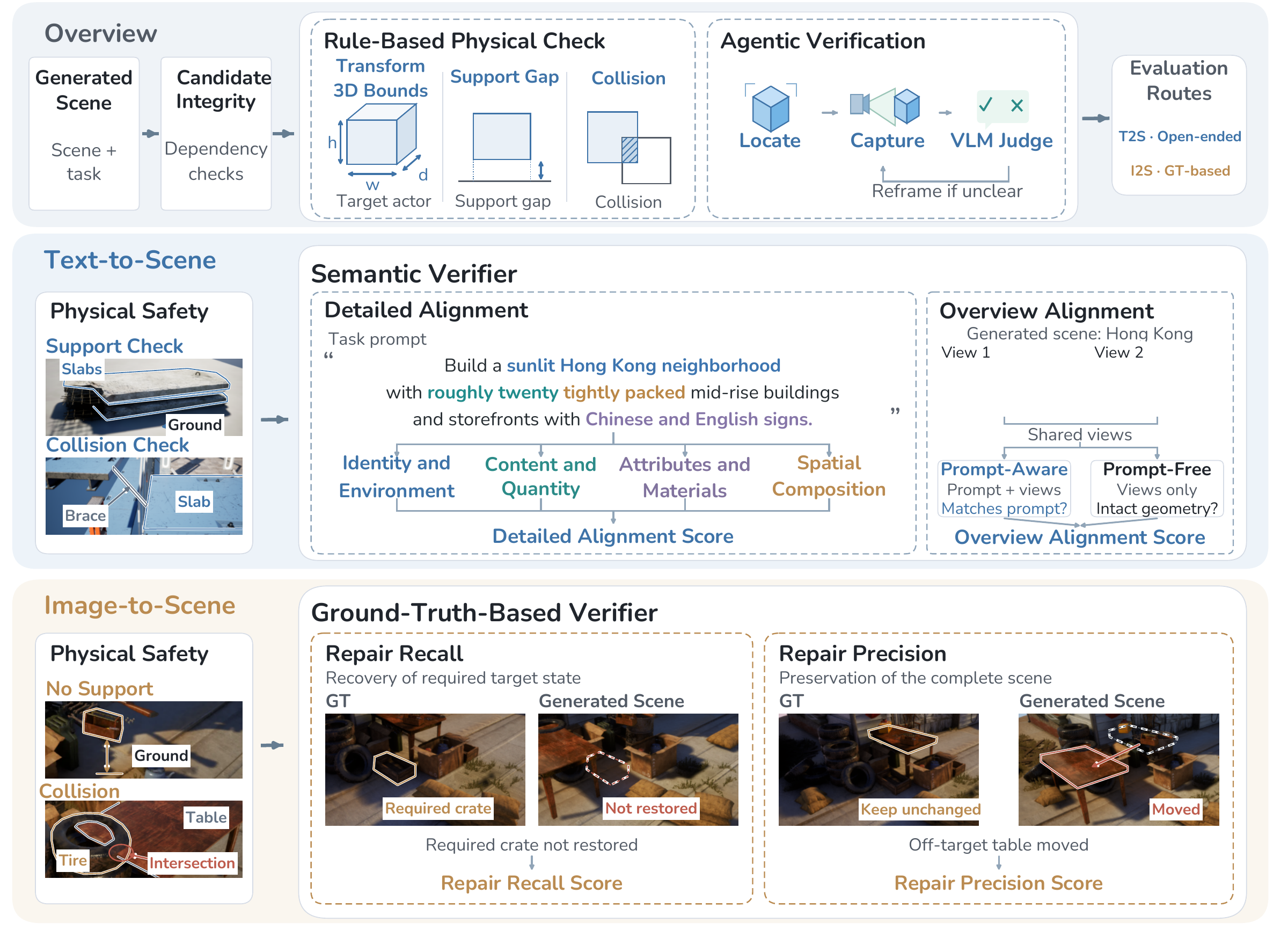}
\caption{Evaluation pipeline. Both settings share Candidate Integrity,
Physical Safety, and Agentic Verification. Text-to-Scene uses
semantic alignment, whereas Image-to-Scene measures target recovery and scene
preservation against withheld ground truth.}
\label{fig:evaluation-pipeline}
\end{figure}

\begin{figure}[tp]
\centering
\includegraphics[width=\linewidth]{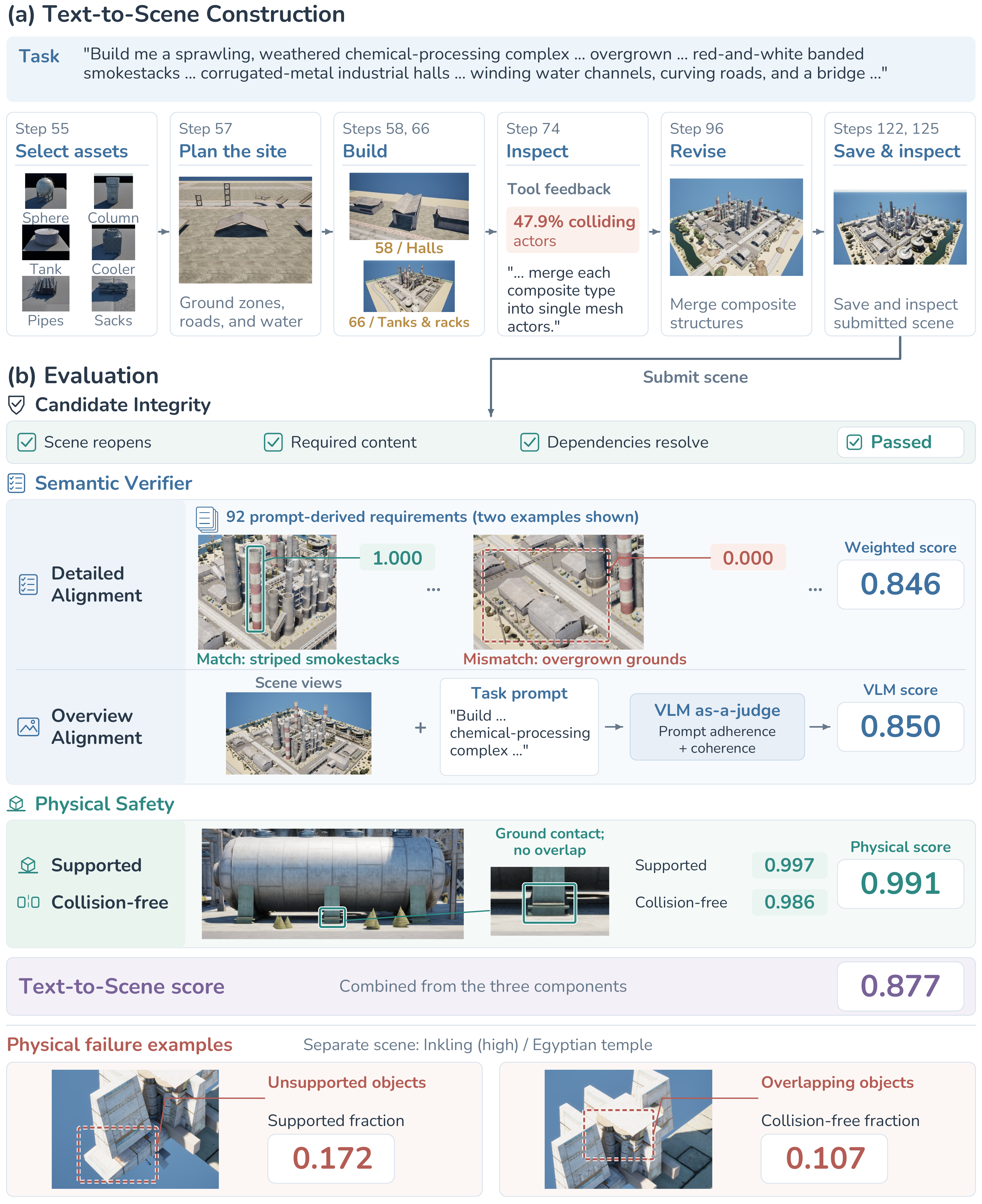}
\caption{Construction and evaluation of a chemical-plant scene generated by Claude Fable 5.1 (max). This historical demonstration case is excluded from the retained benchmark scores.
\textbf{(a) Scene construction.} Given an asset catalog and a task prompt, the coding agent uses code and tools to retrieve assets, plan the layout, and construct the scene. It repeatedly inspects and refines the scene before submission.
\textbf{(b) Evaluation.} The generated scene must pass Candidate Integrity checks before undergoing evaluation by the Semantic Verifier and Physical Safety checks. Selected evidence is shown beside the corresponding component scores, which contribute to the final Text-to-Scene score. The bottom inset contrasts physical failures in a separate scene generated by Inkling (high); its fractions summarize eligible evaluated actors, not only the highlighted examples.}
\label{fig:running-case}
\end{figure}

Code4Scene evaluates the generated scene submitted as a \texttt{.umap} file. After the run, the evaluator reopens this scene in a clean Unreal Engine process and checks its engine state and newly rendered views; the agent's code and interaction trajectory are recorded for analysis but are not scored (Figure~\ref{fig:running-case}). For each case, an agent configuration retries only if its first attempt produces no valid submission, with at most two attempts in total. We retain the first valid submission without score-based selection; cases without one score zero (Appendix~\ref{app:score-aggregation}). Reported token costs and tool counts cover only the retained attempt (Appendix~\ref{app:trajectory-extraction}). During execution, only case-provided resources are accessible. Evaluator-only references \(H\), e.g., the human-assembled scene, held-out views and ground-truth state, remain withheld.

\paragraph{Evaluation overview.}
Generated scenes in both settings must load correctly and satisfy physical constraints, so the pipeline applies shared checks before branching into the task-specific routes shown in Figure~\ref{fig:evaluation-pipeline} (top). \textit{Candidate Integrity} verifies that the generated Unreal Engine scene is a usable artifact, while the \textit{Rule-Based Physical Check} extracts deterministic scene-state measurements such as transforms, 3D bounds, support gaps, and collisions (Appendix~\ref{app:physical-safety}). Because small or occluded objects are hard to judge from a single global render, \textit{Agentic Verification} locates the relevant actors, captures clear views of them, and applies a VLM judge under fixed evaluator settings whenever visual interpretation is required (Appendix~\ref{app:judge-configuration}). Text-to-Scene then uses open-ended semantic verification against the prompt, whereas Image-to-Scene uses ground-truth-based verification against the withheld scene state. Separately, we evaluate \textit{Downstream Scene Usability} with a player agent and report the resulting Embodied Utility Score (EUS) as an auxiliary metric (Section~\ref{sec:evaluation}).

\paragraph{Text-to-Scene evaluation.}
Text-to-Scene follows an open-ended evaluation route because a prompt can admit
multiple valid constructions. Physical Safety evaluates support and collision
for the generated actors using the shared rule-based physical checks. Semantic
alignment is evaluated by the Semantic Verifier at two complementary levels.
Detailed Alignment checks the atomic requirements extracted from the prompt,
organized into four families: Identity and Environment, Content and Quantity,
Attributes and Materials, and Spatial Composition. Overview Alignment evaluates
the generated scene holistically from shared overview views. Its prompt-aware
component assesses whether the scene matches the requested content and overall
intent, while its prompt-free component checks whether the scene's geometry is structurally intact.
Together, Detailed Alignment and Overview Alignment measure prompt fulfillment
at both the requirement and scene levels.

\paragraph{Image-to-Scene evaluation.}
Image-to-Scene follows a ground-truth-based evaluation route because each
corruption has a precisely defined target state. Physical Safety evaluates the
edited actors using support and collision checks analogous to those in Text-to-Scene.
The Ground-Truth-Based Verifier then compares the generated scene with the
withheld ground-truth state. Repair Recall measures recovery of the required
target state, while Repair Precision measures preservation of the complete
scene by accounting for both incorrect target states and unintended changes to
content that should remain unchanged. These two metrics characterize whether
the agent performs the required repair while preserving the rest of the scene.

\subsection{Evaluation Criteria and Metrics}
\label{sec:evaluation}
\label{sec:official-scoring}

For a valid submission, the case scores are
\begin{align}
S_{\mathrm{T2S}} &= 0.2D+0.6O+0.2P,\\
S_{\mathrm{I2S}} &= 0.8F_1+0.2P,\qquad
F_1=\frac{2\mathrm{TP}}{2\mathrm{TP}+\mathrm{FP}+\mathrm{FN}},
\end{align}
where $D$ and $O$ are Detailed and Overview Alignment, $P$ is Physical Safety,
and TP, FP and FN count correctly repaired targets, incorrect or unintended
edits, and unrecovered targets. Physical Safety equally weights support safety
and collision-free validity; unavailable required components receive zero at
their fixed weights. The displayed T2S Semantic score is $0.25D+0.75O$.
Each setting is averaged over all scheduled retained cases, including zero
scores for invalid or missing submissions. Public I2S therefore gives equal
weight to its 67 cases (22 Indoor and 45 Outdoor), rather than equal weight to
the two domains. The overall score is
\[
S_{\mathrm{model}}=\tfrac12\overline{S}_{\mathrm{T2S}}
                   +\tfrac12\overline{S}_{\mathrm{I2S}}.
\]

Code4Scene evaluates coding agents that directly author engine-native 3D scenes, so success requires more than a plausible rendering. We therefore assess three properties of the generated scene: artifact integrity, task fulfillment, and static physical validity. \textit{Candidate Integrity} serves as a validity gate, checking that the generated scene can be reopened, contains the required scene content, and resolves its dependencies (Appendix~\ref{app:submission-integrity}). Task fulfillment uses setting-specific criteria, because construction and editing require different notions of correctness, while physical validity uses shared checks.

\paragraph{Text-to-Scene.}
Text-to-Scene evaluates whether an agent can realize an open-ended description as a coherent 3D scene. Prompts combine requirements on content, appearance, and spatial organization (e.g., \texttt{"houses lining a canal"} or \texttt{"a dense town center surrounded by open farmland"}). Because many different scenes can satisfy the same prompt, evaluation measures semantic and spatial adherence to the prompt rather than similarity to a single reference. \textit{Detailed Alignment} checks individual requirements using the strongest available evidence: engine state can resolve some properties, such as object presence and counts, but most, including appearance, style, and many spatial relations, require visual interpretation (e.g., \texttt{"a weathered facade"} or \texttt{"a Victorian-style building"}). A VLM therefore judges them from targeted rendered views. \textit{Overview Alignment} complements these local checks by evaluating the scene as a whole, including global semantic fit, composition, style, and structural coherence. This captures cases where individual elements are present but do not collectively realize the intended scene. Together, these metrics assess individual requirements and the overall scene (Appendix~\ref{app:semantic-verifier}).

\paragraph{Image-to-Scene.}
Image-to-Scene tests precise control of scene state: the agent must infer a repair from visual evidence and apply it exactly in 3D. The required edit may affect only a small part of a large scene (e.g., \textit{restore the missing globe}), while surrounding content should remain unchanged. Overall image or geometry similarity can therefore remain high even when the repair is missed or unrelated objects are modified. We instead compare the generated scene with the withheld ground-truth state. \textit{Repair Recall} measures how many required changes are recovered, while \textit{Repair Precision} measures whether the resulting states are correct and unrelated scene content is preserved. Their harmonic mean, \textit{Repair F1}, captures both repair completeness and control over the edit scope without being dominated by the many background actors that require no change (Appendix~\ref{app:gt-verifier}).

\paragraph{Shared criteria.}
Task correctness alone does not ensure physical usability. An object may satisfy the semantic requirement while remaining physically invalid (e.g., \texttt{a chair floating above the floor} or \texttt{a cabinet intersecting a wall}). \textit{Physical Safety} therefore measures static physical validity, namely support and interpenetration, directly from Unreal Engine state, applied to generated actors in Text-to-Scene and edited actors in Image-to-Scene (Appendix~\ref{app:physical-safety}). Because static checks cannot show whether a scene supports actual use, we additionally report the \textit{Embodied Utility Score (EUS)} as a separate downstream measure. A fixed player agent navigates predefined routes in the generated scene, and EUS combines object reachability and route completion \citep{yang2024physcene,anderson2018evaluation}. EUS is reported independently of the benchmark score because it evaluates subsequent embodied use of the generated scene (Appendix~\ref{app:eus-details}).

\section{Experiments}
\label{sec:experiments}

We use Code4Scene to measure how reliably current coding agents construct and edit 3D scenes (Section~\ref{sec:main-results}), to validate its evaluator (Section~\ref{sec:evaluator-validation}), and to locate where agents fail (Section~\ref{sec:analysis}).

\subsection{Experimental Setup}
\label{sec:experimental-setup}

The public T2S comparison incorporates sandboxed reruns for 12 systems on the
16 retained prompts. Opus 5 and Inkling retain their earlier T2S results;
I2S and private tasks retain the earlier generated scenes. Repair scores include
the material-export artifact correction described in
Appendix~\ref{app:trajectory-repair-outcomes}. Client and reasoning settings
vary across these cohorts (Appendix~\ref{app:rerun-configurations}); overall
scores combine the stated track-specific configurations. Token costs, total
tool counts and conditional spatial diagnostics use these same selected runs.
The separate Python-operation and execution-feedback profiles retain their
explicitly labelled historical cohort.

We evaluate coding agents powered by commercialized and open-weight models under a common benchmark interface. Our evaluation includes 14 coding
agent configurations on the public set and ten on the private set (Table~\ref{tab:main-results}). Across agent configurations, we keep the benchmark environment fixed: agents receive the same case-specific asset catalog, Unreal Engine tools, skill instructions, and evaluation protocol. This design allows us to compare coding-agent systems as they are practically deployed, while controlling the task environment and evaluator.

Unreal Engine workers run on eight NVIDIA RTX A5000 GPUs, while model inference is served through native clients, OpenRouter, or self-hosted endpoints (Appendix~\ref{app:agent-interface}). Task-performance analyses use the public set unless otherwise specified. Auxiliary evaluator-validation and navigation studies use separate samples, detailed in Appendices~\ref{app:evaluator-validation-results} and~\ref{app:eus-results}.

% Queue the leaderboard with its experimental context, not in the metrics section.
\begin{table*}[!t]
\centering
\compacttable
\caption{Main results on the retained public and private cases. Scores lie in $[0,1]$ ($\uparrow$). Each setting averages all scheduled cases, including failures; $S_{\mathrm{model}}$ weights the settings equally. Public T2S incorporates reruns for 12 systems; I2S and private tasks use the earlier generation cohort, with the exporter-artifact correction applied to Repair F1. Track-specific configurations are listed in Appendix~\ref{app:rerun-configurations}; overall values are composites of these settings. Dark/light highlights mark the best/second-best displayed values.}
\label{tab:main-results}
\small
\setlength{\tabcolsep}{3pt}
\setlength{\fboxsep}{3pt}
\noindent\colorbox{sbTableHead}{\parbox{\dimexpr\textwidth-2\fboxsep}{\textbf{A. Public set (83 cases, 14 systems)}}}\par\vspace{2pt}
\begin{tabularx}{\textwidth}{>{\raggedright\arraybackslash}Xrr>{\columncolor{sbBlue!4}}rrr>{\columncolor{sbOrange!4}}r>{\columncolor{sbInk!4}}r}
\toprule
\rowcolor{sbTableHead}
\textbf{Agent configuration} & \multicolumn{3}{c}{\cellcolor{sbBlue!12}\textbf{Text-to-Scene} (16 cases)} & \multicolumn{3}{c}{\cellcolor{sbOrange!14}\textbf{Image-to-Scene} (67 cases)} & \textbf{Overall} \\
\rowcolor{sbTableSubhead}
& Semantic $\uparrow$ & Physical $\uparrow$ & $S_{\mathrm{T2S}}\uparrow$ & Repair F1 $\uparrow$ & Physical $\uparrow$ & $S_{\mathrm{I2S}}\uparrow$ & $S_{\mathrm{model}}\uparrow$ \\
\midrule
GPT-6 Astra (max) & \tbest[sbBlue]{0.797} & 0.738 & \tsecond[sbBlue]{0.785} & \tsecond[sbOrange]{0.462} & 0.802 & \tsecond[sbOrange]{0.530} & \tbest[sbInk]{0.658} \\
Gemini 3.8 Flash (high) & 0.615 & \tsecond[sbBlue]{0.918} & 0.675 & \tbest[sbOrange]{0.562} & \tsecond[sbOrange]{0.816} & \tbest[sbOrange]{0.613} & \tsecond[sbInk]{0.644} \\
Claude Fable 5.1 (max) & \tsecond[sbBlue]{0.749} & \tbest[sbBlue]{0.980} & \tbest[sbBlue]{0.796} & 0.339 & 0.786 & 0.428 & 0.612 \\
Claude Opus 5 (max) & 0.714 & 0.855 & 0.742 & 0.388 & 0.790 & 0.469 & 0.606 \\
GPT-5.6 Sol & 0.723 & 0.872 & 0.753 & 0.314 & 0.681 & 0.387 & 0.570 \\
GLM-5.3 Flash (max) & 0.633 & 0.678 & 0.642 & 0.200 & \tbest[sbOrange]{0.830} & 0.326 & 0.484 \\
Muse Spark 1.3 & 0.529 & 0.570 & 0.537 & 0.239 & 0.784 & 0.348 & 0.443 \\
Grok 4.6 & 0.677 & 0.701 & 0.682 & 0.049 & 0.736 & 0.187 & 0.434 \\
Qwen 3.8 27B (thinking on) & 0.489 & 0.738 & 0.538 & 0.132 & 0.664 & 0.238 & 0.388 \\
Qwen 3.8 27B (thinking off) & 0.452 & 0.705 & 0.503 & 0.088 & 0.670 & 0.204 & 0.353 \\
Gemma 4 31B (thinking off) & 0.474 & 0.760 & 0.531 & 0.001 & 0.579 & 0.117 & 0.324 \\
Inkling (high) & 0.369 & 0.711 & 0.438 & 0.102 & 0.622 & 0.206 & 0.322 \\
Gemma 4 31B (thinking on) & 0.423 & 0.731 & 0.485 & 0.004 & 0.622 & 0.127 & 0.306 \\
DeepSeek V4.1 Flash & 0.334 & 0.424 & 0.352 & 0.079 & 0.776 & 0.219 & 0.285 \\
\bottomrule
\end{tabularx}
\par\vspace{7pt}
\noindent\colorbox{sbTableHead}{\parbox{\dimexpr\textwidth-2\fboxsep}{\textbf{B. Private set (87 cases, 10 configurations)}}}\par\vspace{2pt}
\begin{tabularx}{\textwidth}{>{\raggedright\arraybackslash}Xrr>{\columncolor{sbBlue!4}}rrr>{\columncolor{sbOrange!4}}r>{\columncolor{sbInk!4}}r}
\toprule
\rowcolor{sbTableHead}
\textbf{Agent configuration} & \multicolumn{3}{c}{\cellcolor{sbBlue!12}\textbf{Text-to-Scene} (10 cases)} & \multicolumn{3}{c}{\cellcolor{sbOrange!14}\textbf{Image-to-Scene} (77 cases)} & \textbf{Overall} \\
\rowcolor{sbTableSubhead}
& Semantic $\uparrow$ & Physical $\uparrow$ & $S_{\mathrm{T2S}}\uparrow$ & Repair F1 $\uparrow$ & Physical $\uparrow$ & $S_{\mathrm{I2S}}\uparrow$ & $S_{\mathrm{model}}\uparrow$ \\
\midrule
Gemini 3.8 Flash (high) & \tsecond[sbBlue]{0.708} & 0.608 & \tsecond[sbBlue]{0.688} & \tbest[sbOrange]{0.588} & \tsecond[sbOrange]{0.846} & \tbest[sbOrange]{0.639} & \tbest[sbInk]{0.664} \\
GPT-5.6 Sol (high) & \tbest[sbBlue]{0.724} & \tbest[sbBlue]{0.696} & \tbest[sbBlue]{0.719} & \tsecond[sbOrange]{0.371} & 0.685 & \tsecond[sbOrange]{0.433} & \tsecond[sbInk]{0.576} \\
Muse Spark 1.3 (medium) & 0.582 & 0.560 & 0.577 & 0.277 & 0.834 & 0.388 & 0.483 \\
GLM-5.3 Flash (max) & 0.579 & 0.543 & 0.572 & 0.259 & \tbest[sbOrange]{0.871} & 0.381 & 0.476 \\
Qwen 3.8 27B (thinking off) & 0.537 & 0.667 & 0.563 & 0.064 & 0.675 & 0.186 & 0.374 \\
Qwen 3.8 27B (thinking on) & 0.486 & 0.674 & 0.523 & 0.100 & 0.674 & 0.215 & 0.369 \\
Gemma 4 31B (thinking on) & 0.474 & 0.651 & 0.510 & 0.004 & 0.586 & 0.120 & 0.315 \\
Inkling (high) & 0.403 & 0.611 & 0.444 & 0.032 & 0.566 & 0.139 & 0.291 \\
Gemma 4 31B (thinking off) & 0.401 & \tsecond[sbBlue]{0.682} & 0.457 & 0.000 & 0.555 & 0.111 & 0.284 \\
DeepSeek V4.1 Flash (high) & 0.172 & 0.364 & 0.211 & 0.081 & 0.784 & 0.222 & 0.216 \\
\bottomrule
\end{tabularx}
\end{table*}

\subsection{Overall Performance}
\label{sec:main-results}

Construction and editing performance are strongly correlated but not interchangeable (Table~\ref{tab:main-results}): across the 14 public-set configurations,
Text-to-Scene and Image-to-Scene scores have a Spearman correlation of
\(\rho=0.67\), and different systems lead the two settings, with Claude Fable 5.1 leading construction (\(0.796\)) and Gemini 3.8 Flash leading editing (\(0.613\)).
GPT-6 Astra has the highest observed overall mean, \(0.6577\), followed by Gemini at \(0.6443\). Their paired difference is \(0.0134\), with a 95\% source-cluster interval of \([-0.0402,0.0665]\); this interval does not resolve an overall advantage (Appendix~\ref{app:trajectory-coverage}).
Astra performs better on construction (\(0.785\) vs.\ \(0.675\)), while
Gemini performs better on editing (\(0.613\) vs.\ \(0.530\)).
Precise editing remains challenging, with the best Repair F1 reaching
only \(0.562\).

\begin{takeaway}[label=take:construction-editing]
Construction and editing probe different capabilities: scene-level spatial reasoning and precise control of scene state. Astra scores higher than Gemini on construction, while Gemini scores higher on editing. Their overall difference remains unresolved by the paired source-cluster interval (Table~\ref{tab:main-results}).
\end{takeaway}

% The full-page running case precedes evaluator validation.
% Reserve room for the heading and wrapped chart without boxing the prose.
\Needspace{16\baselineskip}
\subsection{Evaluator Validation}
\label{sec:evaluator-validation}

% Keep the compact chart alongside its evaluator-validation discussion.
\begin{wrapfigure}[13]{r}{0.49\textwidth}
    \vspace{-8pt}
    \centering
        \newcommand{\evalfigfont}{\fontfamily{Nunito-TLF}\selectfont}
        % Explicit sizes match the type of the other main-text figures.
        \newcommand{\evalfiglabel}{\fontsize{9}{10}\selectfont}
        \newcommand{\evalfignum}{\fontsize{8}{9}\selectfont}
        {\evalfigfont\evalfignum\setlength{\tabcolsep}{3pt}
        \begin{tabular}{@{}ccc@{}}
            \textcolor{sbBlue}{$\blacksquare$}\ Individual
            & \textcolor{sbBlue!55}{$\blacksquare$}\ PL consensus
            & \textcolor{sbOrange}{$\blacksquare$}\ Kendall's $\tau_b$
        \end{tabular}}\par

        \vspace{2pt}
        \begin{tikzpicture}[x=3.80cm,y=0.30cm,
            every node/.style={font=\evalfigfont\evalfiglabel}]
            % Shared scale for agreement proportions and rank correlation.
            \foreach \x/\tick in {0/0,0.25/.25,0.5/.50,0.75/.75,1/1.0} {
                \draw[sbInk!12,line width=0.35pt] (\x,0.10) -- (\x,8.85);
                \node[anchor=north,font=\evalfigfont\evalfignum] at (\x,-0.04) {\tick};
            }
            \draw[sbInk!55,line width=0.45pt] (0,0.10) -- (1,0.10);

            \newcommand{\scorebar}[3]{%
                \fill[#3,rounded corners=0.5pt]
                    (0,{#1-0.30}) rectangle (#2,{#1+0.30});
                \node[anchor=west,font=\evalfigfont\evalfignum] at ({#2+0.014},#1)
                    {\pgfmathprintnumber[fixed,precision=3]{#2}};
            }

            \node[anchor=east,align=right,text width=1.65cm,font=\evalfigfont\evalfiglabel\bfseries]
                at (-0.045,7.50) {Adopted};
            \scorebar{8.35}{0.7937}{sbBlue}
            \scorebar{7.50}{0.8690}{sbBlue!55}
            \scorebar{6.65}{0.7380}{sbOrange}

            \node[anchor=east,align=right,text width=1.65cm]
                at (-0.045,4.50) {Individual fit};
            \scorebar{5.35}{0.7619}{sbBlue}
            \scorebar{4.50}{0.8214}{sbBlue!55}
            \scorebar{3.65}{0.6430}{sbOrange}

            \node[anchor=east,align=right,text width=1.65cm]
                at (-0.045,1.50) {PL fit};
            \scorebar{2.35}{0.7460}{sbBlue}
            \scorebar{1.50}{0.7976}{sbBlue!55}
            \scorebar{0.65}{0.5950}{sbOrange}

            \node[anchor=north,font=\evalfigfont\evalfiglabel] at (0.5,-1.12)
                {Metric value ($\uparrow$)};
        \end{tikzpicture}
        \vspace{-10pt}
        \caption{Text-to-Scene evaluator agreement on the retrospective
        human-study holdout.}
        \label{fig:evaluator-validation}
    \vspace{-8pt}
\end{wrapfigure}
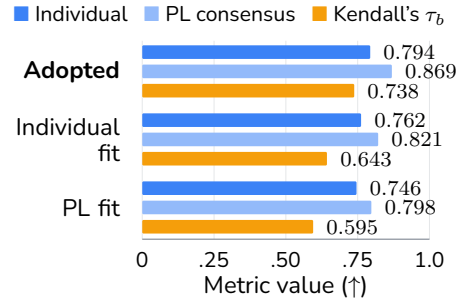

\paragraph{Agreement with human judgments.}

Because Text-to-Scene scores rely on VLM judgments, we test whether they agree with human preferences on a dedicated prompt sample. On its retrospective holdout, the adopted policy agrees with
86.90\% of Plackett-Luce consensus pairs and achieves Kendall's $\tau_b=0.738$ (Figure~\ref{fig:evaluator-validation}). Individual-preference agreement is
79.37\%, versus 80.95\% between annotators on the same prompts.
Earlier human-guided exploration used the full study sample, so this is internal
validation, not a test on unseen prompts; it establishes neither above-human
performance nor Image-to-Scene alignment
(Appendix~\ref{app:evaluator-validation-results}).

\paragraph{VLM judge repeatability.}
We evaluate the repeatability of the VLM judge by rerunning Detailed Alignment scoring three times on six hand-selected Text-to-Scene cases while holding the scenes, evidence, prior decisions, and request schedules fixed. Across repeated evaluations, pairwise Kendall's $\tau_b$ ranges from $0.889$ to $0.944$, while the standard deviation of agent-configuration means ranges from $0.0010$ to $0.0132$ (Appendix~\ref{app:judge-repeatability}). These results characterize scoring stability under fixed inputs.

\subsection{Analysis}
\label{sec:analysis}

We examine where agents fall short in spatial reasoning and in controlling scene state.

\begin{figure*}[t]
\centering
\includegraphics[width=\textwidth,trim=0bp 3bp 0bp 5bp,clip]{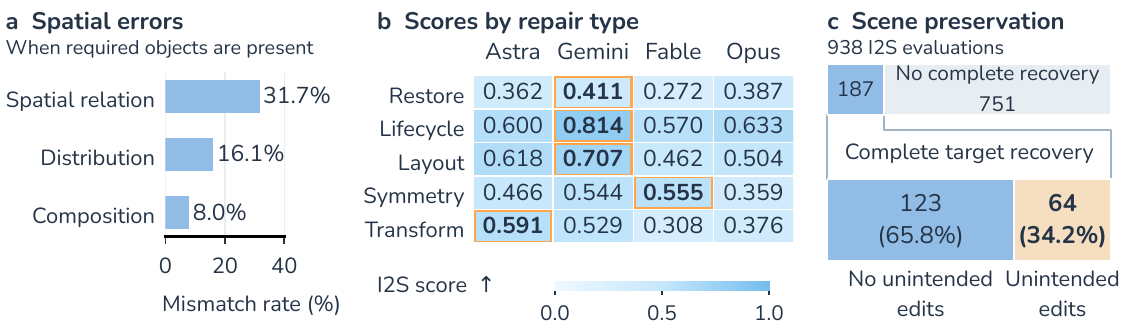}
\caption{
(a) Current-run spatial errors with required objects present: 74/387 direct mismatches across the analyzable evidence; coverage is described in Appendix~\ref{app:conditional-spatial-analysis}. (b) Mean I2S score by repair type for four agents; orange marks the best of the four. (c) Complete recoveries with or without unintended edits.}
\label{fig:analysis-diagnostics}
\end{figure*}

\paragraph{From scene content to spatial composition.}
\label{sec:reconstruction-analysis}
Frontier agents' strengths in realizing scene content coexist with errors in
spatial organization. Astra leads the Identity, Content and Attributes
families, yet Spatial Composition is the lowest-scoring family for all 14
public-set configurations. Even Astra, which has the highest Spatial mean,
reaches only $0.587$ (Appendix~\ref{app:task-category-analysis}). In the selected current runs, the difficulty
persists when content is present: 74 of 387 analyzable spatial requirements whose
relevant entities pass separate existence checks still receive direct
mismatch judgments (19.1\%; 95\% source-cluster interval, 12.1--26.6\%; Figure~\ref{fig:analysis-diagnostics}a). The diagnostic covers 219 of 224 model--case pairs; two invalid candidates and three unavailable semantic reports contribute no diagnostic evidence, while retaining their official score treatment.
These failures show a gap between generating the requested elements and satisfying the relations that organize them
(Appendix~\ref{app:conditional-spatial-analysis}).

\begin{takeaway}[label=take:spatial-composition]
Spatial Composition remains the weakest requirement family for all 14 agents. The current-run diagnostic also finds errors when the required objects are present (Figure~\ref{fig:analysis-diagnostics}a).
\end{takeaway}

\paragraph{Distinct strengths in visual scene repair.}
\label{sec:repair-category-analysis}
The repair types behind the overall editing score (Appendix~\ref{app:case-construction-details}) reveal different frontier
strengths (Figure~\ref{fig:analysis-diagnostics}b). Astra leads Transform
with a category score of $0.591$, ahead of Gemini's $0.529$, whereas Gemini leads Lifecycle and Layout; on Layout, Gemini scores $0.707$ versus Astra's $0.618$. Fable narrowly leads Symmetry ($0.555$ vs.\ Gemini's $0.544$ on eight cases; Appendix~\ref{app:task-category-analysis}). Thus, Gemini's overall editing lead does not extend to every repair type, and a single editing score hides which kinds of scene-state changes each agent controls well.

\begin{takeaway}[label=take:repair-types]
Editing ability varies substantially across repair types. Astra performs best on Transform repairs, while Gemini is stronger on Lifecycle and Layout operations (Figure~\ref{fig:analysis-diagnostics}b).
\end{takeaway}

\paragraph{Failure to preserve the scene after target recovery.}
\label{sec:failure-analysis}
Completing the required repair does not always keep changes within the
intended scope. On Indoor editing tasks, Astra fully recovers all
targets in 20 cases and Gemini in 17. Requiring complete recovery with no
unintended changes reverses this ordering: Astra achieves 15 such repairs and Gemini 16 (Appendix~\ref{app:trajectory-repair-outcomes}). All five of Astra's complete Indoor recoveries with false positives include changes to
non-target content. Across all 14 configurations, 64 of 187 complete recoveries (34.2\%) still contain unintended additions or changes
(Figure~\ref{fig:analysis-diagnostics}c). Recovering the desired target state
therefore does not by itself demonstrate control over the rest of the scene.
Error types and sensitivity checks are reported in
Appendix~\ref{app:trajectory-execution}.

\begin{takeaway}[label=take:recovery-preservation]
Recovering the target does not guarantee precise editing. Across all 14 configurations, 34.2\% of complete recoveries still contain unintended changes to the surrounding scene (Figure~\ref{fig:analysis-diagnostics}c).
\end{takeaway}

% Temporarily use the complete appendix related work instead of the short version.
% \input{sections/02_related_work}
% Conclusion is included at the end of the appendix below.

% In the preprint, disclosures are integrated into the relevant appendices
% and acknowledgments instead of repeated before the references.

% Times-compatible glyphs keep author accents intact in PDF text extraction.
\begingroup
\fontencoding{T1}\fontfamily{qtm}\selectfont
\bibliography{references}
\endgroup
\bibliographystyle{iclr2027_conference}

\clearpage
\appendix
% Active only after \appendix; matches code/analysis/appendix_palette.py.
\definecolor{sbInk}{HTML}{243447}
\definecolor{sbBlue}{HTML}{45A5E5}
\definecolor{sbOrange}{HTML}{FFA64D}
\definecolor{sbPanel}{HTML}{EFF8FF}
\definecolor{sbTableHead}{HTML}{E6F3FD}
\definecolor{sbTableSubhead}{HTML}{F4FAFF}
\colorlet{sbGreen}{sbBlue}
\colorlet{sbPurple}{sbBlue}
\colorlet{sbRed}{sbOrange}

% Keep appendix floats at the top, including pages containing only floats.
\makeatletter
\setlength{\@fptop}{0pt}
\setlength{\@fpsep}{6pt plus 1pt minus 1pt}
\setlength{\@fpbot}{0pt plus 1fil}
\makeatother
\addtocontents{toc}{\protect\setcounter{tocdepth}{2}}
\begingroup
\renewcommand{\contentsname}{Appendix Contents}
\small
\tableofcontents
\endgroup
\clearpage
\section{Benchmark Details}
\label{app:benchmark-details}

This appendix documents task setup and agent execution; evaluator settings and
analysis protocols appear in Appendices~\ref{app:evaluation-protocol}
and~\ref{app:trajectory-methods}. We will release the public-set task code,
evaluation and analysis scripts, agent configuration records, the shared skill
specification and tool schemas, and judge instructions. Access to third-party
scene assets will remain subject to their original licenses.

% Internal reproducibility checklist, not an assertion of existing availability.
% Statistics: package frozen case-level data, cohort manifests, aggregation
% policies, and commands that regenerate the reported tables and figures.
% Evaluation: record submitted scenes and asset requirements, Unreal/verifier
% versions, judge instructions/settings, matching tolerances, and missing-value
% handling alongside the evaluation scripts.
% Agent execution: record task inputs, prompts/skills, tool schemas, model-client
% mappings, software versions, reasoning/sampling settings, seeds, budgets,
% hardware, and commands. Preserve closed-service and stochastic limitations.
% Release preparation: verify access conditions and licenses, distinguish public
% and private task availability, and avoid identifying links in the review draft.

\subsection{Coding-Agent Interface}
\label{app:agent-interface}

All evaluated systems interact with Unreal Engine through the same tool
interface and receive the same skill specification describing how to use that
interface. Systems differ in the coding-agent client and the model behind it;
the model-facing names in the results are therefore shorthand for complete
agent configurations rather than estimates of model-only effects.

\paragraph{Tool interface.}
The 25 tools in Table~\ref{tab:agent-tools} support asset discovery, scene
editing, state inspection, Python execution, planning, verification, and
persistence. Asset discovery provides semantic search over the case-specific
catalog, metadata including physical dimensions and spawn paths, and previews
where available. Agents can
edit actors directly or use the optional deterministic planner to validate
scene graphs, solve transforms, inspect plans, and build scenes in batches.
Editor Python supports additional operations and asynchronous jobs. Inspection
provides screenshots, scene measurements, and targeted geometric checks.
The benchmark proxy blocks tools outside this interface.

\paragraph{Skill text.}
All 170 retained cases provide the same skill specification in the agent's
initial prompt. It documents the available tools and their usage contracts, with general operating rules (for example, calling \textrm{setup\_environment} first and last) and short code snippets, but no scene-specific construction plan. The agent must decide how to search, plan, construct, inspect, and revise
the scene. The specification has four parts, covering asset retrieval,
editor Python execution, scene-graph planning and scene building.

\paragraph{Execution environment.}
We use Unreal Engine 5.8.0 on an Ubuntu 24.04.3 LTS server with eight NVIDIA
RTX A5000 GPUs (24\,GB each) and two AMD EPYC 7513 CPUs. These specifications
describe the Unreal execution host; model inference uses the services
described below.

\paragraph{Coding-agent clients in the earlier generation cohort.}
GPT-5.6 Sol (high)~\citep{openai2026gpt56sol} uses Codex, Claude Fable 5.1 (max)~\citep{anthropic2026fable51} and Claude Opus 5 (max)~\citep{anthropic2026opus5} use Claude Code, and Gemini 3.8 Flash (high)~\citep{google2026gemini38flash} uses Antigravity. Inkling (high)~\citep{thinkingmachines2026inkling} uses Hermes for Text-to-Scene and indoor Image-to-Scene and Claude Code for
outdoor Image-to-Scene, both through OpenRouter. Muse Spark 1.3 (medium)~\citep{meta2026musespark13}, GLM-5.3 Flash (max)~\citep{zai2026glm53flash}, Grok 4.6 (high)~\citep{spacexai2026grok46}, and DeepSeek V4.1 Flash (high)~\citep{deepseek2026v41flash} use the benchmark's API-based agent loop with OpenRouter. Qwen 3.8 27B~\citep{qwen2026qwen3827b} and Gemma 4 31B~\citep{gemmateam2026gemma4} use the same loop
with self-hosted OpenAI-compatible endpoints for both thinking settings.
GPT-6 Astra (max)~\citep{openai2026gpt6astra} includes both Codex and API-based OpenRouter runs, with the
client assignment recorded per case. All clients access tools through a shared
metering proxy. Client versions are recorded where available and may vary across runs.

\paragraph{Public T2S rerun configurations.}
\label{app:rerun-configurations}
The November public T2S refresh reruns 12 systems; Opus 5 and Inkling retain
their earlier results. Sol uses Codex with \texttt{gpt-5.6-sol@max}; Muse uses
\texttt{meta/muse-spark-1.3-contributor@max}; DeepSeek uses
\texttt{deepseek/deepseek-v4.1-flash@max+deepinfra}. Grok uses its native
client with \texttt{grok-4.6@xhigh}, recorded as max effort by the run index,
rather than the earlier OpenRouter loop. Astra uses Codex in these T2S reruns.
The remaining rerun model labels retain their reported reasoning settings.
These client-specific settings are not a common compute scale. I2S and private
tasks retain the earlier configurations above. Accordingly, the refreshed
public and combined tables omit a single effort label for Sol, Muse, DeepSeek
and Grok; their overall values are composites across the documented settings,
not measurements of one fixed configuration across every track and split.

% Resolved 2026-09-24: document the model-to-client mapping, inference routes,
% Unreal version, and execution hardware above, based on retained run records.
% \papertodo{Add the model-to-client mapping and the Unreal version.}
% Internal release-packaging reminder, not a remaining manuscript TODO.
% \papertodo{Complete the supplementary manifest linking each agent configuration
% to its per-run client settings and versions, harness revision, skill text and
% hashes, the 25 tool schemas, judge instructions, and recorded limits.}

\paragraph{Resource limits.}
Text-to-Scene runs allow 800 tool calls and 180 minutes. Image-to-Scene runs allow 120 tool calls and 45 minutes for Indoor cases and 80 tool calls and 20 minutes for Outdoor cases.

\paragraph{Retries and submission selection.}
An agent configuration receives a second attempt only when the first produces
no valid submission. The first valid submission is retained, with no
score-based comparison between attempts. If neither attempt produces a valid
submission, the case receives zero. Reported token costs and tool counts
include only the retained attempt, not discarded attempts, and therefore do
not represent cumulative expenditure across retries.

\begin{table}[!t]
\centering
\compacttable
\caption{The 25 shared agent tools. Arguments marked $^*$ are required;
all others are optional.}
\label{tab:agent-tools}
\scriptsize
\setlength{\tabcolsep}{3pt}
\begin{tabularx}{\linewidth}{>{\raggedright\arraybackslash}p{0.13\linewidth}>{\raggedright\arraybackslash}p{0.42\linewidth}X}
\toprule
\rowcolor{sbTableHead}
\textbf{Group} & \textbf{Tool (arguments)} & \textbf{Result} \\
\midrule
Asset discovery & \textrm{search\_assets}(query$^*$, category, k) & Ranked matching assets \\
 & \textrm{get\_categories}() & Asset categories with counts and descriptions \\
 & \textrm{get\_asset\_metadata}(ids$^*$) & Category, description, tags, dimensions in metres, spawn path \\
 & \textrm{get\_asset\_preview}(id$^*$) & Rendered preview image, if available \\
\midrule
Scene editing & \textrm{spawn\_actor}(name$^*$, static\_mesh$^*$, location$^*$, rotation, scale) & Spawns a static-mesh actor \\
 & \textrm{spawn\_blueprint\_actor}(actor\_name$^*$, blueprint\_id$^*$, location$^*$, rotation, scale) & Spawns a Blueprint actor \\
 & \textrm{set\_actor\_transform}(name$^*$, location, rotation, scale) & Updates an actor's transform \\
 & \textrm{delete\_actor}(name$^*$) & Deletes one actor \\
 & \textrm{delete\_all\_spawned}() & Deletes all actors spawned in the run \\
\midrule
State inspection & \textrm{get\_actors\_in\_level}() & All actors in the scene \\
 & \textrm{find\_actors\_by\_name}(pattern$^*$) & Actors whose names match the pattern \\
\midrule
Programmatic control & \textrm{execute\_python\_script}(script$^*$, execution\_mode) & Starts an asynchronous editor Python job; returns its job ID and log path \\
 & \textrm{read\_job\_log}(job\_id, log\_path, max\_bytes, wait\_s) & Job status and output \\
\midrule
Planning & \textrm{validate\_scene\_graph}(graph$^*$) & Validity report, including entries that would be dropped \\
 & \textrm{solve\_scene\_graph}(graph$^*$, mode, half\_extent\_m) & Deterministic world transforms with de-overlap \\
 & \textrm{render\_plan}(solved) & Top-down image of the solved plan \\
 & \textrm{tier\_derivation}(solved) & Build tier for each placed entry \\
 & \textrm{ground\_pass}(scene\_text$^*$, half\_extent\_m) & Spawns ground tiles across the build area \\
 & \textrm{build\_tier}(tier, entries) & Spawns one solved tier or explicit entries in batches \\
\midrule
Verification & \textrm{take\_screenshot}(filename, filepath) & Viewport screenshot \\
 & \textrm{measure\_scene}() & Actor count, collision and floating rates, offending actors, extent \\
 & \textrm{check\_floating}(actor\_names, threshold\_cm) & Actors whose gap to the surface below exceeds the threshold \\
 & \textrm{check\_collisions}(actor\_names, touch\_tolerance\_cm) & Overlapping actor pairs \\
\midrule
Environment and persistence & \textrm{setup\_environment}(ground\_size, time\_of\_day) & Adds missing lighting, sky, fog and ground \\
 & \textrm{save\_scene\_as}(name$^*$, revert\_to\_original) & Saves the scene as a new \textrm{.umap} \\
\bottomrule
\end{tabularx}
\end{table}

\subsection{Case Construction}
\label{app:case-construction-details}

The source collection spans historical and fantasy settings, contemporary
urban and transit scenes, industrial and working sites, domestic and office
interiors, and stylized environments. This collection emphasizes the
compositional irregularity, asset heterogeneity, and authored intent of
human-assembled environments, at the cost of the scalability offered by
procedural generation (Appendix~\ref{app:environment-generation}).
Text-to-Scene cases use 130\,m or
200\,m square plates. The human-assembled scenes used for Image-to-Scene contain
approximately 260 to 6{,}700 actors outdoors and 95 to 3{,}168 actors indoors.
Each case specifies its source scene and required asset packages.
The Image-to-Scene evidence contains two reference views per indoor
case and one per outdoor case; both use the same editing formulation.

Image-to-Scene corruptions must be stable actor-level changes inferable from
the reference evidence. Sources are excluded when procedural hierarchies do not
expose stable targets, occlusion prevents reliable reference views, required
content cannot be packaged reproducibly, or the scene does not support a clear
actor-level edit.

\paragraph{Authored operation groups.}
Image-to-Scene cases retain the authors' operation-category assignments.
The short labels in the analysis correspond to Repair/Completion (Restore),
Object Lifecycle (Lifecycle), Layout/Rearrangement (Layout),
Symmetry/Pattern (Symmetry), and Geometric Transformation (Transform).
In that order, the full suite contains 18, 18, 9, 9, and 18 indoor cases
and 22, 14, 17, 7, and 12 outdoor cases. These groups describe authored tasks;
they are not labels inferred from agent scores. The selected-case mapping is
retained with the Figure~\ref{fig:i2s-operation-heatmap} data
(Appendix~\ref{app:task-category-analysis}).
% \papertodo{Document the authors' category-assignment criteria, especially the
% boundary between Repair/Completion and Object Lifecycle, with representative
% task IDs.}

In the original authored 80-case Outdoor suite, label-chain lengths one
through five occurred in 25, 3, 26, 25, and 1 cases, respectively. Label chain length counts top-level
authored operation entries, not atomic transformations or agent
tool calls, because one entry may encode multiple atomic changes.

\paragraph{Difficulty labels.}
The initial difficulty prior considers the number and spatial extent of
affected targets, the number and coupling of atomic operations, the relational
constraints to recover, and ambiguity in the reference evidence. Results use
the finalized difficulty assignments rather than this initial prior.
Table~\ref{tab:difficulty-results} reports full-suite scores under the finalized
Easy/Medium/Hard assignments; tier counts are given in
Appendix~\ref{app:difficulty-results}.

\paragraph{Indoor answer keys.}
Indoor answer keys provide the state of the human-assembled scene together with an
author-provided aligned mapping between intended input and ground-truth actor
states. They also include an author-provided task-to-operation grouping used
for the descriptive operation-category analysis in
Section~\ref{sec:repair-category-analysis}. Unlike outdoor records, indoor
category labels and actor-state mappings do not specify atomic edit sequences
or chain lengths.

\subsection{Public and Private Sets}
\label{app:public-private}

The retained benchmark contains 170 cases: 26 Text-to-Scene and 144
Image-to-Scene cases, divided into an 83-case public set and an 87-case private set.
Task code will be released for the public set but not for the private set.
The public set contains 16 Text-to-Scene, 22 Indoor and 45 Outdoor cases;
the private set contains 10, 50 and 27. All 14 configurations are evaluated
on the public set; ten are also evaluated on the private set.
Table~\ref{tab:main-results} reports these cohorts separately. Unless marked
otherwise, analyses use the retained public cases. Full-suite scores combine
the retained sets for the ten configurations with all 170 cases.

\paragraph{Selection.}
The original 95-case public set was selected so that its plain per-setting mean, without case
weights or calibration, estimates the corresponding full-suite mean. Within
each setting, greedy forward selection added the case that most reduced the
largest estimation error over the available agent configurations and score columns,
with each column's error normalized by the variation of random subsets of the
same size. The selection pool contained six agent configurations evaluated on every
setting, namely GPT-5.6 Sol (high), Gemini 3.8 Flash (high),
Qwen 3.8 27B (thinking off), Qwen 3.8 27B (thinking on),
Gemma 4 31B (thinking off) and Gemma 4 31B (thinking on), plus
Gemini 3.7 Flash (high) for Text-to-Scene only. Selection used only case scores, not difficulty labels,
content packages or categories. One selected Text-to-Scene case was later
withdrawn from the benchmark and replaced by another Text-to-Scene case. The candidate sets are nested (61, 75, 95 and 128 cases),
before the asset-use exclusions described in Section~\ref{sec:benchmark-suite}. The current 83-case set removes the 12 retired public cases without rerunning selection or transferring their scores to replacements.

\paragraph{Historical selection-time validation and size.}
Table~\ref{tab:public-set-size} reports historical leave-one-model-out error
before asset-use filtering; these values do not describe the current 83-case set. Each agent configuration is removed in turn, the cases
are reselected from the remaining agent configurations, and the removed
agent configuration's full-suite mean is predicted by its mean on the selected cases.
The 95-case set has mean absolute error $0.011$ on setting totals and $0.017$
on their components, with no reversed pair of totals. Random subsets of the same size had $1.3$--$1.7$ times the error of greedy
selection. Greedy selection also had lower
error than stratification by content package and category on both repair
settings, and tied it on Text-to-Scene. Anchor-point selection by
k-medoids \citep{vivek2024anchor} was competitive on Indoor but requires
per-case weights, so uniform weights were retained. Keeping only cases with
mean score between 0.3 and 0.7 \citep{ndzomga2026efficient} was two to four
times worse than greedy selection and worse than random sampling. That filter
targets model ranking on pass/fail tasks; discarding easy and hard cases biases continuous-score means.

For the size, we use two external reference points: an average estimation
error near $0.02$, as reported by tinyBenchmarks with 100 curated examples
\citep{polo2024tinybenchmarks}, and the 44--70\% task reduction that preserves
agent rankings in \citet{ndzomga2026efficient}. The 61-case set was optimized
for totals only, and its component error exceeds $0.02$. The 128-case set
reduces the suite by less than 44\%. The 75- and 95-case sets meet both reference
points on average. The additional 20 cases of the 95-case set are Outdoor
cases; they reduce the Outdoor total error from $0.019$ to $0.010$ and the
Outdoor Physical Safety error, the largest Outdoor column error at 75 cases,
from $0.029$ to $0.011$. Both sets use the same Indoor cases; their Physical Safety and repair errors
($0.022$ and $0.029$) still exceed $0.02$.

\begin{table}[!t]
\centering
\compacttable
\caption{Historical candidate public-set sizes before asset-use filtering. Errors are leave-one-model-out mean absolute errors of
full-suite means on the $[0,1]$ scale, averaged over setting totals or their
components. Reversed pairs count agent configuration pairs whose full-suite scores
differ by at least $0.02$ but are ordered oppositely, over totals and
components.}
\label{tab:public-set-size}
\small
\setlength{\tabcolsep}{4pt}
\begin{tabular}{rcrccc}
\toprule
\rowcolor{sbTableHead}
Cases & \shortstack{T2S/Indoor/\\Outdoor} & Kept & \shortstack{Total\\error} &
\shortstack{Component\\error} & \shortstack{Reversed\\pairs} \\
\midrule
61  & 16/25/20 & 32.1\% & 0.016 & 0.025 & 8 \\
75  & 20/25/30 & 39.5\% & 0.014 & 0.019 & 1 \\
\textbf{95} & 20/25/50 & 50.0\% & 0.011 & 0.017 & 1 \\
128 & 28/50/50 & 67.4\% & 0.007 & 0.011 & 0 \\
\bottomrule
\end{tabular}
\end{table}

\paragraph{Validation under current scoring.}
These are descriptive comparisons of the published track-specific composites. Public T2S includes reruns while private T2S retains the earlier cohort; configuration differences can therefore contribute to the observed public--full gaps.

We compare the retained 83-case public set and 170-case full suite for the ten
configurations evaluated on both (Table~\ref{tab:public-private-validation}).
The overall model score differs by $0.011$ on average and at most $0.026$.
No overall ranking pair with a full-suite gap of at least $0.02$ reverses.
At that gap threshold, two T2S pairs (Qwen thinking on and Muse; Qwen thinking off
and Gemma thinking off), one Outdoor pair (Muse and DeepSeek) and one I2S pair
(Qwen thinking off and Inkling) reverse. The public I2S mean weights Indoor
by $22/67$ and Outdoor by $45/67$; its absolute difference from the full-suite
I2S mean is $0.016$ on average and at most $0.036$.
This validation uses the filtered cases and does not reuse the historical
selection-time error estimates.

\begin{table}[!t]
\centering
\compacttable
\caption{Public-set versus full-suite scores under current scoring for the ten
agent configurations evaluated on both. $\Delta$ is the public-set score minus the
full-suite score. Kendall's $\tau$ compares rankings. Reversed pairs have full-suite gaps
of at least $0.02$ but the opposite public-set order.}
\label{tab:public-private-validation}
\small
\setlength{\tabcolsep}{4pt}
\begin{tabular}{lccccc}
\toprule
\rowcolor{sbTableHead}
Score & Mean $|\Delta|$ & Max $|\Delta|$ & Mean $\Delta$ & Kendall $\tau$ &
\shortstack{Reversed\\pairs} \\
\midrule
$S_{\mathrm{T2S}}$ & 0.018 & 0.054 & $+0.007$ & 0.911 & 2 \\
Indoor & 0.036 & 0.103 & $+0.030$ & 0.956 & 0 \\
Outdoor & 0.013 & 0.022 & $+0.003$ & 0.911 & 1 \\
$S_{\mathrm{I2S}}$ & 0.016 & 0.036 & $-0.003$ & 0.956 & 1 \\
$S_{\mathrm{model}}$ & 0.011 & 0.026 & $+0.002$ & 0.911 & 0 \\
\bottomrule
\end{tabular}
\end{table}

\paragraph{Use of public-set scores.}
We assess public-set representativeness through the score differences above
\citep{miller2024adding}. These validation errors describe the ten shared
configurations; scores for the four public-only configurations refer to the
retained 83 cases. The public set approximates setting means but is not
proportional across content packages, categories or difficulty tiers; its
category and tier results describe the public cases rather than estimating
full-suite values. Sub-metrics outside the selection objective, such as
individual Physical Safety leaves, deviated by up to $0.056$ at selection time.
Estimates may fail far outside the selection pool's score range \citep{zhang2025benchmark}.

\section{Running Cases}
\label{app:running-cases}

\definecolor{rcPanel}{HTML}{EFF8FF}
\definecolor{rcTitle}{HTML}{243447}
\definecolor{rcGood}{HTML}{45A5E5}
\definecolor{rcBad}{HTML}{FFA64D}
\definecolor{rcPanelI}{HTML}{FFF5E9}
\definecolor{rcTitleI}{HTML}{243447}
\definecolor{rcTruth}{HTML}{A8D8F5}
\newcommand{\rerunviews}[1]{%
  \includegraphics[width=.247\linewidth]{figures/running_cases/rerun_20261109/#1_0.png}\hfill%
  \includegraphics[width=.247\linewidth]{figures/running_cases/rerun_20261109/#1_1.png}\hfill%
  \includegraphics[width=.247\linewidth]{figures/running_cases/rerun_20261109/#1_2.png}\hfill%
  \includegraphics[width=.247\linewidth]{figures/running_cases/rerun_20261109/#1_3.png}}

\newtcolorbox{rcprompt}[1]{enhanced, colback=rcPanel, colframe=rcPanel,
  arc=2pt, boxrule=0pt, left=4pt, right=4pt, top=3pt, bottom=3pt,
  fontupper=\scriptsize, before upper={\textcolor{rcTitle}{\textbf{#1}}\quad}}
\newtcolorbox{rctask}[1]{enhanced, colback=rcPanelI, colframe=rcPanelI,
  arc=2pt, boxrule=0pt, left=4pt, right=4pt, top=3pt, bottom=3pt,
  fontupper=\scriptsize, before upper={\textcolor{rcTitleI}{\textbf{#1}}\quad}}
\newcommand{\rcrun}[4]{%
  \begin{tcolorbox}[enhanced, colback=white, colframe=#1, boxrule=0.6pt,
    before skip=3pt, after skip=3pt,
    arc=2pt, left=1pt, right=1pt, top=1pt, bottom=1pt, toptitle=1pt,
    bottomtitle=1pt, colbacktitle=#1, coltitle=sbInk, fonttitle=\scriptsize,
    title={\textbf{#2}\hfill #3}]
  \IfFileExists{#4}{\includegraphics[width=\linewidth]{#4}}{\rerunviews{#4}}
  \end{tcolorbox}}

\begin{figure}[!tp]
\centering
\begin{rcprompt}{Prompt: Nordic Harbour}
Build me a compact, sunlit Copenhagen-style canal district with narrow
cobblestone streets and rectangular waterways enclosing dense urban blocks.
Arrange roughly thirty attached four- to six-storey townhouses in continuous
street and canal-front rows, using slender fa\c{c}ades, repetitive white-framed
windows, arched doors and carriage entrances, modest cornices, and steep tiled
roofs packed with dormers and brick chimneys. Include occasional exposed party
walls between staggered building rows. Finish the fa\c{c}ades in weathered muted
red, mustard yellow, dusty blue, warm grey, beige, and exposed brick, with red,
orange, and charcoal roof tiles. Run a principal waterfront street along a
stone quay, crossed by smaller streets and linked across the canals by short,
gently arched stone-and-metal bridges, including one opening onto a broad paved
promenade with rounded landings. Form the canal edges from stepped pale stone,
dark timber retaining walls, and iron railings around reflective rippling
water. Add regularly spaced leafy trees, ornate black lamps, stone bollards,
benches, European road signs, zebra crossings, puddles, grime, and sparse
weeds.\par
\textit{The text prompt also contains the ground instruction and the 79-asset catalog, omitted here.}
\end{rcprompt}
\rcrun{rcGood}{Higher-scoring: Claude Fable 5.1 (max), rank 4 of 14}
  {$S_{\mathrm{T2S}}$ 0.804 \enspace$|$\enspace Detailed 0.623 \enspace$|$\enspace Overview 0.798 \enspace$|$\enspace Physics 1.000}
  {nordic_harbour_fable}
\rcrun{rcBad}{Lower-scoring: Inkling (high), rank 11 of 14}
  {$S_{\mathrm{T2S}}$ 0.396 \enspace$|$\enspace Detailed 0.021 \enspace$|$\enspace Overview 0.373 \enspace$|$\enspace Physics 0.841}
  {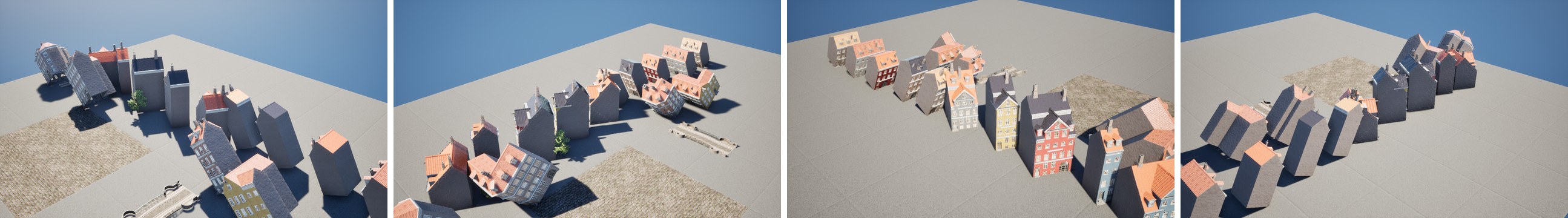}
\caption{Text-to-Scene case: Nordic Harbour. Fable's rerun contains 2{,}678 exported actors,
arranging townhouses around rectangular canals with quays and bridges.
Inkling places 24 actors, largely tilted townhouses on a bare plate without
visible canals or quays.}
\label{fig:running-nordic-harbour}
\par\vspace{2pt}
\begin{rcprompt}{Prompt: Medieval Big Farm Town}
Build me a sprawling medieval farm town set in a rugged, enclosed mountain
basin, with a dense fortified settlement occupying the rocky northern half and
broad, open farmland spreading southward. Ring the scene with massive irregular
gray boulders, steep mossy outcrops, scattered grassy plateaus, and narrow
turquoise streams that descend through gaps in the rocks, including several
small white waterfalls. Place roughly sixty to eighty compact timber-and-stone
buildings inside and around a rectangular walled town: mostly one- and
two-story houses with dark wooden framing, pale infill or masonry walls, and
steep orange-brown tiled roofs. Arrange them tightly and organically along
narrow dirt lanes, small courtyards, canals, and bridges, with taller square
gate towers and watchtowers punctuating the long timber-and-stone walls.
Outside the walls, transition to a much sparser patchwork of fenced fields,
orchards, livestock enclosures, winding dirt tracks, isolated cottages, sheds,
tents, carts, stacked materials, and scattered rocks. Distribute several tall
wooden windmills across the fields, some attached to heavy stone or timber
tower bases, and place additional lookout buildings on remote rocky peaks and
grassy ledges. Let streams skirt both sides of the settlement, feed small
turquoise pools within the town, and pass beneath simple wooden footbridges.
Use uneven grassy ground with worn pale paths, low fences, shrubs, tall grass,
and thin, irregularly spaced trees, keeping the town center crowded and layered
while the agricultural outskirts remain open, loosely ordered, and visibly
bounded by cliffs and boulder formations.\par
\textit{The text prompt also contains the ground instruction and the 158-asset catalog, omitted here.}
\end{rcprompt}
\rcrun{rcGood}{Higher-scoring: GPT-6 Astra (max), rank 3 of 14}
  {$S_{\mathrm{T2S}}$ 0.795 \enspace$|$\enspace Detailed 0.761 \enspace$|$\enspace Overview 0.910 \enspace$|$\enspace Physics 0.483}
  {farm_town_astra}
\rcrun{rcBad}{Lower-scoring: DeepSeek V4.1 Flash (max), rank 6 of 14}
  {$S_{\mathrm{T2S}}$ 0.682 \enspace$|$\enspace Detailed 0.514 \enspace$|$\enspace Overview 0.665 \enspace$|$\enspace Physics 0.902}
  {farm_town_deepseek}
\caption{Text-to-Scene case: Medieval Big Farm Town. Astra's rerun contains 2{,}704 exported actors, with a walled settlement and open fields outside it. DeepSeek's rerun contains 1{,}137 exported actors, with houses and windmills spread across the basin and less separation between settlement and farmland.}
\label{fig:running-farm-town}
\end{figure}

Figures~\ref{fig:running-nordic-harbour} and~\ref{fig:running-farm-town}
show two Text-to-Scene cases. Each figure gives the prompt's scene description
and the four overview views for a higher-scoring and a lower-scoring run,
with scores and per-case ranks among the 14 agent configurations.

\begin{figure}[!tp]
\centering
\begin{rctask}{Task: Indoor Atomic v1 054 (Detective Office)}
% Task summary uses the paper's scene terminology; raw task data is unchanged.
Repair the current indoor scene so it matches the two provided reference images
of the human-assembled scene. Make only the minimum changes needed, and preserve all
unrelated actors and properties.\par
\textit{Corruption, not shown to the agent: the world globe's mesh is replaced with a fan.}
\end{rctask}
\rcrun{rcTruth}{Ground truth, withheld from the agent}{}
  {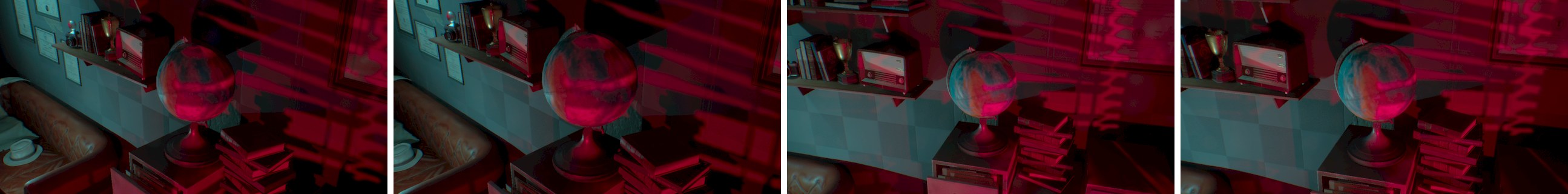}
\rcrun{rcGood}{Successful: Claude Opus 5 (max), tied 1st of 14}
  {$S_{\mathrm{I2S}}$ 1.000\enspace$|$\enspace F1 1.000 (TP/FP/FN 1/0/0)\enspace$|$\enspace Physics 1.000}
  {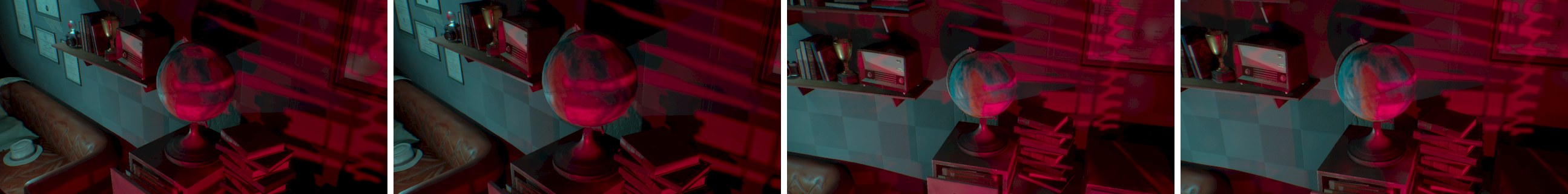}
\rcrun{rcBad}{Unsuccessful: Gemma 4 31B (thinking off), tied last of 14}
  {$S_{\mathrm{I2S}}$ 0.100\enspace$|$\enspace F1 0.000 (TP/FP/FN 0/1{,}310/1)\enspace$|$\enspace Physics 0.500}
  {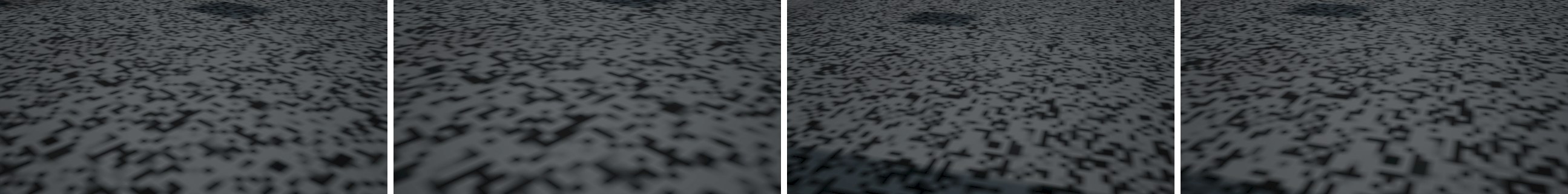}
\caption{Image-to-Scene running case: Indoor Atomic v1 054. The successful
run restores the world globe and changes nothing else. The unsuccessful run does not restore the globe and deletes nearly the whole room, keeping 31 of 1{,}325 actors. Rows are brightened identically.}
\label{fig:running-indoor-054}
\end{figure}

\begin{figure}[!tp]
\centering
\begin{rctask}{Task: Comp 31 Old Industrial Pallet Bay Repair (Old Industrial Courtyard)}
Repair the current outdoor scene so it matches the provided reference image.
Make only the minimum changes needed, and preserve all unrelated actors and
properties.\par
\textit{Corruption, not shown to the agent: a duplicate of the stacked pallets
is added $2.8$\,m away, and the original stack is moved by $2$\,m and rotated
by $35^\circ$.}
\end{rctask}
\rcrun{rcTruth}{Ground truth, withheld from the agent}{}
  {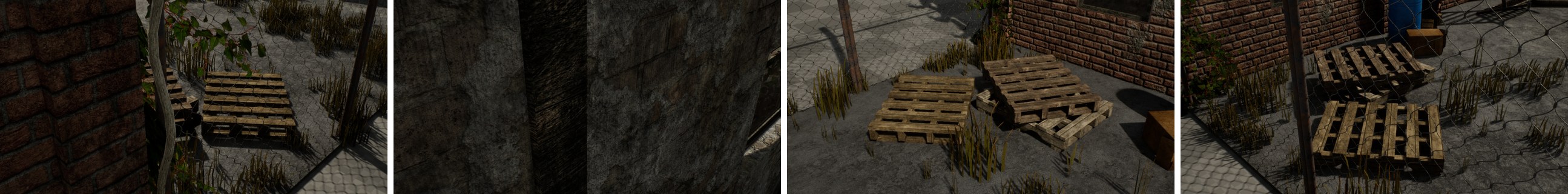}
\rcrun{rcGood}{Successful: GPT-6 Astra (max), rank 1 of 14}
  {$S_{\mathrm{I2S}}$ 1.000\enspace$|$\enspace F1 1.000 (TP/FP/FN 2/0/0)\enspace$|$\enspace Physics 1.000}
  {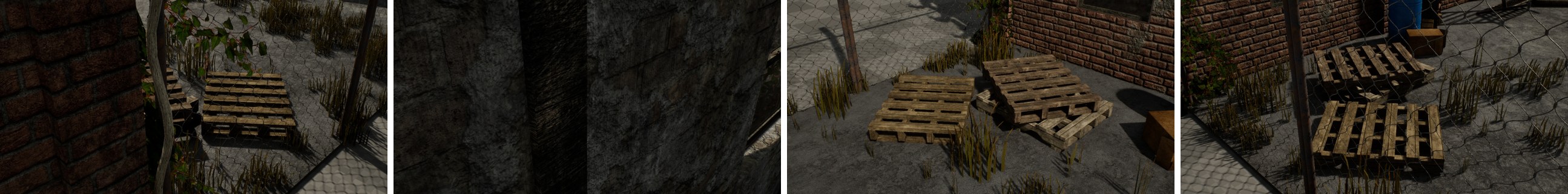}
\rcrun{rcBad}{Unsuccessful: DeepSeek V4.1 Flash (high), tied 12th of 14}
  {$S_{\mathrm{I2S}}$ 0.100\enspace$|$\enspace F1 0.000 (TP/FP/FN 0/5/2)\enspace$|$\enspace Physics 0.500}
  {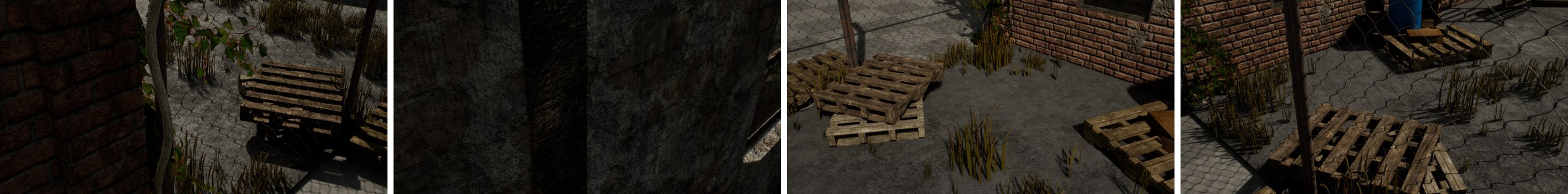}
\caption{Image-to-Scene running case: Comp 31 Old Industrial Pallet Bay Repair.
The successful run removes the duplicate and returns the original stack, with
no other change. The unsuccessful run keeps the duplicate, restores neither stack, and makes four unintended edits.}
\label{fig:running-comp-31}
\end{figure}

Figures~\ref{fig:running-indoor-054} and~\ref{fig:running-comp-31} show two
Image-to-Scene cases. Each figure summarizes the task instruction, then shows
the withheld ground truth and the successful and unsuccessful submissions, all rendered from
the evaluator's four repair-target cameras, which are identical across rows.

% Detailed implementation and validation protocols for the evaluated benchmark.
\section{Evaluation Protocol}
\label{app:evaluation-protocol}

This appendix defines the evidence and scoring protocols shared by the public
and private sets. Validation and navigation results appear in
Appendices~\ref{app:evaluator-validation-results} and~\ref{app:eus-results}.

\subsection{Evaluation Outputs and Verifier Routing}

The evaluator reports component scores, supporting measurements, applicability,
and evaluation status. A \textsc{Measured} result has a finite score in $[0,1]$;
it denotes a completed measurement, not a passed quality threshold.
\textsc{Not Applicable}, \textsc{Not Evaluated}, and \textsc{Error} have no
numeric measurement. \textsc{Valid} and \textsc{Invalid} describe submission
integrity. The conversion from component results to reported benchmark scores,
including zero-score rules, is defined in Appendix~\ref{app:score-aggregation}.

Every case undergoes Candidate Integrity and Physical Safety checks.
Text-to-Scene additionally uses Detailed Alignment and Overview Alignment;
Image-to-Scene uses actor-level Precision, Recall and Repair F1
(Appendix~\ref{app:gt-verifier}). Evaluation runs in an isolated environment,
and its reports are unavailable to agents during construction.

\paragraph{Judge configuration.}
\label{app:judge-configuration}
Every VLM judgment made during evaluation uses Qwen 3.8 27B
(\textrm{Qwen/Qwen3.8-27B}) with thinking disabled, temperature 0, seed 0,
structured output, and a 16{,}000-token response limit. This configuration covers Detailed Alignment requirement judgments, both Overview Alignment passes, and the Image-to-Scene Agentic Verification comparisons of paired target views, which are reported but not scored. Actor Repair F1 itself uses structured scene state.

Table~\ref{tab:verifier-matrix} summarizes the evaluation groups.
\begin{table*}[!hb]
\centering
\caption{Evaluation groups and their inputs. Ground truth is withheld from agents.}
\label{tab:verifier-matrix}
\small
\setlength{\tabcolsep}{6pt}
\begin{tabularx}{\textwidth}{
    >{\raggedright\arraybackslash}p{0.25\textwidth}
    >{\raggedright\arraybackslash}p{0.32\textwidth}
    >{\raggedright\arraybackslash}X}
\toprule
\rowcolor{sbTableHead}
\textbf{Group} & \textbf{Input} & \textbf{Criterion} \\
\midrule
\textbf{Candidate Integrity}\newline
{\footnotesize all cases}
& Generated scene, execution record, and dependencies
& Loadability, minimum content, scene-graph validity, dependency resolution, and asset-library manifest agreement \\
\addlinespace[1pt]
\rowcolor{sbTableSubhead}
\textbf{Physical Safety}\newline
{\footnotesize all eligible cases}
& Scene geometry and targeted engine measurements
& Supported actors and collision-free actors \\
\addlinespace[1pt]
\textbf{Semantic Verifier}\newline
{\footnotesize Text-to-Scene}
& Prompt, frozen requirements, scene state, and rendered evidence
& Detailed Alignment and Overview Alignment \\
\addlinespace[1pt]
\rowcolor{sbTableSubhead}
\textbf{Ground-Truth-Based Verifier}\newline
{\footnotesize Image-to-Scene}
& Input scene, generated scene, and withheld human-assembled scene
& Actor repair TP/FP/FN, precision, recall and F1 \\
\addlinespace[1pt]
\textbf{Downstream EUS}\newline
{\footnotesize reported separately}
& Navigation mesh, eligible targets, and fixed-walker rollouts
& Approachability and route completion \\
\bottomrule
\end{tabularx}
\end{table*}

\subsection{Candidate Integrity Checks}
\label{app:submission-integrity}

The evaluator independently exports the generated scene and applies three
integrity checks:
\begin{enumerate}[leftmargin=*,nosep,topsep=3pt]
\item \emph{Scene snapshot.} Validate the exported scene-graph schema and
check the task's minimum actor count.
\item \emph{Content and dependencies.} Verify that referenced packages
resolve under the frozen dependency policy.
\item \emph{Asset-library manifest.} Verify agreement among the submission,
scorer and frozen content snapshot.
\end{enumerate}

Ending at a resource limit does not itself invalidate a saved
\textrm{.umap}; the generated scene remains eligible for these checks.
Downstream verification requires all three checks to pass. A failed or
errored gate leaves downstream measurements unevaluated; reporting follows
Appendix~\ref{app:score-aggregation}.

\subsection{Physical Safety}
\label{app:physical-safety}

Physical Safety combines support and solid-penetration checks. For a nonempty
evaluated support population,
$r_{\mathrm{unsup}}=N_{\mathrm{unsup}}/N_{\mathrm{tested}}$ is the fraction
of actors for which the profile finds no support. This is a lower-is-better
defect rate; its safety counterpart is $1-r_{\mathrm{unsup}}$.
For $N_{\mathrm{eval}}>0$, the penetration score
$S_{\mathrm{pen}}=1-N_{\mathrm{viol}}/N_{\mathrm{eval}}$ is the collision-free
fraction of eligible actors with trusted verdicts.
For nonnegative weights
\(\mathbf w^P=(w^P_{\mathrm{supp}},w^P_{\mathrm{pen}})\) summing to one,
\[
S_{\mathrm{physics}}(\mathbf w^P)
=w^P_{\mathrm{supp}}(1-r_{\mathrm{unsup}})
 +w^P_{\mathrm{pen}}S_{\mathrm{pen}}.
\]
The adopted policy uses \(\mathbf w^P=(0.5,0.5)\) in both settings;
alternative fitted weights are used only in the Text-to-Scene human-alignment
comparison (Table~\ref{tab:human-alignment-weights}). We omit the weight
argument when referring to the adopted Physical Safety score.
Required checks without numeric measurements, including explicitly
inapplicable checks, contribute zero at their fixed weights. Case scores are
averaged equally, rather than pooling actor counts across cases.
An evaluated check with no eligible actors receives the protocol's neutral
score of one; this differs from an unavailable measurement. An empty-scope
score does not certify whole-scene safety; reports list each check's eligible population.

\paragraph{Support.}
Both task settings use a $5$ cm support-gap tolerance. Text-to-Scene evaluates
generated actors after excluding infrastructure and degenerate helpers.
Its geometric support test uses world-space axis-aligned bounding boxes
(AABBs): an actor is supported when its bottom is at most $5$ cm above the
world ground plane, or another actor lies beneath it with a vertical gap in $[0,5]$ cm, including flush contact, and that actor's footprint, seen from above, contains its pivot within a small margin.
Image-to-Scene evaluates edited actors using targeted Unreal support checks,
with a $5$ cm ground-gap tolerance and, when applicable, a $5$ cm lateral-support
tolerance. These are static support checks, not tests of dynamic stability.

\paragraph{Solid penetration.}
The scene-wide Text-to-Scene population and the edited-actor Image-to-Scene
population use deterministic scene-relative physics roles. Non-solid actors,
support surfaces, and oversized environment proxies are excluded as scored
targets. A trusted collision with an environment proxy can nevertheless count
as a containment failure for a target actor. AABB overlap is only a broad phase;
a violation requires confirmed Unreal collision-body overlap and native
\textrm{FHitResult} minimum-translation depth exceeding the actor's tolerance.
For Text-to-Scene, the tolerance is $5\%$ of the shortest full world-AABB span,
clamped to $[5,50]$ cm; Image-to-Scene uses a fixed $5$ cm tolerance.
These are actor-contact thresholds, not scene-level pass/fail thresholds.
Incomplete evidence can prove a violation but cannot establish a clean actor;
a leaf with unresolved evidence and no confirmed violation remains
\textsc{Not Evaluated}. Reports retain evaluated and unresolved actor counts,
roles, contact depths, and effective tolerances. A reported fraction must
therefore be interpreted together with its measurement coverage.

\subsection{Semantic Verifier for Text-to-Scene}
\label{app:semantic-verifier}

\paragraph{Detailed Alignment requirements.}
Each Text-to-Scene prompt is parsed once into a graph of requirements with
bindings to scene evidence. Validation checks prompt grounding, semantic
coverage, reference resolution, scope, and graph consistency. The requirements
are fixed before evaluation. They are generated automatically or with agent
assistance and are not human-reviewed.

\paragraph{Evidence ladder.}
Stage~1 uses the trusted generated-scene inventory, identity bindings, and
deterministic geometry for claims that can be resolved structurally. Stage~2
localizes the actors or collections needed by unresolved claims and captures
targeted RGB evidence. Stage~3 sends only unresolved visual claims to the
multimodal judge, with a bounded alternate view and a separate best-evidence
binary arbitration when valid images remain. Camera selection controls
observations but is not positive evidence. Missing targets or required
images yield \textsc{Not Evaluated}; transport, parsing, contract, or capture
failures yield \textsc{Error}. Reports retain all decisions and evidence.

\paragraph{Detailed Alignment aggregation.}
Each accepted atomic decision has score $s_j\in[0,1]$. Binary matches and
mismatches receive one and zero, deterministic checks may return intermediate
scores, and an unresolved applicable requirement receives zero while retaining
its weight,
\[
q_j=
\begin{cases}
s_j, & \text{for an accepted decision},\\[1mm]
0, & \text{for an unresolved requirement}.
\end{cases}
\]
Duplicate summaries already represented by scored descendants add no weight,
and a retained conjunction is capped by its nested requirements. Known
Coverage separately records the scoring weight for which the evidence supports
a definite decision.

Atomic predicates are aggregated first within prompt clauses and then within
four Detailed Alignment families. For family $f$, let \(\mathcal C_f\)
contain its nonempty clauses and \(\mathcal R_{fc}\) the active predicates in
clause $c$. For any requirement-level quantity $x$,
\[
\mathcal A_f(x)=\frac{1}{|\mathcal C_f|}
\sum_{c\in\mathcal C_f}\frac{1}{|\mathcal R_{fc}|}
\sum_{j\in\mathcal R_{fc}}x_j.
\]
Applying this operator to $q_j$ gives the family score $S_f$. With
nonnegative family weights \(\mathbf w^D\),
\begin{equation}
S_{\mathrm{detailed}}
=\frac{\sum_{f\in\mathcal F}w^D_fS_f}
       {\sum_{f\in\mathcal F}w^D_f},
\label{eq:detailed-family-score}
\end{equation}
where \(\mathcal F\) contains the families applicable to the prompt. An absent
family is removed and the remaining weights are renormalized; an unresolved
requirement inside an applicable family is not removed. The published
aggregate is a point score, accompanied by Known Coverage, family scores,
and atomic decisions. If \(k_j=1\) when evidence determines a definite decision and
\(k_j=0\) otherwise, then \(K_f=\mathcal A_f(k)\); aggregate Known Coverage
uses the same applicable-family weights as \(S_{\mathrm{detailed}}\). A
conjunction with a known-false child is determined.
Table~\ref{tab:human-alignment-weights} lists adopted and fitted weights.

\paragraph{Conditional Spatial Composition analysis.}
\label{app:conditional-spatial-analysis}
This diagnostic reads requirement graphs and direct judgments from the same selected reports as the current public T2S scores, including unchanged Opus and Inkling runs. It makes no new evaluator calls.

Figure~\ref{fig:analysis-diagnostics}a reports mismatch rates by predicate type within the Spatial Composition family: spatial relations between entities (e.g., along, between, inside), distributions of groups of entities (e.g., clusters or rows), and scene-level composition (e.g., overall density, openness, or a loose grid). A requirement is included only when every
relevant entity passes a separate affirmative
existence check without additional qualifiers. Missing, contradictory or
unresolved existence evidence is excluded. Across 219 analyzable model--case pairs out of 224 scheduled, these checks retain 387 of 3,933 Spatial Composition predicates (9.8\%), spanning 105 model--case pairs and nine source prompts. Two invalid candidates and three reports with unavailable semantic evidence are excluded from this diagnostic, not assigned successful judgments. The pooled direct mismatch rate is 74/387 (19.1\%; 95\% source-cluster interval, 12.1--26.6\%). Giving equal weight to clauses within each eligible case and then to cases yields 23.3\% (12.1--40.9\%); this is a different estimand from the pooled predicate rate. Collection presence does not establish exact cardinality or
member identity. The analysis uses direct
predicate judgments before parent constraints are applied, separating spatial
mismatches from inherited score reductions.

\paragraph{Overview Alignment.}
A scene-graph and clearance-aware camera plan produces exactly four
generated-scene-only RGB overview views. A Prompt-Aware multimodal call scores
global prompt alignment, composition and layout, style and atmosphere, and
completeness and polish. With nonnegative weights \(\mathbf w^O\) summing to
one, their combination is
\begin{equation}
S_{\mathrm{prompt}}
=w^O_{\mathrm{align}}S_{\mathrm{align}}
+w^O_{\mathrm{layout}}S_{\mathrm{layout}}
+w^O_{\mathrm{style}}S_{\mathrm{style}}
+w^O_{\mathrm{complete}}S_{\mathrm{complete}}.
\label{eq:overview-prompt-score}
\end{equation}
An independent Prompt-Free call uses the same views to score intrinsic
structural integrity, including stretching, shearing, tearing, collapse, and
incoherent fragmentation. It does not score missing requested content, style,
scene density, support, or inter-object collision. The resulting score is
\[
\widetilde S_{\mathrm{overview}}
=S_{\mathrm{prompt}}\left(0.75+0.25S_{\mathrm{struct}}\right).
\]
If the evaluator identifies a specific severe defect with confidence at least
$0.8$ and corroborating evidence from at least two views, then
\[
S_{\mathrm{overview}}
=\min\left(\widetilde S_{\mathrm{overview}},0.40\right);
\]
otherwise \(S_{\mathrm{overview}}=\widetilde S_{\mathrm{overview}}\). The
visible-scene summary is explanatory and is never used as a scoring
intermediate. Table~\ref{tab:human-alignment-weights} reports the adopted and
fitted Prompt-Aware weights.

\paragraph{Judging criteria and scoring rubrics.}
For Detailed Alignment, visual judgments use only evidence visible in rendered
RGB images and cite the views used. For an unresolved visual requirement,
the judge returns \textsc{Match}, \textsc{Mismatch}, or \textsc{Unknown}; insufficient visibility
of the subject, relation scope, or counted collection yields
\textsc{Unknown}. A broader or functionally related category cannot substitute
for a named object. Object identity is checked against the expected category
and its aliases, separately from count, position, material, and style.
When applied, final binary arbitration returns \textsc{Match} or
\textsc{Mismatch} and records residual uncertainty as confidence.

For Overview Alignment, Prompt-Aware scoring treats the task prompt as the
evaluation target and uses anchors 0 (absent or contradictory), 0.25 (weak),
0.5 (partial), 0.75 (mostly satisfied), and 1 (clearly complete) for the
dimensions in Eq.~\ref{eq:overview-prompt-score}. A requested element counts
if clearly visible in at least one view; capture quality and structural
integrity are excluded from this score. Prompt-Free scoring assesses intrinsic
geometry integrity using anchors 1 (no visible defect), 0.75 (minor localized
defects), 0.5 (clear defects in several structures or one major structure),
0.25 (widespread severe deformation), and 0 (unusable). Floating, unsupported,
or surreal layouts are assessed under prompt alignment or Physical Safety;
severe geometric defects require evidence from at least two views.

\paragraph{Semantic Verifier aggregation.}
Detailed Alignment and Overview Alignment form one Semantic Verifier. For
nonnegative final weights \(\alpha+\beta+\gamma=1\) with
\(\alpha+\beta>0\),
\begin{equation}
S_{\mathrm{semantic}}
=\frac{\alpha S_{\mathrm{detailed}}+
        \beta S_{\mathrm{overview}}}{\alpha+\beta}.
\label{eq:semantic-verifier-score}
\end{equation}
For a case with fixed evidence \(\mathbf x\), Physical Safety enters the
Text-to-Scene case score as
\[
\begin{aligned}
S_{\mathrm{T2S}}(\mathbf x;\theta)
&=(\alpha+\beta)S_{\mathrm{semantic}}+\gamma S_{\mathrm{physics}}\\
&=\alpha S_{\mathrm{detailed}}+\beta S_{\mathrm{overview}}
  +\gamma S_{\mathrm{physics}},
\end{aligned}
\]
where \(\mathbf x\) contains the frozen requirement decisions, Overview
Alignment and Physical Safety measurements, and applicability information, and
\(\theta=(\mathbf w^D,\mathbf w^O,\mathbf w^P,\alpha,\beta,\gamma)\).
The component scores on the right are computed from \(\mathbf x\) using their
respective weight blocks in \(\theta\); these arguments are suppressed for
brevity. Unavailable required components retain their fixed weights and receive
zero credit. Table~\ref{tab:human-alignment-weights}
reports the adopted and individual-fit values.

\subsection{Ground-Truth-Based Verifier: Actor Repair F1}
\label{app:gt-verifier}

Actor Repair F1 measures target recovery and penalizes incorrect outputs and
unintended changes to unrelated content. It uses independently exported input,
ground-truth and generated scenes, with each affected actor as one repair target.

\paragraph{Defining the repair targets.}
Comparing the input with the ground truth determines which actors must be
added, removed or restored. These targets are fixed before inspecting the
generated scene, using the authored input--ground-truth correspondence and shared
evidence fields. Let $n_+$ denote targets that should be present after repair
(additions and state restorations), and $n_-$ denote required removals.
An actor remains one target even when several of its properties need repair.
Every benchmark case has at least one target. Camera and scene-capture actors
are excluded, and changes to runtime identifiers, labels or role tags alone
are not semantic edits.

\paragraph{Determining repair success.}
A target that should be present is recovered only when a candidate actor has
the correct asset and class and passes every applicable state test in
Table~\ref{tab:repair-thresholds}. Asset and class agreement uses a normalized
structural descriptor comprising actor class, mesh asset or Blueprint class,
and component-asset signature. State tests cover position, rotation, signed
scale, and any bounds or attributes recorded in the ground truth. The whole
target state is checked, including properties that did not require repair.
A one-to-one matching maximizes the number of recovered targets; one candidate
actor cannot satisfy two targets, and partially restored targets receive no
partial credit. Removal requires semantic absence of the original actor, not just renaming.

\begin{table}[!t]
\centering
\caption{Nominal acceptance tests for desired-present repair targets.
All applicable tests must pass; positions are in cm. These benchmark
tolerances are not human-calibrated perceptual equivalences.}
\label{tab:repair-thresholds}
\small
\begin{tabularx}{\linewidth}{lX}
\toprule
Quantity & Acceptance test \\
\midrule
Asset and class & Exact normalized structural descriptor agreement \\
World position & $\|p_c-p_g\|_2\leq 5$ cm \\
World rotation & $2\arccos(\min(1,|q_c^\top q_g|))\leq 5^\circ$ \\
Signed scale & $\max_j|s_{c,j}/s_{g,j}-1|\leq 0.05$ \\
Bounds centre, if recorded in GT & $\|b_c-b_g\|_2\leq 5$ cm \\
Bounds extent, if recorded in GT & $\max_j|e_{c,j}/e_{g,j}-1|\leq 0.05$ \\
Attributes, if recorded in GT & Exact property-dictionary, normalized material-set and component/material-slot agreement \\
\bottomrule
\end{tabularx}
\end{table}

\paragraph{Counting correct and incorrect outcomes.}
Let $m$ be the number of successfully matched targets and $d$ the number of
successful required removals. To count incorrect outputs, define $C$ as the
candidate actors that are newly added, semantically changed, or surviving
targets that should be present. This includes a corrupted target left
unchanged. An unchanged surviving actor that should have been removed instead
counts as a missed removal. Let $o$ count unintended losses of background actors,
including actors removed or repurposed to satisfy a repair target; overlapping
losses are counted once. The outcome counts are
\begin{equation}
\begin{aligned}
\mathrm{TP}&=m+d,\\
\mathrm{FP}&=|C|-m+o,\\
\mathrm{FN}&=(n_+-m)+(n_--d).
\end{aligned}
\label{eq:repair-counts}
\end{equation}
True positives (TP) count completed targets, false negatives (FN) count missed
targets, and false positives (FP) count incorrect candidate outputs and
unintended background losses.
Every target contributes either a TP or an FN, so
$\mathrm{TP}+\mathrm{FN}=n_++n_-$. This also covers pure deletions.

\paragraph{Computing Repair F1.}
Precision measures the share of counted repair outcomes that are correct,
recall measures the fraction of targets recovered, and F1 combines them:
\begin{equation}
P=\frac{\mathrm{TP}}{\mathrm{TP}+\mathrm{FP}},\qquad
R=\frac{\mathrm{TP}}{\mathrm{TP}+\mathrm{FN}},\qquad
F_1=\frac{2\mathrm{TP}}{2\mathrm{TP}+\mathrm{FP}+\mathrm{FN}}.
\label{eq:repair-f1}
\end{equation}
For illustration, suppose a task requires restoring a missing globe,
repositioning a chair, and deleting an extra box. A generated scene restores the
globe and deletes the box, but leaves the chair misplaced and deletes an
unrelated plant. The globe and box contribute two TP; the misplaced chair
contributes one FP and one FN; the lost plant adds one FP. Hence
$(\mathrm{TP},\mathrm{FP},\mathrm{FN})=(2,2,1)$, giving $P=0.5$,
$R=2/3$ and $F_1=4/7$.

We set $P=0$ when its denominator is zero. Nonempty target sets make the recall
and F1 denominators positive. Unchanged background actors do not contribute
true negatives: arbitrary new actors and edit states do not define a finite
negative population. Cases assigned zero under
Appendix~\ref{app:score-aggregation} receive zero Precision, Recall and F1;
these reporting zeros are not measured TP/FP/FN counts.

\paragraph{Aggregating case scores.}
Precision, Recall and F1 are averaged over scheduled cases with equal case
weights (Appendix~\ref{app:score-aggregation}). Macro F1 is not the harmonic
mean of macro Precision and Recall; pooling counts would give greater weight
to cases with more targets and false positives. Repair F1 enters the I2S
aggregate defined in Eq.~\ref{eq:i2s-score}.

\paragraph{Matching actors across input and generated scenes.}
The input--generated-scene correspondence identifies which actors were preserved,
changed, added or removed before the outcome counts above are computed.
It does not use actor names or GUIDs. Matching first maximizes unchanged
correspondences, preserving the multiplicity of unaffected background actors
before matching target actors. Remaining actors with the same structural
descriptor are assigned at minimum cost,
$\delta/(1+\delta)+0.2\varphi/180+0.05k$, where $\delta$ is position distance
in cm, $\varphi$ is quaternion angular distance in degrees, and $k$ is the
number of changed fields. Remaining same-class actors at the same pose can
anchor an asset replacement or cleared mesh. Such correspondence identifies
an edit; repair success still requires the ground-truth tests above.
In an asset replacement, removing the old asset can satisfy a required
removal, while the replacement is assessed separately as a candidate presence.

\paragraph{Evidence and numerical tolerances.}
Unchanged-state detection uses tolerances for editor save/export round trips:
$0.1$ cm per position axis, $0.01^\circ$ quaternion rotation, and $10^{-4}$
absolute scale per axis, together with agreement of recorded attributes.
These tolerances identify changes; the larger thresholds in
Table~\ref{tab:repair-thresholds} determine repair success. Equivalent Euler
representations and renamed or recreated equivalent actors do not create
spurious additions or deletions. Bounds-only changes of background actors are
not independent edits, but target bounds remain subject to acceptance tests.
Comparisons use absolute Unreal world coordinates, without fitting a global
translation or yaw offset or allowing object-symmetry equivalences.

Rotations are compared as unit quaternions, and signed scale detects
reflections. Missing, malformed or nonfinite required evidence fails. A zero
ground-truth scale is invalid; a zero bounds-extent component passes only if
the candidate component is also zero. Bounds and attribute channels absent
from the ground truth are not required; the generated scene cannot disable a
required channel. Numeric comparisons allow a $10^{-9}$ boundary guard.
Thresholded one-to-one pose evaluation has precedent in Scan2CAD and
BOP~\citep{avetisyan2019scan2cad,hodan2020bop}; their error functions, symmetry
policies and thresholds are not adopted here. The nominal profile is
$(5\text{ cm},5^\circ,5\%)$; Appendix~\ref{app:selective-repair-analysis} also tests strict
$(2.5\text{ cm},2.5^\circ,2.5\%)$ and relaxed
$(10\text{ cm},10^\circ,10\%)$ tolerances on public cases.

\subsection{Score Aggregation and Reporting}
\label{app:score-aggregation}
The Text-to-Scene case score assigns 20\% to Detailed Alignment, 60\% to
Overview Alignment, and 20\% to Physical Safety. The Image-to-Scene case score
assigns 80\% to actor Repair F1 and 20\% to Physical Safety:
\begin{equation}
S_{\mathrm{I2S}}=\lambda_{\mathrm{rep}}F_1
 +(1-\lambda_{\mathrm{rep}})S_{\mathrm{physics}},
\qquad \lambda_{\mathrm{rep}}=0.80.
\label{eq:i2s-score}
\end{equation}
The same weights apply to all configurations within each setting. T2S weights
were informed by human judgments and are assessed in
Section~\ref{sec:evaluator-validation}. Scores lie in $[0,1]$; multiplying by
100 gives the percentage scale used in some figures.

Every retained scheduled case remains in its evaluation subset's denominator; asset-use exclusions apply uniformly across configurations.
Case \(i\) contributes
\[
s_i
=
\begin{cases}
S_i, & \text{for a valid submission with an authoritative primary score},\\
0, & \text{for an authoritative invalid verdict, a missing submission, or a missing result}.
\end{cases}
\]
All subscores, including repair Precision, Recall and F1, are zero for an
invalid submission, including dependency-check failures. Missing submissions and missing results therefore also count as zero and remain in the denominator, as for the seven absent Muse Spark 1.3 (medium) submissions.
Within an otherwise valid case, unavailable required
components retain their weights and receive zero credit; explicit
inapplicability and optional visual fallbacks follow the metric-specific rules
in Appendices~\ref{app:physical-safety}, \ref{app:semantic-verifier},
and~\ref{app:gt-verifier}.

Table~\ref{tab:invalid-breakdown} lists the nine zero-score pairs among 1,162
public evaluations. Unresolved dependencies arise when the generated scene refers
to assets created during the run that were not saved with it. The seven absent
Muse submissions correspond to Outdoor cases built from MiddleEasternTown,
which is distinct from the excluded MiddleEast pack.

\begin{table}[!t]
\centering
\compacttable
\caption{Public evaluations assigned zero, by configuration and reason
(9 of 1,162 pairs). Configurations with no such cases are omitted.}
\label{tab:invalid-breakdown}
\small
\setlength{\tabcolsep}{3.5pt}
\begin{tabular}{llccccc}
\toprule
\rowcolor{sbTableHead}
Agent configuration & Track & \shortstack{Unresolved\\dependencies} & \shortstack{Empty\\scene} &
\shortstack{No\\submission} & \shortstack{Missing\\result} & Total \\
\midrule
Muse Spark 1.3 (medium)      & Outdoor & -- & -- & 7  & -- & 7 \\
Qwen 3.8 27B (thinking on)  & T2S     & 1  & -- & -- & -- & 1 \\
Qwen 3.8 27B (thinking off) & T2S     & -- & 1  & -- & -- & 1 \\
\midrule
Total               &         & 1  & 1  & 7  & 0  & 9 \\
\bottomrule
\end{tabular}
\end{table}

For an evaluation subset \(t\) with all \(N_t^{\mathrm{score}}\) scheduled contributions resolved,
the mean score is
\[
\overline S_t=\frac{1}{N_t^{\mathrm{score}}}\sum_{i=1}^{N_t^{\mathrm{score}}}s_{ti}.
\]
Otherwise the aggregate remains unreported, without dropping missing cases
or publishing score bounds. The overall model score is
\begin{equation}
S_{\mathrm{model}}
=0.50\overline S_{\mathrm{T2S}}
+0.50\overline S_{\mathrm{I2S}}.
\label{eq:model-score}
\end{equation}
Here the Image-to-Scene mean gives each scheduled case equal weight:
\[
\overline S_{\mathrm{I2S}}=
\frac{N_{\mathrm{in}}\overline S_{\mathrm{I2S,indoor}}+
N_{\mathrm{out}}\overline S_{\mathrm{I2S,outdoor}}}
{N_{\mathrm{in}}+N_{\mathrm{out}}}.
\]
The public set uses 22 Indoor and 45 Outdoor cases, giving domain weights
$22/67$ and $45/67$; the private set uses 50 and 27, giving $50/77$ and $27/77$.
The retained full suite uses 72 cases in each domain and therefore equal domain weights.
Component scores are reported alongside the aggregate. Known Coverage, visual
availability, resource usage, and embodied utility are auxiliary measurements
and do not enter \(S_{\mathrm{model}}\). An overall score requires complete
results for both settings. Case-level T2S scores are rounded to four decimal
places and I2S scores to six before aggregation; tables use three decimals.

\subsection{Worked Scoring Examples}
\label{app:evaluator-scoring-examples}

\begin{figure}[!htb]
\centering
\includegraphics[width=\linewidth]{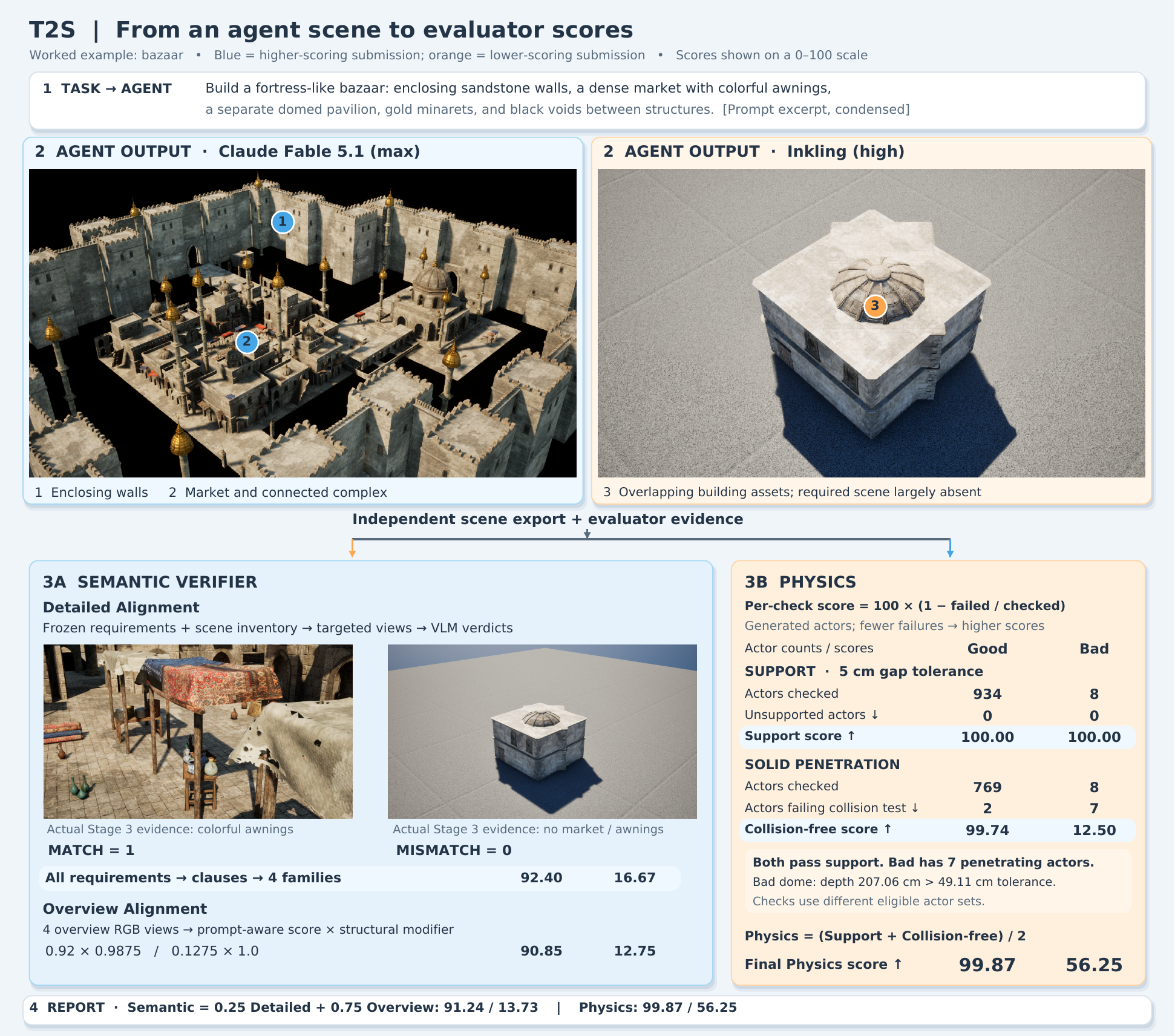}
\caption{\textbf{Text-to-Scene scoring flow on Bazaar.}
Claude Fable 5.1 (max) and Inkling (high) receive the same prompt. Targeted
views in panel 3A resolve the colorful-awnings requirement; panel 3B shows
support and penetration checks with their respective eligible actor
populations (Appendix~\ref{app:physical-safety}). The dome marker denotes an
engine-confirmed collision. Captures are annotated for clarity.}
\label{fig:t2s-scoring-flow}
\end{figure}

\begin{figure}[!htb]
\centering
\includegraphics[width=\linewidth]{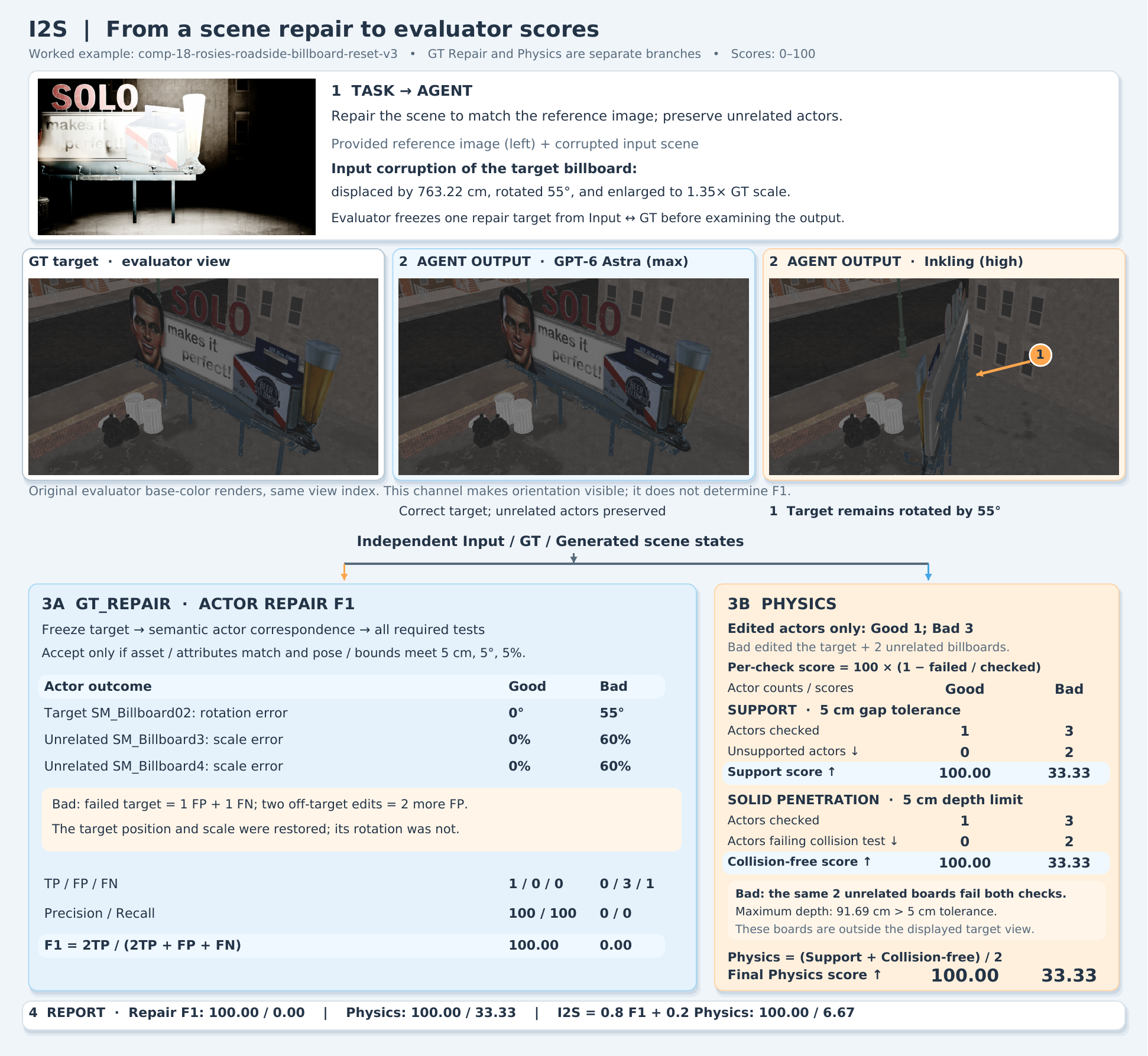}
\caption{\textbf{Image-to-Scene scoring flow on roadside billboard reset.}
GPT-6 Astra (max) restores the target and preserves unrelated actors. Inkling
(high) leaves a $55^\circ$ rotation error and enlarges two unrelated billboards;
these off-target billboards also fail the physical checks and lie outside the
displayed view. The reference image is the agent input; comparison images are
evaluator base-color renders. Repair F1 uses structured scene states, while
Physical Safety uses engine measurements.}
\label{fig:i2s-scoring-flow}
\end{figure}

Figures~\ref{fig:t2s-scoring-flow} and~\ref{fig:i2s-scoring-flow} trace
two tasks from agent output through the evidence and decisions used by each
verifier. Each pairs two valid submissions to the same task, selected for
visible differences and available evaluation evidence. The examples illustrate
the scoring procedure rather than estimate performance across tasks. The Bazaar
demonstration is historical and is excluded from the retained benchmark scores.
Displayed component scores use a 0--100 scale; binary requirement verdicts
and the intermediate calculations retain their native 0--1 values.

\subsection{Embodied Utility Score Implementation}
\label{app:eus-details}

Scene-generation evaluations have measured reachability through layout
connectivity or object clearance
\citep{yang2024physcene,tam2026sceneeval}. Embodied navigation evaluates
whether routes are actually traversed \citep{anderson2018evaluation}.
EUS combines these complementary checks in reopened Unreal scenes
\citep{ye2025simworld}. The fixed follower isolates scene usability from policy decisions
\citep{chang2024partnr}; SceneSmith likewise separates reach and execution
\citep{pfaff2026scenesmith}.

\paragraph{Definition and scope.}
EUS measures navigation by a player agent with a person-sized collision body
and a fixed route-following controller in Unreal Engine. It evaluates scene
usability rather than general interaction or learned-policy performance.
The generated scene is reopened in a fresh editor session. Let \(T\) be its pool of eligible
floor-level, collidable actors. Approachability is the share of this content
connected to the main navigable region,
\begin{equation}
A=\frac{\bigl|\{t\in T:\mathrm{approachable}(t)\}\bigr|}{|T|}.
\label{eq:eus-a}
\end{equation}
Walk validation is a simulated capsule's completion rate on selected
non-trivial routes,
\begin{equation}
W=\frac{n_{\mathrm{arrived}}}{n_{\mathrm{walks}}},
\qquad
r_{\mathrm{arr}}=\min\bigl(300,\max(120,0.25h)\bigr)\ \mathrm{cm},
\label{eq:eus-w}
\end{equation}
where \(r_{\mathrm{arr}}\) is the arrival radius and \(h\) is the smaller half-extent of the
settled navigation mesh. The raw per-scene result is
\begin{equation}
\mathrm{EUS}=
\begin{cases}
0, & |T|<15\ \text{or certified }A=0,\\[1mm]
A\,W, & A\text{ and }W\text{ are measured},\\[1mm]
\textsc{withheld}, & \text{the navigation measurement is unreliable}.
\end{cases}
\label{eq:eus}
\end{equation}
A withheld result is not a certified zero and does not establish that the
scene is unusable.

\paragraph{Eligible targets and the floor-level pool.}
The target pool uses live actors and their collision geometry from the reopened
\textrm{.umap}. Lights, fog, reflection captures,
sky and particle effects, cameras, engine bookkeeping actors, and surfaces
such as floors, roads, and sidewalks are excluded. Semantic categories are
used when available; otherwise documented name patterns are applied with
near-miss checks. An actor whose base lies more than \(150\) cm above or
\(50\) cm below the navigation point beneath it is treated as scenery and
removed from both numerator and denominator. A floor-level actor that no
navigation cell reaches remains in the denominator as unreachable.

\paragraph{Measuring \(A\).}
Actor footprints come from colliding-component bounds, with the transform
pivot used only when collision bounds are unavailable. A single component
wider than \(200\) m in plan or \(50\) m in height is excluded from navigation
bounds so that an environment proxy cannot dominate the build volume. The
navigation mesh uses agent radius \(50\) cm, agent height \(180\) cm, maximum
slope \(35^\circ\), and minimum region area \(150\); the evaluator writes and
reads back these values for every scene. A build that does not settle within
\(300\) s is withheld.

The settled mesh is sampled by deterministic three-dimensional lattices with
\(200\) cm pitch for connected components, \(25\) cm pitch for target
projection, and \(100\) cm vertical strata. Each target footprint is projected
through nine samples to cells within \(100\) cm in plan, preferring cells on
the target's own storey. A target is approachable when its projected cell has
a complete path to the largest connected component. At most three builds are
attempted; certification requires two consecutive builds to agree on \(A\)
within \(0.05\) and on the number of on-mesh cells within \(0.5\%\).

\paragraph{Measuring \(W\).}
The walk stage rebuilds the same mesh and settings. Its anchor is the
floor-level member of the largest component nearest its centroid and within
\(150\) cm of the ground median. Targets already within \(r_{\mathrm{arr}}\) are excluded;
the remaining targets are selected at the \(0.20\), \(0.55\), and \(0.90\)
quantiles of route distance. The engine plans each route, and a capsule follows
it at \(200\) cm/s. Two five-second intervals with less than \(10\) cm progress
terminate a stalled walk. A walk succeeds only if the capsule ends within
\(r_{\mathrm{arr}}\) of its goal.

For a scene with half-extent below \(1200\) cm, the protocol instead selects
six object-to-object routes of length at least \(4r_{\mathrm{arr}}\). If no non-trivial route
exists, \(W\) is withheld. Only this oracle route follower contributes to EUS;
more autonomous navigation agents are diagnostic and do not enter the score.

\begin{figure}[!t]
    \centering
    \includegraphics[width=\linewidth]{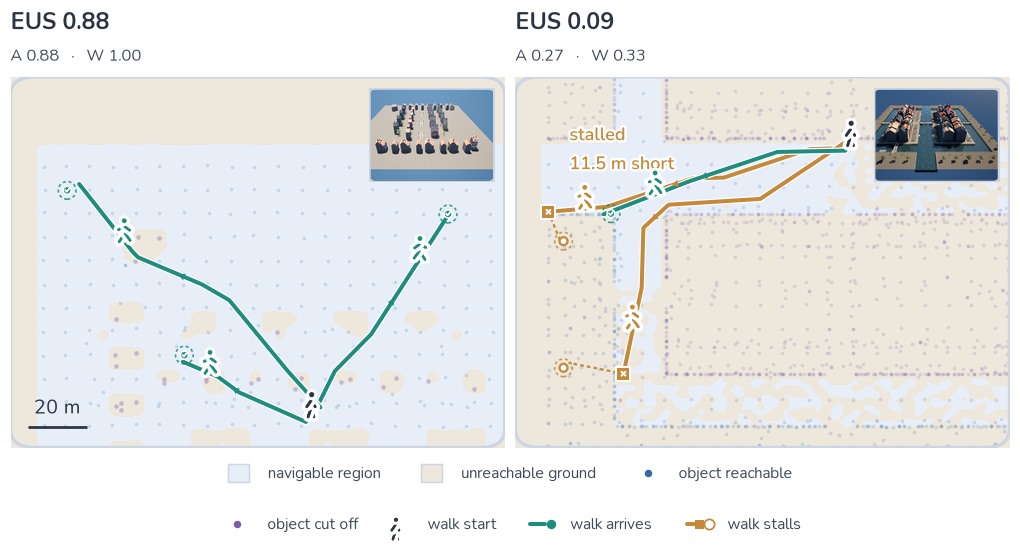}
    \caption{EUS scoring illustration on the same Nordic Harbour prompt. For approachability $A$, most floor-level objects in the left scene are reachable ($A=0.88$), whereas most floor-level objects in the right scene are not ($A=0.27$). For walk validation, three routes lead from a floor-level anchor to targets at the near, mid, and far route-distance quantiles; all three are completed in the left scene and one in the right. The overall EUS is therefore $0.88$ for the left scene and $0.09$ for the right.}
    \label{fig:eus-score-comparison}
\end{figure}

\paragraph{Withholding and aggregation.}
Failure to settle or certify the mesh, or to obtain a non-trivial route,
produces a withheld result with a recorded reason. A scene
blocked by an evaluation-environment content failure is excluded rather than
attributed to the model. For the conservative reported aggregate, a withheld
scene contributes zero while retaining its distinct withheld status. For task
set \(t\), define \(c_{ti}\) as the measured EUS (including certified zeros)
or zero if withheld, and let $N_t^{\mathrm{EUS}}$ be the number of retained scene records.
The aggregate is
\[
\overline{\mathrm{EUS}}_t=\frac{1}{N_t^{\mathrm{EUS}}}\sum_{i=1}^{N_t^{\mathrm{EUS}}}c_{ti}.
\]
Table~\ref{tab:eus-leaderboard} reports measured-only means and coverage
alongside \(\overline{\mathrm{EUS}}_t\). Standard errors summarize between-scene variation; fixed routes do
not estimate variability across randomly sampled walks. Indoor results are grouped by base room because edits share its geometry. Figure~\ref{fig:eus-score-comparison} shows an example of how EUS is calculated for a scene.

\subsection{Human-Alignment Protocol and Weight Fitting}
\label{sec:human-alignment}

The human-alignment study evaluates whether the composite Text-to-Scene score
reproduces blinded human preferences, following evaluator-validation protocols
for generative systems~\citep{ghosh2023geneval,zheng2023judging,ning2026vibeworlding}.
It contains 16 prompts and seven agent
configurations per prompt, yielding 112 generated scenes. The human-review cohort comprises Qwen 3.8 27B (thinking off), Qwen 3.8 27B (thinking on),
Gemma 4 31B (thinking off), Gemma 4 31B (thinking on), Gemini 3.8 Flash (high),
GPT-5.6 Sol (high), and GLM-5.3 Flash (max).

Each task presents three anonymous candidates and asks reviewers for a strict
best-to-worst ranking, without ties or abstentions. A candidate is represented
by four fixed overview images from its verifier run, with view switching and
image enlargement. Reviewers see the scene
description and the criterion being judged, while model identities, verifier
scores, and other reviewers' answers remain hidden. Seven overlapping triplets
form a balanced incomplete-block design over the seven candidates. Each
candidate appears in three triplets, and each of the 21 candidate pairs appears
in exactly one triplet. The tasks are distributed among seven graduate and undergraduate student
annotators. For the primary Overall Task
Success criterion, three different annotators rank each triplet, yielding
\(16\times7\times3=336\) triplet rankings and 1,008 pairwise preferences.

The interface separates Overall Task Success from six diagnostic criteria,
namely Identity and Environment, Content and Quantity, Spatial Composition,
Attributes and Materials, Holistic Overview Alignment, and Physical
Plausibility. The four semantic criteria map to the Detailed Alignment
families, Holistic Overview Alignment maps to Overview Alignment, and Physical
Plausibility concerns visible unsupported placement and penetration. The
primary reference is the independent Overall Task Success ranking, which is
compared directly with \(S_{\mathrm{T2S}}\); it is not constructed by combining
the diagnostic human rankings. Reviewers see the same four overview images used
by Overview Alignment, but not other components' engine measurements or targeted views.

For each prompt and criterion, a Plackett-Luce model estimates a consensus
ordering from the original triplet rankings~\citep{turner2020plackettluce}.
The implementation applies an L2 coefficient of \(0.1\) to the log-worths so
that unanimous preferences yield finite estimates. Each triplet is also
expanded into three pairwise
preferences. Inter-annotator agreement is measured on the same candidate pair,
while verifier agreement is reported against both individual preferences and
the Plackett-Luce consensus. Kendall's \(\tau_b\) is the primary rank
association statistic, with Spearman's \(\rho\), nominal Krippendorff's
\(\alpha_{\mathrm K}\), leave-one-reviewer-out agreement, and agreement between disjoint
reviewer subsets reported as diagnostics. Agreement and rank correlation are
computed within prompts and macro-averaged. Confidence intervals use
whole-prompt bootstrap resampling with 5,000 draws for agreement and rank
correlation and 2,000 draws for \(\alpha_{\mathrm K}\), to account for shared candidates
within prompts.

\paragraph{Aggregation weight fitting.}
\label{sec:human-weight-fitting}
We fit \(\theta\) to maximize agreement with individual human Overall Task
Success preferences. A deterministic, score-independent split of case
identifiers (seed 20260915) assigns 12 prompts to development and four to
retrospective holdout evaluation. Let \(\mathcal P_{\mathrm{dev}}\) denote the
development prompts, \(\mathcal Q_p\) the 21 candidate pairs for prompt \(p\),
and \(\mathcal R_{pij}\) the three reviewers of pair \((i,j)\).
Let \(\mathbf x_{pi}\) denote the fixed evidence for candidate \(i\) on
prompt \(p\), as defined for \(S_{\mathrm{T2S}}(\mathbf x;\theta)\) above. Write
\(y_{pijr}=+1\) if reviewer \(r\) prefers candidate \(i\) and \(-1\)
otherwise, and let
\(\Delta_{pij}(\theta)=S_{\mathrm{T2S}}(\mathbf x_{pi};\theta)
-S_{\mathrm{T2S}}(\mathbf x_{pj};\theta)\). The primary objective is
\begin{equation}
\begin{aligned}
&\underset{\theta\in\Theta}{\operatorname{maximize}}\;
  \mathcal A_{\mathrm{dev}}(\theta),\\
&\mathcal A_{\mathrm{dev}}(\theta)
=\frac{1}{|\mathcal P_{\mathrm{dev}}|}
\sum_{p\in\mathcal P_{\mathrm{dev}}}\frac{1}{|\mathcal Q_p|}
\sum_{(i,j)\in\mathcal Q_p}\frac{1}{|\mathcal R_{pij}|}
\sum_{r\in\mathcal R_{pij}}
\mathbf 1\!\left[y_{pijr}\Delta_{pij}(\theta)>\varepsilon\right],
\end{aligned}
\label{eq:human-weight-objective}
\end{equation}
where \(\varepsilon=10^{-8}\) identifies numerical ties, which receive no
credit against strict preferences. Prompts contribute equally. The feasible
set \(\Theta\) uses nonnegative normalized weight blocks, a fixed final
Physical Safety share \(\gamma=0.20\), \(\alpha\geq0.20\), and positive
total weight for each applicable-family set.

The search combines a 0.05 simplex grid, 8,192 seeded joint proposals, and up
to five coordinate-refinement passes with 0.01 pairwise weight transfers.
Equal-agreement candidates are ordered by pairwise logistic loss
\(\log(1+\exp[-10y\Delta])\), averaged with the same prompt and pair weights,
and then by squared distance to the comparison policy. This is a bounded
numerical search rather than exhaustive continuous optimization. Scores are
reconstructed from saved component scores, preserving rounding, applicability
rules, and the fixed Overview Alignment modifier and cap; neither VLM
judgments nor engine measurements are refitted. The secondary fit instead uses
within-prompt Plackett-Luce consensus labels. All fitted parameters are frozen
before holdout evaluation.

The adopted weights came from earlier human-guided exploration, which used
all 16 prompts. The split therefore provides retrospective internal validation
of the fitted alternatives. Component measurements remain fixed throughout.

\begin{table}[!t]
\centering
\caption{Aggregation weights for the adopted score and two preference fits.
Each vector follows the displayed subscore order and sums to one; both fits
fix the final Physics share at $\gamma=0.20$. Fitting and the retrospective
holdout protocol are defined in Appendix~\ref{sec:human-weight-fitting}.}
\label{tab:human-alignment-weights}
\small
\setlength{\tabcolsep}{4pt}
\begin{tabularx}{\linewidth}{>{\raggedright\arraybackslash}Xccc}
\toprule
\rowcolor{sbTableHead}
\textbf{Weights and order} & \textbf{Adopted}
& \shortstack{\textbf{Individual}\\\textbf{fit}}
& \shortstack{\textbf{PL}\\\textbf{fit}} \\
\midrule
$\mathbf w^D$: identity/environment, content/quantity,
attributes/materials, spatial composition
& \cellcolor{sbBlue!9}$(.25,.40,.15,.20)$ & $(.18,.30,.13,.39)$ & $(.08,.20,.17,.55)$ \\
$\mathbf w^O$:
prompt alignment, layout, style/atmosphere, completeness/polish
& \cellcolor{sbBlue!9}$(.40,.25,.20,.15)$ & $(0,0,1,0)$ & $(.10,0,.90,0)$ \\
$\mathbf w^P$: supported fraction,
collision-free fraction
& \cellcolor{sbBlue!9}$(.50,.50)$ & $(0,1)$ & $(0,1)$ \\
$(\alpha,\beta,\gamma)$: Detailed, Overview, Physics
& \cellcolor{sbBlue!9}$(.20,.60,.20)$ & $(.25,.55,.20)$ & $(.30,.50,.20)$ \\
\bottomrule
\end{tabularx}
\end{table}

Empirical comparisons appear in Appendix~\ref{app:evaluator-validation-results}.

\subsection{Fixed-Evidence Repeatability Protocol}
\label{app:judge-repeatability-protocol}

\paragraph{Scenes and repeated judgments.}
The diagnostic cohort contains nine agent configurations on six prompts, namely
\emph{Battleground}, \emph{Classical Modular Gardens}, \emph{Cyberpunk},
\emph{Industrial Harbor}, \emph{Medieval Buildings Volume 1}, and
\emph{Street New York}. The prompts were selected for scene-type coverage
and reusable evidence, without optimizing model rankings or score differences.
They form a purposive subset, not a random sample of the full benchmark.
The cohort includes Gemini 3.7 Flash (high) but excludes six main-leaderboard configurations: GPT-6 Astra (max), Claude Fable 5.1 (max), Claude Opus 5 (max), Grok 4.6 (high), Muse Spark 1.3 (medium), and DeepSeek V4.1 Flash (high). Each of the 54 generated scenes was scored
three times. Scenes, requirement bundles, image order, earlier-stage
decisions, and the recorded visual-request schedule were fixed; each repeat
made fresh VLM requests. No scene generation, image capture, or Overview
Alignment scoring was repeated.

\paragraph{Judge and aggregation.}
The study used the fixed judge configuration in
Appendix~\ref{app:judge-configuration}. The saved request schedule
contained 8{,}764 logical requests per repeat, excluding transport retries.
The 54 scenes and three repeats yield 162 scene-level scores.
Requests were served by one to three replicas with matching model metadata
and varying concurrency; the repeats include this serving variation.
Detailed Alignment used the same family
weights in every repeat, $(0.25,0.40,0.20,0.15)$ for identity/environment,
content/quantity, spatial composition, and attributes/materials, respectively.

The repeated-score results appear in Appendix~\ref{app:judge-repeatability}.

\subsection{Evaluator Integrity}

The packaging audit scans task packages for source maps, external
actors, Level Instances, packed actors, prefabs that encode the target
assembly, cached layouts, screenshots, and revealing metadata. Mutation tests
cover proxy geometry, off-camera objects, disabled collision, excessive
supports, camera changes, and protected-region edits to verify that these
shortcuts are not rewarded.

\section{Additional Results and Analysis}
\label{app:trajectory-methods}

This appendix supplements Section~\ref{sec:analysis} with complete task
profiles, repair diagnostics and navigation results. It also covers
analysis cohorts, difficulty strata, agent behavior and evaluator validation.

\subsection{Analysis Cohorts and Measurements}
\label{app:trajectory-cohort}
\label{app:trajectory-coverage}

The public set contains 14 configurations on 16 T2S, 22 Indoor and 45 Outdoor
cases (1,162 configuration--case pairs). The private set contains ten
configurations on 10 T2S, 50 Indoor and 27 Outdoor cases (870 pairs), selected
by removing the retained public task IDs from the retained full suite.
Together, the two sets cover 26 T2S and 72 cases per repair domain for the ten
common configurations. GPT-6 Astra (max), Grok 4.6 (high), Claude Opus 5
(max) and Claude Fable 5.1 (max) have only public coverage. Comparisons across
task sets use the same ten configurations. Public and private tasks are
disjoint but can share source scenes.

All benchmark score, cost, trajectory, spatial and repair analyses use this
retained-case selection. Current scores, token costs, total tool-call counts and
conditional spatial diagnostics use the public T2S refresh described in
Appendix~\ref{app:rerun-configurations}. The separately labelled Python-operation
and execution-feedback profiles retain the earlier generation cohort.
Repair diagnostics exclude the exporter artifact. Score means include failed submissions under
Appendix~\ref{app:score-aggregation}; trajectory statistics use recorded runs.
Scores use $[0,1]$ unless a 0--100 scale is stated. The separately collected
human-judgment, fixed-evidence judge-repeatability and navigation studies retain
their explicitly stated auxiliary cohorts.

\paragraph{Trajectory measurements.}
\label{app:trajectory-extraction}
Each configuration--case pair contributes one retained attempt under
Section~\ref{sec:evaluation-protocol}. Tool counts and token-cost estimates
exclude discarded attempts, rather than summing across retries.
Tool counts include requests with a metered start, including failed and
unfinished requests, with duplicate request IDs counted once. Non-polling
counts exclude asynchronous status checks (the \texttt{read\_job\_log} tool). Mean calls use recorded
trajectories, including zero-call runs.

A fixed API-basename dictionary classifies parseable Python submissions into
eight operation families. Families may co-occur; percentages use parseable
scripts and describe attempted operations. Execution feedback links
synchronous responses and asynchronous logs to their originating submissions;
repeated error reports count once per submission. A startup acknowledgement
alone does not establish execution feedback. Missing code or feedback cannot rule out an operation or error.

\paragraph{Uncertainty.}
Confidence intervals use a source-cluster bootstrap. Each cluster groups
cases derived from the same source environment and their results across
configurations. Clusters are resampled with replacement within task domains,
retaining all cases and configurations in each sampled cluster.
Score and trajectory comparisons use 10,000 draws with seed 23917 (16/10 public/private
T2S sources; each task set has nine Indoor rooms from eight source-level
clusters, and nine Outdoor source clusters). Spatial and selective-repair diagnostics use
10,000 draws with seed 20260925. Repair resampling is stratified by domain,
retaining observed domain weights. Reported intervals are pointwise percentile
95\% intervals over retained sources; each configuration--case pair has one
selected run. They describe source heterogeneity, not generation variability,
and do not use multiplicity-adjusted significance thresholds. Current score intervals preserve the fixed weights: one half for T2S and one half for I2S, with Indoor/Outdoor weighted 22/67 and 45/67 within public I2S. All models share the same source draws. Table~\ref{tab:overall-uncertainty} reports the resulting intervals; the Astra--Gemini overall paired difference is $0.0134$ [$-0.0402$, $0.0665$].

\begin{table}[!t]
\centering
\compacttable
\caption{Public-set scores with pointwise 95\% source-cluster bootstrap intervals. Paired systems share resampled sources within each task domain; the T2S/I2S and Indoor/Outdoor aggregation weights remain fixed. These intervals describe variation over retained sources, not repeated generation.}
\label{tab:overall-uncertainty}
\footnotesize
\setlength{\tabcolsep}{3pt}
\begin{tabularx}{\linewidth}{>{\raggedright\arraybackslash}Xrrr}
\toprule
\rowcolor{sbTableHead}
System & T2S & I2S & Overall \\
\midrule
GPT-6 Astra (max) & 0.785 [0.760, 0.810] & 0.530 [0.456, 0.610] & 0.658 [0.619, 0.700] \\
Gemini 3.8 Flash (high) & 0.675 [0.634, 0.717] & 0.613 [0.527, 0.706] & 0.644 [0.596, 0.696] \\
Claude Fable 5.1 (max) & 0.796 [0.764, 0.824] & 0.428 [0.357, 0.519] & 0.612 [0.573, 0.659] \\
Claude Opus 5 (max) & 0.742 [0.701, 0.778] & 0.469 [0.380, 0.573] & 0.606 [0.558, 0.660] \\
GPT-5.6 Sol & 0.753 [0.722, 0.780] & 0.387 [0.344, 0.429] & 0.570 [0.544, 0.595] \\
GLM-5.3 Flash (max) & 0.642 [0.600, 0.684] & 0.326 [0.263, 0.379] & 0.484 [0.446, 0.518] \\
Muse Spark 1.3 & 0.537 [0.405, 0.656] & 0.348 [0.278, 0.415] & 0.443 [0.369, 0.512] \\
Grok 4.6 & 0.682 [0.654, 0.708] & 0.187 [0.150, 0.223] & 0.434 [0.411, 0.456] \\
Qwen 3.8 27B (thinking on) & 0.538 [0.444, 0.621] & 0.238 [0.190, 0.299] & 0.388 [0.336, 0.439] \\
Qwen 3.8 27B (thinking off) & 0.503 [0.407, 0.587] & 0.204 [0.164, 0.250] & 0.353 [0.303, 0.401] \\
Gemma 4 31B (thinking off) & 0.531 [0.469, 0.595] & 0.117 [0.105, 0.128] & 0.324 [0.292, 0.356] \\
Inkling (high) & 0.438 [0.387, 0.490] & 0.206 [0.160, 0.251] & 0.322 [0.288, 0.356] \\
Gemma 4 31B (thinking on) & 0.485 [0.425, 0.545] & 0.127 [0.112, 0.142] & 0.306 [0.275, 0.337] \\
DeepSeek V4.1 Flash & 0.352 [0.228, 0.477] & 0.219 [0.170, 0.277] & 0.285 [0.218, 0.353] \\
\bottomrule
\end{tabularx}
\end{table}

\subsection{Difficulty-Stratified Results}

\paragraph{Difficulty-stratified scores.}
\label{app:difficulty-results}
Table~\ref{tab:difficulty-results} reports scores for ten configurations on the
retained 170-case suite using the authored Easy/Medium/Hard mapping. Tier counts
are 8/8/10 for T2S, 25/22/25 for Indoor and 24/24/24 for Outdoor. Cases have equal weight within each tier; scores follow
Appendix~\ref{app:score-aggregation}.

Across configurations, I2S means decrease from Easy to Medium to Hard:
$0.327/0.272/0.244$; the corresponding T2S means are $0.600/0.522/0.501$.
These tiers are descriptive strata; individual configurations can reverse
the ordering.
\begin{table*}[!t]
\centering
\compacttable
\caption{Difficulty-stratified scores for ten agent configurations on the retained 170-case suite of public and private cases (Appendix~\ref{app:difficulty-results}).}
\label{tab:difficulty-results}
\footnotesize
\setlength{\tabcolsep}{3pt}
\begin{tabularx}{\textwidth}{>{\raggedright\arraybackslash}Xrrrrrrr}
\toprule
\rowcolor{sbTableHead}
\textbf{System} & \shortstack{T2S\\Easy} $\uparrow$ & \shortstack{T2S\\Medium} $\uparrow$ & \shortstack{T2S\\Hard} $\uparrow$ & \shortstack{I2S\\Easy} $\uparrow$ & \shortstack{I2S\\Medium} $\uparrow$ & \shortstack{I2S\\Hard} $\uparrow$ & $S_{\mathrm{model}}\uparrow$ \\
\midrule
Gemini 3.8 Flash (high) & \tsecond[sbBlue]{0.722} & \tsecond[sbBlue]{0.662} & \tsecond[sbBlue]{0.662} & \tbest[sbOrange]{0.638} & \tbest[sbOrange]{0.647} & \tbest[sbOrange]{0.597} & \tbest[sbInk]{0.654} \\
GPT-5.6 Sol & \tbest[sbBlue]{0.805} & \tbest[sbBlue]{0.700} & \tbest[sbBlue]{0.719} & 0.456 & \tsecond[sbOrange]{0.407} & \tsecond[sbOrange]{0.372} & \tsecond[sbInk]{0.576} \\
GLM-5.3 Flash (max) & 0.690 & 0.523 & 0.628 & 0.423 & 0.359 & 0.285 & 0.485 \\
Muse Spark 1.3 & 0.668 & 0.485 & 0.515 & \tsecond[sbOrange]{0.492} & 0.297 & 0.314 & 0.461 \\
Qwen 3.8 27B (thinking on) & 0.584 & 0.533 & 0.491 & 0.297 & 0.230 & 0.151 & 0.379 \\
Qwen 3.8 27B (thinking off) & 0.628 & 0.542 & 0.430 & 0.230 & 0.202 & 0.152 & 0.360 \\
Gemma 4 31B (thinking on) & 0.497 & 0.514 & 0.477 & 0.128 & 0.126 & 0.117 & 0.309 \\
Gemma 4 31B (thinking off) & 0.568 & 0.546 & 0.415 & 0.117 & 0.110 & 0.114 & 0.308 \\
Inkling (high) & 0.544 & 0.404 & 0.386 & 0.230 & 0.158 & 0.121 & 0.305 \\
DeepSeek V4.1 Flash & 0.295 & 0.308 & 0.292 & 0.258 & 0.183 & 0.217 & 0.259 \\
\bottomrule
\end{tabularx}
\end{table*}

\subsection{Task Requirement Analysis}
\label{app:insight-diagnostics}

\label{app:task-category-analysis}
\begin{figure*}[!htbp]
% Updated from canonical CSVs with build_f1_figure5.py; original task mapping retained.
\centering
\includegraphics[width=\textwidth]{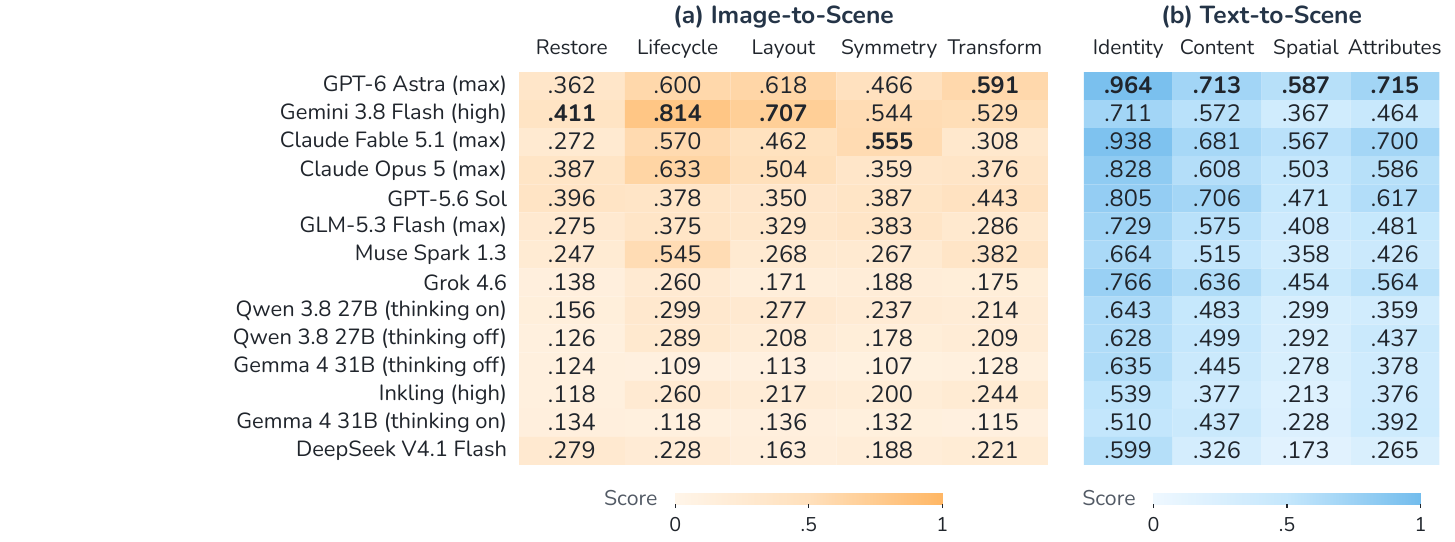}
\caption{Task-category performance of 14 agent configurations on the 83 retained public cases:
(a) Image-to-Scene scores by operation category and
(b) Text-to-Scene Detailed Alignment.}
\label{fig:i2s-operation-heatmap}
\end{figure*}

Figure~\ref{fig:i2s-operation-heatmap} gives the complete repair-category and
construction-requirement profiles for all 14 public-set configurations,
extending the frontier comparison in Section~\ref{sec:analysis}. Beyond the contrasts discussed there, Gemini leads Restore ($0.411$) across all 14 configurations. Spatial Composition is the weakest construction family for every configuration.

\subsection{Agent Behavior Analysis}
\label{app:trajectory-subset14}
\label{app:trajectory-full10}
\label{app:trajectory-private10}
\label{sec:harness-ablations}
\label{sec:joint-trajectory-evaluation}

\paragraph{Interaction volume and scene quality.}
\label{app:trajectory-effort}
Table~\ref{tab:trajectory-paired} compares Astra and Gemini on the same public
tasks using current scores and their selected-run tool counts. Astra averages
97.2 non-polling calls per construction case versus Gemini's 123.6, a difference
of $-26.44$ [$-55.00$, $1.75$], while scoring 10.97 points higher
[6.04, 15.78]. On Indoor, Astra uses 8.23 more calls and scores 5.20 points
higher; on Outdoor it uses 1.64 more calls and scores 14.90 points lower.
The T2S and Outdoor score-difference intervals exclude zero, with opposite
signs. These selected-run comparisons do not estimate the causal effect of
additional tool calls or generation-to-generation variance.
\begin{table}[!t]
\centering
\compacttable
\caption{Current public-task paired differences, Astra minus Gemini, with 95\% source-cluster intervals. Scores use 0--100; tool counts exclude polling and correspond to the selected scored runs. T2S uses the reruns; I2S uses the earlier scenes with the corrected repair scores.}
\label{tab:trajectory-paired}
\footnotesize
\setlength{\tabcolsep}{3pt}
\begin{tabularx}{\linewidth}{>{\raggedright\arraybackslash}Xrr}
\toprule
\rowcolor{sbTableHead}
Task & $\Delta$ score [95\% CI] & $\Delta$ calls [95\% CI] \\
\midrule
Text-to-Scene & 10.97 [6.04, 15.78] & -26.44 [-55.00, 1.75] \\
Indoor & 5.20 [-7.38, 16.67] & 8.23 [5.25, 12.06] \\
Outdoor & -14.90 [-27.74, -2.03] & 1.64 [-0.32, 3.43] \\
\bottomrule
\end{tabularx}
\end{table}

\paragraph{Historical operation content behind tool calls.}
\label{app:trajectory-repair-profiles}
\begin{figure*}[!t]
\centering
\includegraphics[width=\textwidth]{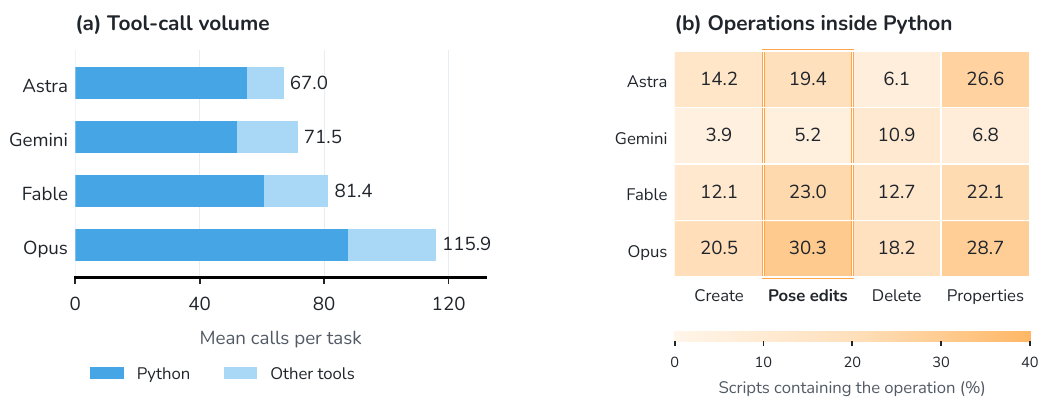}
\caption{Historical construction behavior, before the T2S reruns, of Astra, Gemini, Fable and Opus, the four
highest-ranked public configurations overall, on the same 16 retained T2S tasks.
(a) Mean non-polling calls per task, split into Python and other tools.
(b) Percentage of parseable Python scripts containing each operation family shown;
families can co-occur. The outlined column shows pose-editing operations.
All four configurations make zero named transform calls. Counts and definitions are in
Appendices~\ref{app:trajectory-repair-profiles} and~\ref{app:trajectory-extraction}.}
\label{fig:joint-trajectory-2}
\end{figure*}

Figure~\ref{fig:joint-trajectory-2} retains the earlier generation cohort to distinguish named-tool calls from operations inside Python. These per-script profiles are separate from the refreshed total calls in Table~\ref{tab:actual-tool-usage}. On the 16 public construction tasks, Astra, Gemini,
Fable and Opus submit 884, 832, 969 and 1,403 Python requests;
883, 384, 969 and 1,402 scripts, respectively, are parseable.
Pose-editing APIs occur in 171, 20, 223 and 425 of those scripts, despite
zero named transform calls for all four configurations. These counts describe
submitted operations and use the available code as their denominator.

Opus averages 115.9 calls per construction task, versus 81.4 for Fable,
while their matched historical mean T2S scores are $0.742$ and $0.772$. Gemini makes no named
Indoor asset searches, yet Python catalog APIs occur in 3/22 public and
12/50 private cases. Code submissions therefore expose editing and retrieval
attempts that named-tool counts alone omit.

\subsection{Failure Analysis}
\label{app:trajectory-execution}

\paragraph{Execution errors.}
\label{app:trajectory-python-feedback}
Object-access and type-related errors recur in frontier configurations:
\textrm{AttributeError} or \textrm{TypeError} occurs in 22/28 of Astra's
error-bearing public I2S submissions and 9/10 of Gemini's. Across all
public configurations and both settings, these families occur in $85.0\%$
of error-bearing submissions. These are shares of submissions with observed
errors, not error rates over all requests; missing feedback does not establish
error-free execution (Appendix~\ref{app:trajectory-extraction}).

\begin{figure}[!t]
\centering
\includegraphics[width=\linewidth]{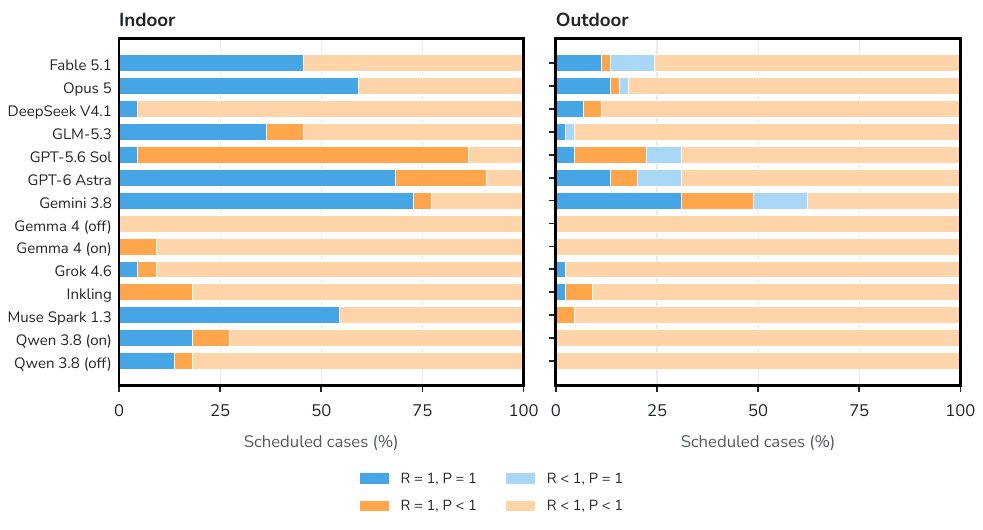}
\caption{Public repair outcomes by configuration (22 Indoor and 45 Outdoor
scheduled cases). All scheduled cases enter the denominators. R denotes
Recall and P denotes Precision; R=1 means complete target recovery, and P=1 means at least one recovered target and no false positives; the renamed-material export artifact described below is excluded.}
\label{fig:selective-repair-by-model}
\end{figure}

\paragraph{Final repair outcomes.}
\label{app:trajectory-repair-outcomes}
\label{app:additional-scores}
\label{app:selective-repair-analysis}
Astra leads Indoor target recovery, but requiring complete recovery with
zero false positives changes the comparison. Among submissions that recover every target, false positives
(FP) denote unmatched added actors or unintended modification, deletion or
repurposing of actors that should be preserved. On the 22 public Indoor cases, Astra
recovers every target in 20 cases and Gemini in 17; adding the requirement
of zero false positives changes these counts to 15 and 16, respectively.
On Outdoor, Gemini produces 22 complete recoveries and 14 without false
positives, compared with Astra's nine and six. Fable produces fewer complete
recoveries overall (16 across the 67 public repair cases), of which 15 have no false positives. Figure~\ref{fig:selective-repair-by-model} separates
complete recovery and false positives for all configurations;
Table~\ref{tab:repair-metrics} reports their domain-specific P/R/F1.

\begin{table}[!t]
\centering
\compacttable
\caption{Actor Precision / Recall / F1 (0--100) on public and private Image-to-Scene tasks, including the material-export-artifact correction. Each value averages all scheduled retained cases, with invalid or missing submissions scored as zero.}
\label{tab:repair-metrics}
\scriptsize
\setlength{\tabcolsep}{3pt}
\begin{tabularx}{\linewidth}{>{\raggedright\arraybackslash}Xrrrr}
\toprule
\rowcolor{sbTableHead}
Agent & Indoor Pub. & Indoor Priv. & Outdoor Pub. & Outdoor Priv. \\
\midrule
Gemini 3.8 Flash (high) & 73.9 / 77.3 / 74.5 & 57.1 / 72.0 / 60.3 & 51.9 / 57.3 / 47.3 & 57.5 / 60.6 / 55.8 \\
GPT-6 Astra (max) & 78.4 / 90.9 / 82.1 & \textemdash & 32.7 / 31.7 / 28.7 & \textemdash \\
Claude Opus 5 (max) & 59.1 / 59.1 / 59.1 & \textemdash & 30.4 / 29.7 / 28.9 & \textemdash \\
Claude Fable 5.1 (max) & 45.5 / 45.5 / 45.5 & \textemdash & 33.8 / 27.9 / 28.2 & \textemdash \\
GPT-5.6 Sol (high) & 31.3 / 86.4 / 43.8 & 30.9 / 76.0 / 41.2 & 26.8 / 38.7 / 25.3 & 23.5 / 49.4 / 29.4 \\
Muse Spark 1.3 (medium) & 54.5 / 54.5 / 54.5 & 34.7 / 38.0 / 35.7 & 8.0 / 11.9 / 8.9 & 11.9 / 16.0 / 12.9 \\
GLM-5.3 Flash (max) & 38.6 / 45.5 / 40.0 & 36.5 / 42.0 / 37.5 & 10.3 / 12.7 / 10.3 & 2.9 / 8.9 / 4.3 \\
Qwen 3.8 27B (thinking off) & 14.5 / 18.2 / 15.2 & 6.8 / 10.0 / 7.3 & 4.2 / 10.7 / 5.7 & 4.4 / 5.6 / 4.8 \\
Qwen 3.8 27B (thinking on) & 19.8 / 27.3 / 21.0 & 11.8 / 18.0 / 12.9 & 7.1 / 16.5 / 9.4 & 3.7 / 7.0 / 4.6 \\
Grok 4.6 (high) & 5.7 / 9.1 / 6.4 & \textemdash & 3.7 / 5.6 / 4.2 & \textemdash \\
Inkling (high) & 4.5 / 18.2 / 7.3 & 0.8 / 4.0 / 1.3 & 10.6 / 17.0 / 11.7 & 5.7 / 9.3 / 6.6 \\
Gemma 4 31B (thinking on) & 0.4 / 9.1 / 0.8 & 0.2 / 2.0 / 0.4 & 0.1 / 6.7 / 0.2 & 0.1 / 3.7 / 0.2 \\
Gemma 4 31B (thinking off) & 0.0 / 0.0 / 0.0 & 0.0 / 0.0 / 0.0 & 0.1 / 5.6 / 0.2 & 0.0 / 1.2 / 0.0 \\
DeepSeek V4.1 Flash (high) & 4.5 / 4.5 / 4.5 & 9.3 / 14.0 / 10.1 & 8.6 / 13.3 / 9.6 & 3.1 / 8.6 / 4.4 \\
\bottomrule
\end{tabularx}
\end{table}

These analyses exclude one export artifact. In the eight Indoor tasks built from the laboratory-loft scene, re-saving the scene renames a lighting blueprint's runtime material instance, which the scene comparison records as a modified non-target actor although no agent edited it. We remove this single false positive from the 88 affected submissions (41 public); the same correction is now applied to official Precision, Recall and F1 (Table~\ref{tab:main-results}; Table~\ref{tab:repair-metrics}), with TP/FN and physical scores unchanged. Across all 14 public configurations, 64 of 187 submissions that recover all targets also contain false positives ($34.2\%$; source-cluster 95\% CI, $25.5$--$42.3\%$). The 64 submissions span 37 tasks and 14 source clusters. Among them, 51 include changes to non-target actors and 27 include unmatched added actors;
Table~\ref{tab:repair-robustness}A separates their overlapping FP types.
FP counts have median three, interquartile range two--four, and maximum 2,886. Case prevalence complements actor counts so that large scenes do not
dominate the interpretation.

\paragraph{Repair tolerances and asset/material changes.}
Table~\ref{tab:repair-robustness}B first halves or doubles the default
error tolerances used to accept a repaired target. Target definitions,
actor correspondence and non-target comparison tolerances remain fixed.
The share of complete recoveries with FP stays near 34\%, indicating that
the recovery--preservation distinction persists across these target
acceptance thresholds. No non-target FP in a complete recovery is solely a within-tolerance pose change
(Appendix~\ref{app:gt-verifier}).

The final row excludes non-target modifications confined to asset/material
properties. These properties record asset and material references;
dynamic material-instance identities can change between saved packages
without establishing a visible change. The exclusion also removes actual
asset or material substitutions, so it is a diagnostic rather than a
correction to official scores. After exclusion, 63 public complete recoveries
retain FP across 13 configurations, including 50 with remaining non-target
changes. Only one of the 64 affected submissions contains only excluded modifications. For Astra, all five affected Indoor submissions and all three affected Outdoor submissions retain FP. Thus,
the distinction extends beyond asset/material differences.

\paragraph{Private-task check.}
Table~\ref{tab:repair-robustness}C compares public and private tasks for
the same ten configurations under the default target tolerances. Each entry reports submissions with
FP out of those recovering all targets. Separately excluding non-target modifications confined to asset/material properties leaves the private count unchanged at 80 of 171.
\begin{table}[!t]
\centering
\caption{\textbf{Unintended changes despite complete target recovery.}
Complete recovery means all required repairs succeed (Recall=1). Here, false
positives (FP) are added actors without a target match or unintended
modifications, deletions or repurposing of actors that should be preserved.
Counts refer to submissions (one configuration on one task), not actors.
A: submissions containing each change type; categories overlap and $n$ is
the number of complete recoveries. B: all 14 public configurations;
threshold triplets give distance (cm), rotation (degrees) and relative
scale/bounds-extent (\%) tolerances. The last row excludes non-target
changes confined to asset/material properties. C: submissions with FP / complete recoveries (percentage), for ten matched public/private configurations. All counts exclude the renamed-material export artifact (Appendix~\ref{app:trajectory-repair-outcomes}).}
\label{tab:repair-robustness}
\label{tab:repair-fp-types}
\label{tab:repair-sensitivity}
\label{tab:selective-repair-private}
\footnotesize
\setlength{\tabcolsep}{3pt}
\textbf{A. Types of unintended changes}\par\smallskip
\begin{tabularx}{\linewidth}{>{\raggedright\arraybackslash}Xrrr}
\toprule
\rowcolor{sbTableHead}
Change type & \shortstack{Public, 14 configs.\\$n=187$} & \shortstack{Public, 10 configs.\\$n=119$} & \shortstack{Private, 10 configs.\\$n=171$} \\
\midrule
Added actors without a target match & 27 & 23 & 38 \\
Modified non-target actors & 50 & 43 & 62 \\
Deleted non-target actors & 2 & 2 & 2 \\
Non-target actors repurposed for a repair & 2 & 2 & 0 \\
Unrecovered target state & 0 & 0 & 0 \\
\bottomrule
\end{tabularx}
\medskip
\textbf{B. Effect of repair tolerances and asset/material exclusions}\par\smallskip
\begin{tabularx}{\linewidth}{>{\raggedright\arraybackslash}Xrrr}
\toprule
\rowcolor{sbTableHead}
Analysis setting & \shortstack{All targets\\recovered} & \shortstack{Of these,\\with FP} & \shortstack{Share\\with FP} \\
\midrule
Strict (2.5 cm, $2.5^\circ$, 2.5\%) & 184 & 64 & 34.8\% \\
Default (5 cm, $5^\circ$, 5\%) & 187 & 64 & 34.2\% \\
Relaxed (10 cm, $10^\circ$, 10\%) & 190 & 64 & 33.7\% \\
Default; exclude changes confined to asset/material properties & 187 & 63 & 33.7\% \\
\bottomrule
\end{tabularx}
\medskip
\textbf{C. Public/private comparison for the same ten configurations}\par\smallskip
\begin{tabularx}{\linewidth}{Xcc}
\toprule
Domain & Public & Private \\
\midrule
Indoor & 30/75 (40.0\%) & 64/138 (46.4\%) \\
Outdoor & 23/44 (52.3\%) & 16/33 (48.5\%) \\
\bottomrule
\end{tabularx}
\end{table}

\paragraph{Observed tool use.}
\label{app:trajectory-resources}
Table~\ref{tab:actual-tool-usage} reports actual tool use by task setting and
public/private set. Counting conventions follow
Appendix~\ref{app:trajectory-extraction}.
\begin{table}[!t]
\centering
\compacttable
\caption{Mean non-polling tool calls per recorded selected run. Public T2S uses the current rerun cohort (except unchanged Opus and Inkling); I2S and private tasks retain their original runs. All 224 public T2S traces are observed. A dash denotes no private evaluation. Track-specific configurations follow Appendix~\ref{app:rerun-configurations}.}
\label{tab:actual-tool-usage}
\footnotesize
\setlength{\tabcolsep}{3pt}
\begin{tabularx}{\linewidth}{>{\raggedright\arraybackslash}Xrrrrrr}
\toprule
\rowcolor{sbTableHead}
System & T2S Pub. & T2S Priv. & Indoor Pub. & Indoor Priv. & Outdoor Pub. & Outdoor Priv. \\
\midrule
GPT-6 Astra (max) & 97.2 & \textemdash & 18.2 & \textemdash & 14.0 & \textemdash \\
Gemini 3.8 Flash (high) & 123.6 & 73.0 & 10.0 & 10.5 & 12.4 & 12.5 \\
Claude Fable 5.1 (max) & 72.1 & \textemdash & 11.3 & \textemdash & 14.0 & \textemdash \\
Claude Opus 5 (max) & 115.9 & \textemdash & 18.5 & \textemdash & 20.4 & \textemdash \\
GPT-5.6 Sol & 89.9 & 56.4 & 23.8 & 28.1 & 22.2 & 23.9 \\
GLM-5.3 Flash (max) & 167.4 & 115.2 & 24.8 & 22.6 & 13.8 & 15.4 \\
Muse Spark 1.3 & 78.4 & 74.4 & 56.3 & 55.1 & 17.7 & 21.3 \\
Grok 4.6 & 61.4 & \textemdash & 16.7 & \textemdash & 28.9 & \textemdash \\
Qwen 3.8 27B (thinking on) & 170.4 & 57.4 & 54.8 & 61.3 & 56.8 & 53.6 \\
Qwen 3.8 27B (thinking off) & 81.4 & 70.4 & 61.6 & 58.0 & 53.7 & 56.0 \\
Gemma 4 31B (thinking off) & 13.1 & 18.5 & 29.0 & 30.3 & 26.3 & 23.0 \\
Inkling (high) & 27.2 & 28.0 & 11.2 & 13.6 & 21.0 & 21.0 \\
Gemma 4 31B (thinking on) & 32.2 & 33.3 & 83.9 & 85.6 & 23.2 & 23.0 \\
DeepSeek V4.1 Flash & 158.7 & 72.8 & 11.5 & 14.5 & 24.9 & 22.4 \\
\bottomrule
\end{tabularx}
\end{table}

\paragraph{Estimated cost.}
\label{app:cost-estimate}
Figure~\ref{fig:teaser}(b) pairs current scores with token usage from their
selected generation attempts. Discarded attempts are excluded. Cache reads,
cache writes, fresh input and output are priced separately using the fixed
rates read between 7 and 23 September 2026, not current prices or billed spend.
Proprietary models use developer rates; Muse uses Meta's contributor tier.
Open-weight models use the recorded OpenRouter price snapshot, which is
counterfactual for self-hosted Gemma and Qwen. Standard tiers, base context/time
rates and observed cache reuse are assumed; non-token charges and discounts
are excluded.

All 192 new T2S selected scenes match their scoring-report hashes. Of these,
190 have reported total token usage that reconciles with the native event
stream. Two interrupted Qwen thinking-on runs have only per-request lower
bounds (\$1.882 and \$0.728 at the same rates); these remain separate and are
excluded from point estimates. Opus and Inkling retain their earlier T2S runs.
The 224 public T2S cases therefore have 222 reported-total cost records.

Table~\ref{tab:rerun-cost-coverage} reports cost and coverage by system. Domain
means use cases with reported totals and are weighted 16/22/45 over the 83
public tasks. Combined coverage ranges from 30/83 for Muse to 83/83 for Gemma
thinking off; Astra has 56/83 priced cases, including 21/45 Outdoor cases.
Missing usage is not assigned zero. The displayed frontier is descriptive
under these fixed prices and observed-usage coverage; token counts alone are
not cost ratios.
\begin{table}[!t]
\centering
\compacttable
\caption{Estimated token cost in USD at the fixed September 2026 price snapshot. T2S costs use the same selected runs as the current scores. Combined cost weights the observed domain means by 16/83, 22/83 and 45/83. Coverage counts cases with reported total usage; partial lower bounds are excluded. These are list-price estimates, not billed expenditure.}
\label{tab:rerun-cost-coverage}
\footnotesize
\setlength{\tabcolsep}{3pt}
\begin{tabularx}{\linewidth}{>{\raggedright\arraybackslash}Xrrrr}
\toprule
\rowcolor{sbTableHead}
System & T2S cost & Coverage & Combined cost & Coverage \\
\midrule
GPT-6 Astra (max) & 18.084 & 16/16 & 6.039 & 56/83 \\
Gemini 3.8 Flash (high) & 3.723 & 16/16 & 1.780 & 81/83 \\
Claude Fable 5.1 (max) & 25.347 & 16/16 & 7.280 & 77/83 \\
Claude Opus 5 (max) & 33.021 & 16/16 & 9.863 & 82/83 \\
GPT-5.6 Sol & 7.956 & 16/16 & 3.015 & 78/83 \\
GLM-5.3 Flash (max) & 1.106 & 16/16 & 0.287 & 38/83 \\
Muse Spark 1.3 & 0.315 & 16/16 & 0.115 & 30/83 \\
Grok 4.6 & 8.012 & 16/16 & 1.781 & 32/83 \\
Qwen 3.8 27B (thinking on) & 2.414 & 14/16 & 0.675 & 78/83 \\
Qwen 3.8 27B (thinking off) & 0.558 & 16/16 & 0.327 & 72/83 \\
Gemma 4 31B (thinking off) & 0.017 & 16/16 & 0.019 & 83/83 \\
Inkling (high) & 0.212 & 16/16 & 0.634 & 76/83 \\
Gemma 4 31B (thinking on) & 0.034 & 16/16 & 0.029 & 81/83 \\
DeepSeek V4.1 Flash & 0.568 & 16/16 & 0.154 & 68/83 \\
\bottomrule
\end{tabularx}
\end{table}

\subsection{Downstream Usability}
\label{app:eus-results}

The navigation study uses cohorts distinct from the main score and Physical
Safety comparisons. Table~\ref{tab:eus-leaderboard} reports construction results
for nine configurations (30 scheduled prompts; 29 retained GLM records),
Outdoor repair on the same 71 tasks for eight agent configurations, and Indoor
base-room coverage. These are historical auxiliary study cohorts, not additional
public/private splits, and include source scenes later retired from the benchmark.
Their results do not describe the retained 83-case public set. Methods and withholding rules are defined in
Appendix~\ref{app:eus-details}.

\paragraph{Downstream usability for embodied agents.}
\label{sec:downstream-usability}
\label{sec:distributional-embodied-results}
A generated scene is useful only if agents can still move through it. The auxiliary navigation study compares generated Outdoor scenes from editing with their
matched inputs to assess changes in downstream usability
(Section~\ref{sec:evaluation}). Gemini shows the largest observed mean improvement:
EUS increases from $0.264$ to $0.323$, a gain of $0.059$. Other evaluated
agent configurations show mean changes between $-0.031$ and $+0.014$.
 Study coverage, paired uncertainty and component
scores are reported in Appendix~\ref{app:eus-results}.

\begin{figure}[!t]
\centering
\includegraphics[width=\linewidth]{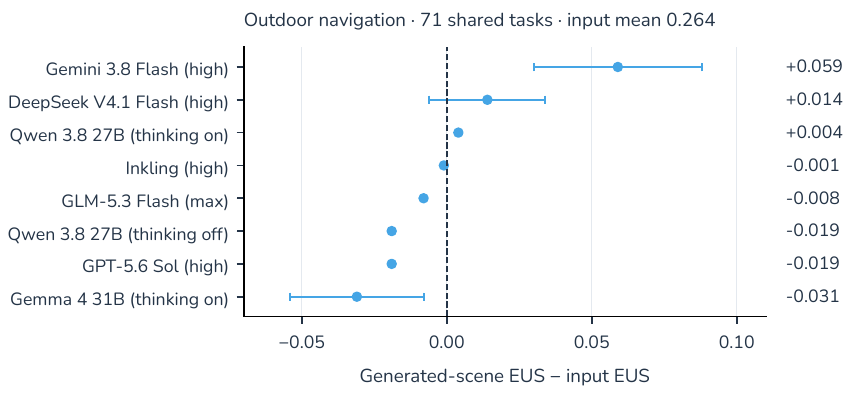}
\caption{Outdoor navigation changes relative to the matched input in the historical auxiliary cohort (71 tasks).
Dots show generated-scene-minus-input means on matched navigation tasks, using the
conservative withholding rule. Whiskers show paired standard
errors for Gemini, DeepSeek and Gemma; the other five marks show means only.
Differences use the reported three-decimal means. Generated-scene means, measured-only
means, components and coverage appear in Table~\ref{tab:eus-leaderboard}.}
\label{fig:eus-summary}
\end{figure}

\begin{takeaway}[label=take:downstream-usability]
Paired navigation tests separate editing gains from inherited usability. Gemini shows the largest observed EUS improvement ($+0.059$; Appendix Figure~\ref{fig:eus-summary}).
\end{takeaway}

\paragraph{Outdoor changes relative to the input.}
The matched task inputs were probed with the same navigation settings and
conservative withholding rule as the generated scenes. Figure~\ref{fig:eus-summary}
centres the comparison on the paired generated-scene-minus-input change, separating
inherited navigability from the change after editing. Seven mean changes lie
between $-0.031$ and $+0.014$; Gemini's is $+0.059$ with paired SE $0.029$.
Gemini and Sol have similar Outdoor walkability ($W=0.793$ and $0.799$),
but Gemini has higher approachability ($A=0.411$ versus $0.342$) and
generated-scene EUS ($0.323$ versus $0.245$). Gemini's higher EUS thus accompanies
broader target approachability, with comparable route-completion rates
(Table~\ref{tab:eus-leaderboard}).

\begin{table}[!t]
\centering
\caption{Embodied Utility Score (EUS); uncertainties are standard errors.
Conservative means assign zero to withheld
records; measured means omit them. Coverage is the measured share; Scored counts measured records; Status lists zero / withheld counts. Panel~A has 29 GLM-5.3 Flash (max) records and 30 per other agent;
Panel~B has 71 per agent. Panel~C summarizes indoor rooms.}
\label{tab:eus-leaderboard}
\footnotesize
\setlength{\tabcolsep}{3pt}
\setlength{\fboxsep}{2pt}
\noindent\colorbox{sbBlue!12}{\parbox{\dimexpr\linewidth-2\fboxsep}{\textbf{A. Text-to-Scene Construction (30 scheduled prompts)}}}\par\vspace{2pt}
\begin{tabularx}{\linewidth}{>{\raggedright\arraybackslash}Xccccccc}
\toprule
\rowcolor{sbTableHead}
Agent configuration & EUS $\uparrow$ & Measured $\uparrow$ & Coverage & $A\uparrow$ & $W\uparrow$ & Scored & Status \\
\midrule
Gemini 3.8 Flash (high) & \tbest{0.479 $\pm$ 0.050} & \tbest{0.513} & 93\% & \tbest{0.533} & 0.905 & 28 & 0 / 2 \\
GPT-5.6 Sol (high) & \tsecond{0.477 $\pm$ 0.068} & \tsecond{0.511} & 93\% & \tsecond{0.519} & \tbest{0.926} & 27 & 1 / 2 \\
GLM-5.3 Flash (max) & 0.387 $\pm$ 0.066 & 0.432 & 90\% & 0.478 & 0.778 & 24 & 2 / 3 \\
Qwen 3.8 27B (thinking off) & 0.373 $\pm$ 0.061 & 0.399 & 93\% & 0.420 & 0.810 & 28 & 0 / 2 \\
Gemma 4 31B (thinking on) & 0.369 $\pm$ 0.069 & 0.410 & 90\% & 0.419 & 0.792 & 24 & 3 / 3 \\
Gemma 4 31B (thinking off) & 0.347 $\pm$ 0.064 & 0.372 & 93\% & 0.389 & 0.736 & 24 & 4 / 2 \\
Muse Spark 1.3 (medium) & 0.344 $\pm$ 0.051 & 0.368 & 93\% & 0.380 & 0.852 & 27 & 1 / 2 \\
Inkling (high) & 0.268 $\pm$ 0.060 & 0.309 & 87\% & 0.328 & \tsecond{0.907} & 18 & 8 / 4 \\
Qwen 3.8 27B (thinking on) & 0.247 $\pm$ 0.058 & 0.274 & 90\% & 0.285 & 0.710 & 23 & 4 / 3 \\
\bottomrule
\end{tabularx}

\vspace{3pt}
\noindent\colorbox{sbOrange!14}{\parbox{\dimexpr\linewidth-2\fboxsep}{\textbf{B. Image-to-Scene Editing, Outdoor}}}\par\vspace{2pt}
\begin{tabularx}{\linewidth}{>{\raggedright\arraybackslash}Xccccccc}
\toprule
\rowcolor{sbTableHead}
Agent configuration & EUS $\uparrow$ & Measured $\uparrow$ & Coverage & $A\uparrow$ & $W\uparrow$ & Scored & Status \\
\midrule
Gemini 3.8 Flash (high) & \tbest[sbOrange]{0.323 $\pm$ 0.044} & \tbest[sbOrange]{0.396} & 82\% & \tbest[sbOrange]{0.411} & 0.793 & 58 & 0 / 13 \\
DeepSeek V4.1 Flash (high) & \tsecond[sbOrange]{0.278 $\pm$ 0.037} & \tsecond[sbOrange]{0.319} & 87\% & 0.340 & 0.788 & 62 & 0 / 9 \\
Qwen 3.8 27B (thinking on) & 0.268 $\pm$ 0.037 & 0.312 & 86\% & \tsecond[sbOrange]{0.344} & 0.775 & 60 & 1 / 10 \\
Inkling (high) & 0.263 $\pm$ 0.036 & 0.307 & 86\% & 0.323 & \tsecond[sbOrange]{0.839} & 60 & 1 / 10 \\
GLM-5.3 Flash (max) & 0.256 $\pm$ 0.034 & 0.313 & 82\% & 0.330 & \tbest[sbOrange]{0.854} & 58 & 0 / 13 \\
Qwen 3.8 27B (thinking off) & 0.245 $\pm$ 0.036 & 0.286 & 86\% & 0.323 & 0.783 & 60 & 1 / 10 \\
GPT-5.6 Sol (high) & 0.245 $\pm$ 0.035 & 0.295 & 83\% & 0.342 & 0.799 & 59 & 0 / 12 \\
Gemma 4 31B (thinking on) & 0.233 $\pm$ 0.035 & 0.250 & 93\% & 0.273 & 0.793 & 58 & 8 / 5 \\
\bottomrule
\end{tabularx}

\vspace{3pt}
\noindent\colorbox{sbOrange!7}{\parbox{\dimexpr\linewidth-2\fboxsep}{\textbf{C. Image-to-Scene Editing, Indoor (base rooms, not ranked)}}}\par\vspace{2pt}
\begin{tabularx}{\linewidth}{>{\raggedright\arraybackslash}Xccc}
\toprule
\rowcolor{sbTableHead}
Agent configuration & Rooms measured & Mean EUS & Median EUS \\
\midrule
DeepSeek V4.1 Flash (high) & 6/10 & 0.194 & 0.192 \\
Gemini 3.8 Flash (high) & 6/10 & 0.207 & 0.189 \\
Gemma 4 31B (thinking on) & 8/10 & 0.220 & 0.248 \\
Gemma 4 31B (thinking off) & 9/10 & 0.279 & 0.242 \\
GLM-5.3 Flash (max) & 6/10 & 0.169 & 0.157 \\
GPT-5.6 Sol (high) & 7/10 & 0.227 & 0.289 \\
Inkling (high) & 6/10 & 0.194 & 0.187 \\
Qwen 3.8 27B (thinking off) & 8/10 & 0.158 & 0.139 \\
Qwen 3.8 27B (thinking on) & 8/10 & 0.184 & 0.203 \\
\bottomrule
\end{tabularx}
\end{table}

\paragraph{Construction results.}
Gemini and Sol have nearly identical mean construction EUS ($0.479$ and
$0.477$), with a paired difference of $0.002\pm0.061$ on shared prompts.
Their Outdoor repair means diverge: Gemini gains
$0.059$ over the shared input mean, while Sol changes by $-0.019$.
Similar usability after construction therefore does not imply similar
navigation gains after editing; the construction and repair cohorts are
assessed separately.

\paragraph{Alignment with benchmark scores.}
Eight configurations are shared by the construction EUS summary and the benchmark-score summary available at the time of the EUS study. Both studies used 30 scheduled prompts; GLM
has 29 retained EUS records. The association concerns configuration means,
not paired individual scenes. The alignment uses the eight configurations matched when the EUS summary was produced; DeepSeek has no construction EUS result. Adding Gemma 4 31B (thinking off), whose reasoning mode was confirmed afterwards, gives $r=0.752$ and $\rho=0.667$. From the reported three-decimal configuration means, Pearson
correlation is $r=0.781$ and Spearman rank correlation is $\rho=0.714$. Using
measured-only EUS gives $r=0.731$ and $\rho=0.643$. Construction and repair are analyzed separately because their navigation
cohorts differ.

\paragraph{Indoor repair.}
The indoor cohort contains eight edits of each of ten base rooms; rooms, not edits, are the
source-level units. The number of rooms with a numeric result ranges from six
to nine across systems, and the corresponding room means range from \(0.158\)
to \(0.279\). An unmeasured room has no approachable target beyond the grading
radius for any of its edits. Because room coverage differs across systems,
Panel~C uses alphabetical agent order.

\paragraph{Navigation measurement coverage.}
Text-to-Scene has two to four withheld scenes per configuration, due to mesh
settling, route availability, or certification failures. Outdoor repair has
five to thirteen withheld scenes per configuration, including 58
configuration--task measurements with mesh-settling failures. These results
follow the withholding and exclusion rules in Appendix~\ref{app:eus-details}.

\subsection{Evaluator Validation}
\label{app:evaluator-validation-results}

Table~\ref{tab:human-fit-results} summarizes human agreement, judge
repeatability and repair-weight sensitivity. Human-review and repeatability
protocols appear in Appendices~\ref{sec:human-alignment}
and~\ref{app:judge-repeatability-protocol}.

\begin{table}[!t]
\centering
\caption{Evaluator validation and its scope. Panel A separates three validation
questions; Panel B reports agreement fractions for the retrospective
12/4-prompt split. Fitted weights are in
Table~\ref{tab:human-alignment-weights}.}
\label{tab:human-fit-results}
\footnotesize
\setlength{\tabcolsep}{4pt}
\begin{tabularx}{\linewidth}{>{\raggedright\arraybackslash}p{0.16\linewidth}>{\raggedright\arraybackslash}p{0.20\linewidth}>{\raggedright\arraybackslash}X>{\raggedright\arraybackslash}X}
\toprule
\multicolumn{4}{l}{\textbf{A. Validation questions}} \\
\rowcolor{sbTableHead}
Question & Unit & Result & Scope \\
\midrule
Human agreement & Four holdout prompts, 84 consensus pairs &
73/84 agreement; 95\% CI 76.19--94.05\%; $\tau_b=0.738$ &
Internal validation: earlier exploration used all 16 prompts. \\
Judge repeatability & Nine configurations, six cases, three repeats &
Rank $\tau_b=0.889$--$0.944$; model-mean SD 0.0010--0.0132 &
Fixed evidence and routing; not complete adaptive-verifier variability. \\
Repair-weight sensitivity & Fourteen configurations, 67 retained public repairs &
For tested $\lambda_{\mathrm{rep}}\geq0.5$, rank $\rho\geq0.960$; at most two places shifted &
Fixed scenes, thresholds and cases; no human labels fitted. \\
\bottomrule
\end{tabularx}
\medskip
\begin{tabularx}{\linewidth}{Xccc}
\toprule
\multicolumn{4}{l}{\textbf{B. Policy agreement}} \\
\rowcolor{sbTableHead}
Policy & Dev. individual & Holdout individual & Holdout PL \\
\midrule
Adopted & 0.7672 & \tbest{0.7937} & \tbest{0.8690} \\
Individual fit & \tbest{0.8082} & 0.7619 & 0.8214 \\
PL fit & \tbest{0.8082} & 0.7460 & 0.7976 \\
\bottomrule
\end{tabularx}
\end{table}

\paragraph{Agreement with human judgments.}
The preference fits improve development agreement but do not improve the
retrospective holdout results. Across all 16 prompts, pairwise inter-annotator
agreement is 80.75\% and nominal Krippendorff's $\alpha_{\mathrm K}=0.613$. On the four
holdout prompts, individual-preference agreement is 79.37\% for the adopted
policy and 80.95\% between annotators. Separate reporting distinguishes consensus from individual-preference labels.

\begin{figure}[!t]
\centering
\includegraphics[width=\linewidth]{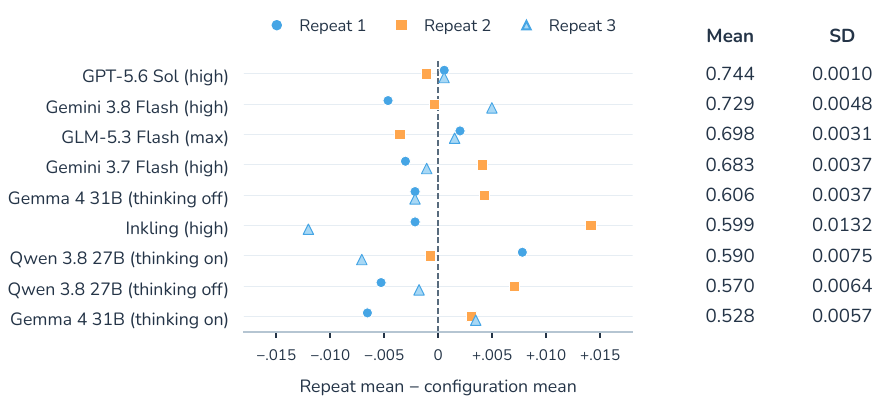}
\caption{Detailed Alignment repeatability on six selected T2S cases. Marks
show each repeat's six-case mean minus the configuration's three-repeat mean;
colors and shapes distinguish repeats. The right columns report the absolute
mean and sample standard deviation across the three repeats. Evidence and
routing are fixed.}
\label{fig:judge-repeatability}
\end{figure}

\paragraph{Detailed Alignment repeatability.}
\label{app:judge-repeatability}
All 162 evaluations completed. Response validation produced 111 fallbacks
recorded as \textsc{Unknown} and handled by the aggregation policy.
Figure~\ref{fig:judge-repeatability} shows that the four highest configurations
retain their ordering, with reversals confined to ranks 5--7. Pairwise rank
agreements are 0.889, 0.944 and 0.944. Inkling has the largest variation in
repeat means.

\paragraph{Repair-weight sensitivity.}
\label{app:i2s-weight-sensitivity}
Table~\ref{tab:i2s-weight-sensitivity} varies the Repair F1 weight in the
existing I2S aggregate. Gemini remains the leader at tested weights
$\lambda_{\mathrm{rep}}\in\{0.25,0.5,0.6,0.7,0.8,0.9,1.0\}$; physics-only scoring selects
GLM. This checks sensitivity on fixed public cases, not whether 0.8 is optimal.
\begin{table}[!t]
\centering
\caption{I2S ranking sensitivity to the Repair F1 weight for 14 configurations on 67 retained public repairs. Correlations and maximum rank shifts are relative to $\lambda_{\mathrm{rep}}=0.8$ in $\lambda_{\mathrm{rep}} F_1+(1-\lambda_{\mathrm{rep}})S_{\mathrm{physics}}$. Scenes, thresholds and case selection remain fixed.}
\label{tab:i2s-weight-sensitivity}
\small
\begin{tabular}{rrrr}
\toprule
$\lambda_{\mathrm{rep}}$ & Spearman $\rho$ & Kendall $\tau_b$ & Max. rank shift \\
\midrule
0.00 & 0.802 & 0.670 & 6 \\
0.25 & 0.925 & 0.802 & 3 \\
0.50 & 0.960 & 0.846 & 2 \\
0.60 & 0.978 & 0.912 & 2 \\
0.70 & 0.996 & 0.978 & 1 \\
0.80 & 1.000 & 1.000 & 0 \\
0.90 & 0.996 & 0.978 & 1 \\
1.00 & 0.987 & 0.956 & 2 \\
\bottomrule
\end{tabular}
\end{table}

\section{Related Work}
\label{app:additional-related-work}
\label{sec:related-work}

\paragraph{Reading the benchmark comparison.}
Table~\ref{tab:benchmark-positioning} distinguishes task evidence, artifacts
and evaluation coverage. Input references specify the desired outcome and
exclude initial-scene observations, code and execution feedback. Open-ended
tasks allow multiple valid realizations; ground-truth-based tasks provide a
target scene or state, including reference sequences. Executable assertions
alone do not constitute a target scene. Checkmarks indicate reported coverage,
not equivalent scope or accuracy; blank cells indicate no corresponding
component reported. Physical validity may concern static geometry or dynamics,
whereas preservation requires explicit checks of non-target content.

SceneActBench compares outputs with hidden geometry, pose and motion targets
without a separate physical-validity score \citep{zhao2026sceneactbench}.
4DBuildBench audits physics mechanisms and spatial relations alongside visual
prompt fulfillment \citep{liu2026simworlds}. CutsceneBench combines cinematic
judgments with reference-sequence checks; its agent-side collision-avoidance
tools are not a benchmark physical metric \citep{he2026cutscene}.
Code4Scene measures static support and penetration.

\paragraph{Environment generation and executable representations.}
\label{app:environment-generation}
Captured spaces preserve acquired geometry, while procedural systems generate
controlled variation through rules \citep{xia2018gibson,deitke2022procthor,
raistrick2023infinigen,raistrick2024infinigenindoors}. Language-conditioned
systems add asset retrieval and layout synthesis; recent coding-agent systems
construct editable environments through planners, engine tools and executable
scene programs \citep{yang2024holodeck,pfaff2026scenesmith,xia2026sage,
kang2026simworldstudio,wang2026scenecode,guo2026worldclaw}. Executable scenes
also support embodied-agent training and evaluation, e.g., urban navigation
and multi-robot collaboration in SimWorld-Robotics
\citep{zhuang2025simworldrobotics} and long-horizon delivery planning in
DeliveryGym \citep{kang2026deliverygym}. Code4Scene evaluates
complete agent configurations using a shared Unreal interface, task assets
and accounting rules. Representation and harness choices can themselves
change performance \citep{liu2026spatialbabel,yao2026harnessbench}; the benchmark
does not isolate spatial reasoning from competence with the execution interface.

\paragraph{Construction, editing and preservation.}
BlenderGym evaluates edits from start scenes toward goal renders
\citep{gu2025blendergym}. AuthorBench combines construction and language-guided
editing with requirement, preservation and physical checks on scene JSON
\citep{xu2026muse}. VWE-Bench uses language to specify changes to placed-asset
worlds, with current-scene images as context; its verified queries check
required edits and reject unauthorized changes \citep{ning2026vibeworlding}.
Code4Scene's Image-to-Scene references instead depict the desired human-assembled
scene. Agents infer the repair from visual differences, then submit a persistent
Unreal scene that is reopened for integrity, physical and task checks.
These differences concern evidence, interfaces and artifacts; preservation
and physical validity are also evaluated in prior work.

\paragraph{Evaluation beyond rendered appearance.}
Requirement-based and multidimensional evaluation separate object, attribute,
relation and visual-quality judgments \citep{hu2023tifa,huang2024vbench,
he2023t3bench}. Symbolic scene evaluation and engine measurements provide
complementary evidence to visual judgments \citep{sengupta2026scenecritic,
tam2026sceneeval}. Code4Scene combines structured scene-state checks with
bounded visual judgments and separately measures navigation with a fixed
route follower (Appendix~\ref{app:eus-details}). This complements navigation proxies in MUSE \citep{xu2026muse} without measuring general embodied
interaction.

\section{Limitations and Broader Impact}
\label{sec:limitations}

\paragraph{Scope and transfer.}
Code4Scene evaluates static scene construction and editing in Unreal Engine. It
does not evaluate authored dynamics, e.g., cloth, fluids, or articulated
mechanisms \citep{liu2026simworlds}. The Text-to-Scene support score is a
bounding-box contact proxy with a world-ground assumption, not a test of mesh
contact, load-bearing stability, or whether a chain of supports reaches the
ground. The auxiliary fixed-walker analysis (EUS) asks
whether a person-sized body can reach and traverse the scene, but does not test
object interaction, manipulation, learned-policy task completion, or multi-agent behavior \citep{chang2024partnr,pfaff2026scenesmith}. Performance can also vary
substantially with the scene representation and action interface \citep{liu2026spatialbabel,yao2026harnessbench}. The execution logs expose issued calls and returned
failures, but they do not fully disentangle competence with the Unreal
interface from scene-construction quality. Comparisons characterize complete
agent configurations, including their clients (Appendix~\ref{app:agent-interface}).

\paragraph{Statistical and evaluator validity.}
Each score cohort retains one scene per agent configuration and case,
without estimating variability across repeated agent trajectories.
The paired source-cluster bootstrap intervals describe variation across
sampled cases with those scenes fixed
(Appendix~\ref{app:trajectory-coverage}). Leaderboard stability under repeated
scene generation remains unmeasured. Agreement with human preferences was
assessed for Text-to-Scene on 16 prompts and seven agent configurations, with only
four prompts in the retrospective holdout. Reviewers saw four overview images,
whereas the automatic score also uses engine measurements and targeted views.
Image-to-Scene has not been validated against human judgments, and the
fixed-evidence VLM study does not measure repeatability of the complete adaptive
evaluator. The actor-repair tolerances and $0.8/0.2$ repair/physics weighting
are design choices, not human-calibrated optima. Table~\ref{tab:i2s-weight-sensitivity}
retrospectively tests ranking sensitivity to weights on fixed scenes, not to
repair thresholds or generation randomness.

\paragraph{Dataset and annotation coverage.}
Public-set agreement with full-suite scores is measured for ten shared
configurations; full-suite error remains unmeasured for the four
agent configurations evaluated only on the public set (Appendix~\ref{app:public-private}).
The public set is
not proportional across categories, content packages or difficulty tiers.
The cases derive from human-assembled Unreal scenes, so their coverage reflects
the available content packages rather than a sample from a defined population.
The authored difficulty labels are not empirically calibrated and do not
guarantee decreasing scores for every model or task setting
(Table~\ref{tab:difficulty-results}).
Indoor cases do not record the same operation-level corruption contracts as outdoor
cases. Their author-provided operation groups support descriptive category
analysis but not operation-chain conclusions
(Section~\ref{sec:repair-category-analysis}). Requirement graphs are produced
automatically or with agent assistance rather than independently hand-annotated;
an error in a graph can therefore propagate into the score.

% AUTHOR REMINDER: Keep participant and institution identities anonymous.
% Confirm applicable consent, compensation, and ethics-review reporting facts;
% do not claim approval or exemption without confirmation.
% AUTHOR REMINDER: Confirm third-party asset terms before the planned release.

\paragraph{Broader impact.}
Reliable construction of executable 3D scenes could increase the availability
of environments for embodied and physical AI. It could also reduce demand for
some manual authoring work or shift creative labor toward supervision and
revision. Failures such as hidden collisions, missing dependencies, or
unintended edits may propagate into downstream experiments if generated scenes
are accepted without independent checks; a high Code4Scene score is not a
general safety certification. Because the cases use existing content packages,
access and redistribution remain subject to the packages' original licenses.

% \section{Conclusion}
% \label{sec:conclusion}

% Code4Scene establishes a common Unreal Engine protocol for evaluating coding
% agents on open-ended construction and reference-guided repair. The component
% scores distinguish semantic alignment, physical validity, target
% recovery, and preservation, thereby exposing failures that an overall score
% would otherwise conflate.

% The frontier configurations exhibit complementary strengths: Astra leads
% Transform repairs, Fable leads construction, and Gemini leads Lifecycle,
% Layout and overall repair. These strengths coexist with failures in Spatial
% Composition and incomplete control of collateral edits. Astra's high Indoor
% recovery rate, alongside Gemini's larger count of complete repairs without
% false positives, shows why target recovery and scene preservation must be
% assessed together. In a separate navigation cohort, Gemini has the largest
% observed Outdoor gain over the input scenes, making the contribution of
% editing visible beyond the final utility score.
% Future work should validate Image-to-Scene scores against human judgments,
% measure run-to-run variability, and extend evaluation from static scene
% validity to dynamic interaction.

% \input{appendix/C_additional_results}
% \input{appendix/D_reproducibility}

\FloatBarrier
\Needspace{5\baselineskip}
\section*{Acknowledgments}

We thank Abhay Anand and Yash Vishe for their support with benchmark
construction, and Haoqiang Kang for helpful feedback on the writing of this
paper.

\paragraph{AI-assisted work.}

Generative AI assisted language editing, figure prototyping, analysis planning
and interpretation, code implementation and debugging, task-prompt drafting,
and requirement-bundle construction. The authors reviewed the manuscript,
figures, and analysis code, verified reported results against the underlying
artifacts, and take responsibility for the final work. Task prompts were
human-reviewed; frozen requirement bundles
received automated checks but no human review
(Section~\ref{sec:benchmark-suite}; Appendix~\ref{app:semantic-verifier}).

\end{document}